%% file: main.tex
\documentclass[11pt, a4paper, logo]{googledeepmind}

\usepackage{amsmath}
\usepackage{amssymb}
\usepackage{mathtools}
\usepackage{amsthm}
\usepackage{microtype}
\usepackage{graphicx}
\usepackage{booktabs} %

\input{math_commands.tex}

\usepackage[utf8]{inputenc} %
\usepackage[T1]{fontenc}    %
\usepackage{hyperref}       %
\usepackage{url}            %
\usepackage{booktabs}       %
\usepackage{amsfonts}       %
\usepackage{nicefrac}       %
\usepackage[labelformat=simple]{subcaption}

\usepackage{multirow}
\usepackage{makecell}
\usepackage{graphicx}
\usepackage{algorithm}
\usepackage{algorithmic}
\usepackage{amssymb} %
\usepackage{pifont} %
\usepackage{adjustbox}
\usepackage{enumitem}
\usepackage{setspace} %
\usepackage{etoolbox}
\usepackage{xspace} %
\usepackage{wrapfig} %
\usepackage{array}
\usepackage{calc}
\usepackage{tabularx}

\usepackage[capitalize,noabbrev]{cleveref}

\usepackage[authoryear, sort&compress, round]{natbib}
\title{Transformers Can Achieve Length Generalization But Not Robustly}

\correspondingauthor{yczhou@cs.toronto.edu}

\renewcommand{\today}{}

\author[1,2]{Yongchao Zhou}
\author[1]{Uri Alon}
\author[1]{Xinyun Chen}
\author[1]{Xuezhi Wang}
\author[1]{Rishabh Agarwal}
\author[1]{Denny Zhou}

\affil[1]{Google DeepMind}
\affil[2]{University of Toronto}

\newcommand{\uri}[1]{{\color{magenta}Uri: #1}}
\newcommand{\yc}[1]{{\color{purple}YC: #1}}
\newcommand{\rish}[1]{{\color{blue}RA: #1}}
\newcommand{\xinyun}[1]{{\color{red}XY: #1}}

\renewcommand{\uri}[1]{}
\renewcommand{\yc}[1]{}
\renewcommand{\rish}[1]{}
\renewcommand{\xinyun}[1]{}

\renewcommand{\paragraph}[1]{\textbf{#1} \hspace{0.6em}}

\newif\ifworkshop
\workshopfalse
\ifworkshop
  \newcommand{\workshoponly}[1]{#1}
  \newcommand{\workshopexclude}[1]{}
\else
  \newcommand{\workshoponly}[1]{}
  \newcommand{\workshopexclude}[1]{#1}
\fi

\begin{abstract}
Length generalization, defined as the ability to extrapolate from shorter training sequences to longer test ones, is a significant challenge for language models. This issue persists even with large-scale Transformers handling relatively straightforward tasks. In this paper, we test the Transformer's ability of length generalization using the task of addition of two integers. We show that the success of length generalization is intricately linked to the data format and the type of position encoding. Using the right combination of data format and position encodings, we show for the first time that standard Transformers can extrapolate to a sequence length that is $2.5\times$ the input length. Nevertheless, unlike in-distribution generalization, length generalization remains fragile, significantly influenced by factors like random weight initialization and training data order, leading to large variances across different random seeds.
\end{abstract}

\begin{document}

\maketitle

\input{1-introduction}

\input{2-background}

\input{3-method}

\input{4-result}

\input{5-related_work}

\section{Conclusion}

Length generalization in Transformers has been a long-standing challenge. 
We evaluate the ability of Transformers to generalize to longer test sequences using the decimal addition task.
Through extensive experiments, we find that there is no inherent limitation in Transformers’ design preventing effective
length generalization. Instead, the missing ingredient is the right combination of data format and position encoding.
We demonstrate that Transformers can achieve almost perfect generalization on sequences up to $2.5\times$ the training length, given appropriate data formatting and position encoding. %

Our thorough empirical analysis of common length generalization techniques reveals a significant dependency between the type of position encoding and the data format. This underscores the importance of synergizing data format with model architecture for optimal generalization.
Despite these advancements, robust length generalization in Transformers remains elusive, even with meticulously finetuned regularization hyperparameters. 

\bibliography{main}

\newpage
\appendix
\onecolumn

\counterwithin{figure}{section}
\counterwithin{table}{section}
\counterwithin{equation}{section}
\counterwithin{algorithm}{section}

\input{appendix/positional_encoding}
\input{appendix/implementation_details}
\input{appendix/additional_results}

\end{document}

%% file: math_commands.tex
\usepackage{amsmath,amsfonts,bm}
\usepackage[normalem]{ulem}

\def\eqref#1{equation~\ref{#1}}

\def\1{\bm{1}}

\def\ve{{\bm{e}}}

\def\vp{{\bm{p}}}

\def\mA{{\bm{A}}}
\def\mB{{\bm{B}}}

\def\mW{{\bm{W}}}
\def\mX{{\bm{X}}}

\DeclareMathAlphabet{\mathsfit}{\encodingdefault}{\sfdefault}{m}{sl}
\SetMathAlphabet{\mathsfit}{bold}{\encodingdefault}{\sfdefault}{bx}{n}

\def\sN{{\mathbb{N}}}

\def\sR{{\mathbb{R}}}

\newcommand{\R}{\mathbb{R}}

%% file: 1-introduction.tex
\section{Introduction}\label{sec:intro}

\workshopexclude{
\begin{figure}[b!]
\centering
\begin{subfigure}[t]{0.5\textwidth}
    \includegraphics[width=1.0\linewidth]{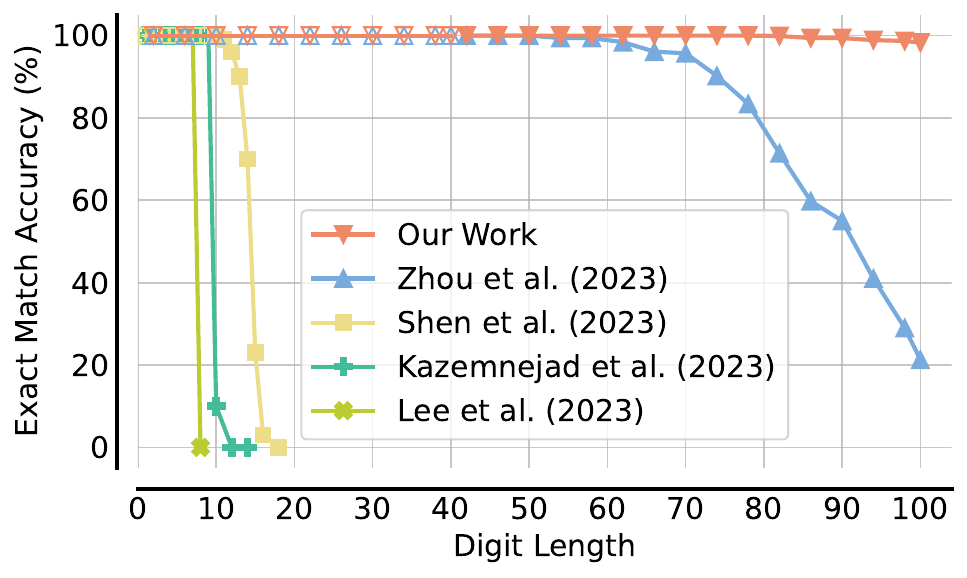}
\end{subfigure}
\caption{%
Using an appropriate position encoding and data formatting, we demonstrate that Transformers can generalize to 100-digit decimal addition tasks with more than 98\% of accuracy when trained up to 40-digit addition, resulting in a length extension ratio of $2.5\times$, 
which is much more than the ratio of \citet{lee2023teaching} ($1.0\times$), \citet{kazemnejad2023impact} ($1.125\times$), \citet{shen2023positional} ($1.1\times$), and \citet{zhou2023algorithms} ($1.5\times$).
Unfilled markers (\textemdash\!\!\!\!\!${\color{white}\blacktriangledown}$\!\!\!\!\!\,\!\!\;$\triangledown$\,) denote in-distribution test results, filled markers (\textemdash\!\!\!\!\!$\blacktriangledown$) denote out-of-distribution results. 
In \citet{zhou2023algorithms} and Our Work, each curve is the best out of 10 trials. For the other three methods, we report the value from their corresponding paper. 
}\label{fig:main_performance}
\end{figure}
}

Transformer-based models have revolutionized natural language understanding and generation across diverse applications \citep{openai2023gpt4tr,team2023gemini}. Despite their impressive abilities in mathematical reasoning \citep{lewkowycz2022solving}, code synthesis \citep{li2022competition}, and theorem proving \citep{wu2022autoformalization}, Transformers often struggle with length generalization, an ability that requires the model to generalize to longer sequences than seen during training \citep{anil2022exploring, abbe2023generalization, zhou2023algorithms}. This limitation raises an essential question: do Transformers genuinely grasp the correct underlying algorithms for a given task, or are they merely resorting to superficial memorization or shortcuts that fail to scale to more complex problems \citep{liu2023transformers}? 

Recent work has scrutinized Transformers' shortcomings in length generalization across formal language learning \citep{deletang2023neural} and algorithmic reasoning tasks \citep{anil2022exploring, zhang2022unveiling, velivckovic2022clrs, dziri2023faith}. These investigations consistently indicate a notable deficiency in length generalization capabilities. This recurring issue raises a crucial question: Is there an inherent limitation in Transformers' design preventing effective length generalization?

In this paper, we systematically examine the Transformer's capability of length generalization, specifically focusing on the $N$-digit decimal addition problem. We view the addition problem as a form of synthetic language learning, which despite its relative simplicity compared to natural language, provides valuable insights into the Transformer's ability to internalize fundamental algorithms. Notwithstanding its simplicity, recent work has demonstrated that Transformers exhibit limited length generalization in this task \citep{lee2023teaching, shen2023positional, kazemnejad2023impact}.

Previous attempts to improve Transformer's length generalization ability primarily focus on two areas: refining position encodings \citep{shen2023positional, press2022train} and optimizing data formats \citep{lee2023teaching, zhou2023algorithms}. Therefore, we perform an extensive empirical evaluation of combinations of widely used position encoding and various data formats, resulting in a recipe for successful length generalization. Our final recipe consists of: FIRE position encodings \citep{li2023functional}, with randomized positions \citep{ruoss2023randomized}, 
in reversed format, with index hints \citep{zhou2023algorithms}.

As shown in \Cref{fig:main_performance}, when trained on only 40 digits, our model successfully extrapolates to sequences of up to 100 digits, exceeding the input length by $2.5\times$.
To the best of our knowledge, this is the strongest known generalization result for text-based Transformers on addition. Nevertheless, we observe that the robustness of this length generalization is fragile, significantly swayed by variables such as random initialization and the training data order. 
\xinyun{The 2 (equally) important conclusions of this paper are: (1) Our approach achieves much better length generalization with XXX positional encoding and YYY data format; and (2) The generalization performance is sensitive to several factors. I personally think the 2nd point was not sufficiently discussed in prior work and is important to emphasize. However, the current introduction does not tell about what is the best combination, and there is very little discussion on the second point. I suggest to rewrite the bullet points and highlight these 2 conclusions, and discussion on redundant techniques is not necessary in the intro.}

Our key contributions are summarized as follows:
\begin{enumerate}[label=(\roman*),topsep=2pt,itemsep=2pt, parsep=2pt, leftmargin=16pt]
    \item We demonstrate that the success in length generalization is markedly influenced by position encoding and data format. Through careful selection of these factors, we achieved extrapolation to lengths that are $2.5 \times$ longer than those seen during training.
    \item Our exploration of established data formatting and augmentation techniques indicates that their effectiveness in length generalization is primarily contingent on the choice of position encoding. 
    \item Despite remarkable generalization to lengths $2.5 \times$ longer than training, we found this generalization to be fragile and heavily relying on factors like random weight initialization and training data order.
\end{enumerate}

%% file: 2-background.tex
\section{Position Encoding and Data Formats} \label{sec:background}

Recently proposed improvements in architectural design, notably in position encoding \citep{shen2023positional, kazemnejad2023impact, ruoss2023randomized} and attention mechanisms \citep{dubois2019location, duan2023interpolation}, aim to address the challenge of length generalization in arithmetic computations with Transformers. However, the effectiveness of such modifications is often constrained, either due to their overly ad-hoc nature or their poor performance on longer sequences. Although scaling the size of models and datasets has been recognized as a generally effective strategy to improve performance, prior research \citep{brown2020language, anil2022exploring} suggests that relying solely on scale might not be sufficient for handling test sequences that are longer than training. Concurrently, with the rising focus on data-centric AI \citep{motamedi2021data}, recent work has investigated refining the data format to enhance the learning efficacy of existing Transformer models. 
In this section, we review some of the most common position encodings (\Cref{sec:pe}) and relevant data formats (\Cref{sec:background_data})

\subsection{Position Encoding for Length Generalization}\label{sec:pe}
The inability of transformers to extrapolate to longer sequences has been primarily attributed to position encoding \citep[PE; ][]{shaw2018self}. In this section, we review existing positional encoding approaches with an emphasis on their length generalization abilities. 

\paragraph{Absolute Positional Encoding (APE).} APE enhances Transformer models with positional information by attaching a positional vector $\vp_i$ to each position $i$. 
This is achieved through a predefined sinusoidal function \citep{vaswani2017attention} or a learnable approach \citep{devlin2018bert}. Then, the vector $\vp_i$ is combined with the token embedding $\ve_i$ before entering the transformer's first layer. Although straightforward, APE often struggles with generalizing to longer sequences, as observed in both NLP \citep{press2022train} and algorithmic tasks \citep{kazemnejad2023impact}.

\paragraph{Additive Relative Positional Encoding (RPE).} \citet{shaw2018self} pioneered the additive RPEs, diverging from standard input-level integration by modifying keys and, optionally, values in each attention layer. This concept was advanced by T5, which employed scalar biases to directly affect pre-softmax attention logits, a method noted for its simplicity yet criticized for limited efficiency and positional differentiation in long sequences \citep{raffel2020exploring, press2022train}. Later approaches such as Alibi \citep{press2022train}, Kerple \citep{chi2022kerple} and FIRE \citep{li2023functional} build on the idea of learned additive bias, proposing different functions to model the scalar bias as a function of the key- and query-indices.
Most pre-softmax attention logits of 
additive RPEs can be generally written as \citep{li2023functional}:
\begin{equation}
    \label{eq:rpe-attn-mat}
    \mA_{\mathrm{RPE}}(\mX) = \mX \mW_Q(\mX \mW_K)^{\top}+\mB,
\end{equation}
where $\mX$, $\mW_Q$, $\mW_K$ denote the input and weight matrices for queries and keys. The bias matrix $\mB\in\R^{n\times n}$ is induced by the \textbf{position encoding function} $b: \sN^{*2}\to\R$, with its $(i,j)$-th entry defined as $b(i,j)$. Instances of $b(i,j)$ include:
\begin{itemize}[topsep=2pt,itemsep=2pt, parsep=2pt, leftmargin=10pt]
    \item T5 \citep{raffel2020exploring}: $b(i,j) = r_{min}\{i-j,K\}$, where $K$ is a hyperparameter and $r_i$ are learned scalars.
    \item Alibi \citep{press2022train}: $b(i,j) = -r\left|i-j \right|$, where $r>0$ is a hyperparameter.
    \item KerpleLog \citep{chi2022kerple}: $b(i,j) = -r_1\log(1+r_2|i-j|)$, where $r_1, r_2>0$ are learnable scalars.
    \item FIRE \citep{li2023functional}: $b(i,j) = f_{\theta}\left(\frac{\psi(i-j)}{\psi(\max\{L, i\})}\right)$, where $f_{\theta}:\R\to\R$ is a learnable MLP parameterized by $\theta$, $\psi: \sN \to \sR_{+}$ is 
    $\psi\left(x)\right)=log\left(cx+1\right)$
    and $c>0, L>0$ are learnable scalars.
\end{itemize}
Additional background on additive RPEs is provided in Appendix \ref{app:additive_rpe}

\paragraph{Rotary Positional Encoding (RoPE).}
RoPE \citep{su2024roformer}
encodes position information in attention logits through rotational encoding of query and key vectors based on their relative positions. Despite being simple and effective, RoPE exhibits limited length generalization  \citep{press2022train, kazemnejad2023impact}. While extensions like Position Interpolation \cite{chen2023extending, peng2023yarn, su2023rerope} enhance RoPE's context length, they do not necessarily improve length generalization on algorithmic tasks where learning the underlying algorithm is crucial.

\paragraph{No Positional Encoding (NoPE).}
While encoder-only Transformers (e.g., BERT \citep{devlin2018bert}) are permutation equivariant without positional encodings, decoder-only counterparts with causal attention, as shown by \citet{haviv2022transformer}, acquire positional understanding autonomously, even without explicit PE. Interestingly, recent findings by \citet{kazemnejad2023impact} further reveal that a model without PE outperforms those with specialized PEs on simple algorithmic tasks.

\paragraph{Randomized Position Encoding.} \citet{ruoss2023randomized} introduced Randomized PE to enhance existing PEs by randomly sampling encodings from a range exceeding test-time lengths while preserving the order. Transformers trained this way adapt to larger positional encodings, effectively eliminating OOD position encodings during testing.

\workshopexclude{
\begin{figure}[t]
\centering
\begin{subfigure}[t]{0.5\textwidth}
    \includegraphics[width=1.0\linewidth]{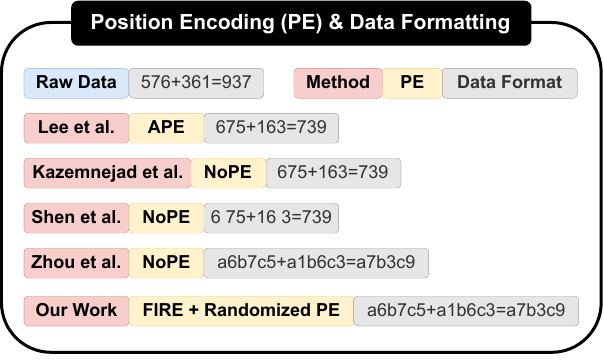}
\end{subfigure}
\caption{%
Comparative overview of PEs and data formats: While most related studies focus on APE or NoPE, our approach integrates FIRE \citep{li2023functional} and Randomized PE \citep{ruoss2023randomized}. All studies utilize a reversed format. \citet{shen2023positional} enhance this with random space augmentation, and both \citet{zhou2023algorithms} and Our Work incorporate index hints.
}\label{fig:pe_data_format}
\end{figure}
}

\subsection{Data Formats}\label{sec:background_data}
Data format plays a pivotal role in enhancing Transformers' length generalization capabilities, primarily by transforming the data into a format that could be more easily learned. 
We give an overview of the existing techniques below.

\paragraph{Reversed Format.} 
Computing addition in an algorithmic way (as taught in elementary school) requires starting with the least significant digit (LSD) and proceeds to the most significant digit (MSD). This sequence contrasts with the standard printed format ($A_3A_2A_1 + B_3B_2B_1 = C_3C_2C_1$, where $A_1$ and $B_1$ are the LSDs, which is not ideally suited for autoregressive models due to their outputting the MSD first. However, the reversed format ($A_1A_2A_3 + B_1B_2B_3 = C_1C_2C_3$) aligns better with these the natural order of computing the digits. It simplifies the learning task to a function that depends only on 
the two corresponding operand digits and the carry from the previous step \citep{lee2023teaching, zhou2023algorithms, shen2023positional}.

\paragraph{Index Hints.} 
\citet{zhou2023algorithms} introduced ``index hints'' in both the query and response of arithmetic tasks. For example, $42 + 39 = 81$ is represented as $a4b2 + a3b9 = a8b1$ during training and inference, enabling transformers to execute indexing via induction heads \citep{olsson2022context}.

\paragraph{Random Space Augmentation.} \citet{shen2023positional} explored the impact of random spacing between digits in addition, aiming to disrupt the model's reliance on absolute positional information. Their results show successful generalization from 10-digit to 11-digit addition, but falters with longer sequences.

\Cref{fig:pe_data_format} lists the position encodings and data formats used in some of the most related work to ours.

%% file: 3-method.tex
\section{A Recipe for Length Generalization in Decimal Addition}
\label{sec:recipe}

The task of decimal addition is composed of two critical subtasks: (a) the identification of the right operands to add;
and (b) the summation of these operands with the preceding carry.
While the summation step ((b)) is relatively easier because it has a finite set of possible inputs, the primary generalization challenge lies in the operand identification ((a)), where precise positional access is crucial. 

Our best model, which leads to the results in \Cref{fig:main_performance}, uses the following combination: 
\begin{enumerate}[topsep=2pt,itemsep=2pt, parsep=2pt, leftmargin=16pt]
    \item \textbf{FIRE position encodings} \citep{li2023functional}: 
    We believe that FIRE position encodings are helpful for length generalization because they are more expressive than other PEs, as shown by \citet{li2023functional}. 
    \item \textbf{Randomized position encodings} \citep{ruoss2023randomized}: We believe that randomized position encodings are crucial to avoid overfitting on the position indices and index differences that were seen during training. 
    \item \textbf{Reversed format}: The reversed format makes it easier for the model to \emph{decompose} the long computation to local, ``markovian'', steps that depend only on the single previous step.
    \item \textbf{Index hints} \citep{zhou2023algorithms}: We believe that index hints are useful 
    because they ease the task of \emph{operand identification} (discussed in (b)), of matching the right operands to add at a certain step.

\end{enumerate} 

We ablate each of these decisions and some other alternative choices in \Cref{sec:experiments}.

%% file: 4-result.tex
\section{Experiments}
\label{sec:experiments}

\workshopexclude{
\subsection{Setup}
\paragraph{Data.} As shown in \Cref{fig:pe_data_format}, we adopt the reversed format with index hints as our default data format. During training, we randomly sample consecutive index hints from a pre-defined ordered set of hints with 102 symbols, thereby enhancing the learning of hint sequences and their order. 
We generated a dataset comprising 30M examples on input lengths 1-40 for training and 1,000 examples per input length for testing. 

\paragraph{Model.} Our base model, following \citet{zhou2023algorithms}, is a 25M parameter Transformer featuring 6 blocks, a 512 hidden size, and a feedforward layer with a hidden dimension of 2048. We also adopt RMSNorm, integrating both PreNorm and PostNorm layers, following the Primer architecture \citep{so2021primer}.
We use the AdamW optimizer \citep{loshchilov2017decoupled} to train the model with a weight decay value of 0.1 and no dropout, for 50,000 steps. The learning rate schedule incorporates an initial 500-step linear warm-up, followed by a cosine decay, starting at 3e-4. The hyperparameters are chosen based on \Cref{sec:hyperparam}.

\paragraph{Randomized PE and Random Space Augmentation.} As will be demonstrated in \Cref{fig:rs_noisy:a,fig:rs_noisy:b}, the success of these techniques is markedly PE-dependent. Hence, we tailor the default hyperparameter choice to best suit each PE.
Further, instead of using random spaces, we use another special token to prevent automatic merging by the tokenizer.

Due to the high variance (which we discuss in the next section), we repeat each experiment five times unless mentioned otherwise.
More implementation details are provided in \Cref{app:imp_details}.

\input{figures/length_generalization}

\subsection{Results}
\label{subsec:main_results}
}
\paragraph{FIRE enables significantly better length generalization.} 
\Cref{fig:length_generalization:a} compares the length generalization capabilities of four positional encodings in the best of 10 trials (See \Cref{sec:loss_acc_reverse_index_hint} for all trials). Trained exclusively on sequences of lengths 1-40, the best trial of FIRE exhibit near-perfect generalization to sequences up to the length of 100. In contrast, other PEs show a visible degradation in generalization accuracy beyond the sequence length of 60. This finding counters the findings of \citet{kazemnejad2023impact} that no positional encoding (NoPE) surpasses complex PE techniques for length generalization. Our findings suggest that a well-designed PE, such as FIRE, is essential for optimal length generalization.

\paragraph{Index hints are crucial.} We compare models trained with and without index hints. 
As shown in \Cref{fig:em_index_hint:a}, index hints significantly enhance length generalization across various PEs, corroborating the findings of \citet{zhou2023algorithms}. Notably, without index hints, NoPE and FIRE demonstrate poor in-distribution generalization for 40-digit additions, a marked deviation from their reasonable performance when trained on 10-digits, as shown in \Cref{fig_app:em_no_index_hint:a}. 
\Cref{fig_app:seed_acc_tlen40_nopos} shows that this phenomenon occurs across all random seeds.  
Conversely, RoPE and KerpleLog exhibit moderate in-distribution generalization but falter in out-of-distribution scenarios. \Cref{sec:loss_acc_reverse_wo_index_hint,sec:loss_acc_reverse_wo_index_hint_10} shows the training loss and test accuracy of these runs.

Analyzing errors in 11-digit additions from models trained on 10-digits revealed a common misalignment issue: the Transformer often adds operands adjacent to the correct ones. An attempt to rectify this by reformatting addition ($A_1B_1, A_2B_2, A_3B_3 = C_1C_2C_3$, with 1 as the least significant bit) failed to improve length generalization, merely shifting the error to adjacent output positions. This highlights the Transformer's inherent limitations in precise position identification.

\input{figures/index_hint}

\input{figures/rs_noisy}

\paragraph{Standard format vs reversed format.} As shown in Figure~\ref{fig:em_index_hint:b}, standard formatting shows limited length generalization in all PEs compared to the reversed format.
FIRE excels in length generalization even with the standard format, even matching RoPE in reverse format. However, FIRE's performance (with standard format) declines beyond 60-digit additions, likely due to increased carry propagation challenges exceeding the model's capacity. 

Looking at the training loss and training next-token accuracy in both formats also shows interesting differences. As shown in \Cref{fig:reverse_loss,fig_app:logpplx_acc_step_4pe}, the standard format training leads to gradual improvement, whereas reverse format yields a sharp performance transition. This transition, 
which is a reminiscent of ``grokking'' phenomenon  \cite{power2022grokking},
shows in this case the ``Eureka moment'' in which the Transformer learns the right addition algorithm.

\input{figures/robustness_figs}

\paragraph{Random space augmentation and randomized position encoding.}
\Cref{fig:rs_noisy:a} reveals divergent impacts of random space augmentation on four PEs. 
The augmentation's efficacy is notably contingent upon the chosen PE. While Random Spaces marginally enhances RoPE and KerpleLog's performance, it markedly deteriorates NoPE and FIRE. A similar PE-specific pattern is evident in Randomized PE, as \Cref{fig:rs_noisy:b} demonstrates. Randomized PE significantly degrades KerpleLog's effectiveness, yet it substantially boosts FIRE. See \Cref{sec:loss_acc_reverse_index_hint_rs,sec:loss_acc_reverse_index_hint_randomized_pe} for training loss and EM accuracy for all trials in each setting.

\paragraph{Length generalization is not robust to neither weight initialization nor training data order.} \Cref{fig:fire_all_seeds} illustrates the varying performance of 10 FIRE trials using identical training data order but distinct weight initializations. Notably, while all trials achieve similar close-to-zero training losses after 10K training steps (\Cref{fig_app:logpplx_logx_all}) and exhibit perfect in-distribution generalization, their out-of-distribution (OOD) length generalization shows significant variance. 
Moreover, the length generalization performance fluctuates significantly across training steps (\Cref{sec:acc_vs_steps_reverse_index_hint}). This observation contrasts with earlier studies suggesting in-distribution loss as a reliable OOD generalization predictor \citep{nagarajan2020understanding}. 

We further examine 15 unique combinations, resulting from 3 weight initialization seeds and 5 data input orders. As shown in \Cref{fig:order_digit_len:a}, there is significant variance across training data orders
even when the weight initialization is constant.
Intriguingly, certain weight initializations demonstrate remarkable resilience to changes in data input order. This observation is reminiscent of the Lottery Ticket Hypothesis~\citep{frankle2018lottery}, which posits the existence of a sparse, equally effective sub-network within a larger neural network. Our findings suggest the presence of ``fortunate'' weight configurations that exhibit robust length generalization, akin to a ``lucky weight ticket.''

\workshopexclude{
While \citet{anil2022exploring} also noticed similar in-distribution accuracy but marked differences in OOD behavior on parity tasks, their OOD performance was quite poor across all runs. Moreover, contrary to the findings of \citet{anil2022exploring} on the impact of hyperparameter variation, our experiments reveal considerable performance fluctuations even with different random seeds. This inconsistency appears unrelated to position encoding (refer to \Cref{fig_app:seed_acc_tlen40} for different PEs), and is more likely due to variations in random weight initialization and data order. %
}

\workshopexclude{
\section{Analysis}\label{sec:analysis}

\paragraph{Error analysis.}
In examining Transformers' error characteristics, we classified erroneous predictions into two categories: those with and without carry. Figure~\ref{fig:rs_noisy:c} shows no significant difference between these categories, thus carry propagation does not majorly impede length generalization.

Additionally, we analyzed the error distribution in 100-digit addition using FIRE, illustrated in \Cref{fig_app:error_analysis}. As shown, \Cref{fig_app:error_analysis} indicates an overall uniform error distribution across all indices, despite some individual model checkpoints showing errors at specific positions. Excluding two near-zero accuracy runs, over 90\% of errors in incorrect examples are single-digit mistakes, following an exponential distribution. Additional results are shown in \Cref{fig_app:error_count_hist,fig_app:error_pos_hist}.

\input{figures/order_digit_len}

\input{figures/scaling}

\input{figures/hyperparmeter_sensitivity}

Despite the imperfect calculation, the FIRE model does not show any systematic error. 
Random errors may stem from phenomena such as attention glitches \cite{liu2023exposing}. Conversely, other PEs systematically fail to identify the start or end of addition, leading to premature termination.

\paragraph{Performance evolution during training.} \Cref{fig:in_vs_ood} shows that while transformers achieve near-perfect in-distribution accuracy early in training, they explore different extrapolation strategies. This ability is remarkable considering the inherent unpredictability and architecture-dependent nature of OOD accuracy. Notably, transformers with FIRE exhibit a generally steady increase in OOD accuracy during training, suggesting that FIRE's inductive bias may be helpful in finding solutions that generalize to different lengths. In contrast, other PE methods display more volatile OOD performance. Interestingly, some methods exhibit a ``grokking-like'' phenomenon, where there is a sudden surge in the OOD accuracy despite no change in in-distribution accuracy.

\paragraph{Sequence length during training.}  
We trained separate models for addition involving up to 10, 20, 30, and 40 digits, and evaluated them on addition of up to 100 digits. As depicted in \Cref{fig:order_digit_len:b,fig_app:scale_4pe}, training length crucially improves performance in longer length generalizations across different PEs. 
Notably, not only that models that were trained on 40 digits generalize better than models that were trained on shorter sequences, the \emph{generalization factor is also increasing}: the model that was trained on 40 digits generalizes to 100 digits (2.5$\times$), while the model that was trained on up to 30 digits generalizes to 45 digits (1.5$\times$), the model that was trained on up to 20 digits generalizes to 25 digits (1.25$\times$), and the model that was trained on up to 10 digits does not generalize beyond training lengths (1.0$\times$).

\paragraph{Scaling model size.} 
The scaling of model size is crucial for improving large language models \citep{thoppilan2022lamda, chowdhery2023palm}. To assess its effect on length generalization, we contrasted models with 25M and 268M parameters. We find that model size variation has a minor effect on length generalization. \Cref{fig:scaling} shows that larger models slightly improve generalization in short digit regimes (1 to 10 and 1 to 20 digit additions) but yield mixed results in longer regimes. While RoPE and KerpleLog show improvements, NoPE and FIRE experience performance degradation with a larger model, indicating model size may not be the primary factor in length generalization.

The efficacy of length generalization in the 25M model prompted us to explore the capabilities of smaller models. 
Specifically, we trained models with 2M and 5M parameters. As \cref{fig:scaling:b,fig_app:scale_fire} illustrate, the 2M model's performance deteriorates with longer sequences, indicating limited model capacity as a potential performance bottleneck. Intriguingly, this model outperforms its larger counterparts (5M and 25M models) in tasks involving 1 to 10 digit addition. Furthermore, the 5M model remarkably achieves 80\% accuracy in 100 digit addition, trained only on 1 to 40 digit tasks, surpassing the 268M model's performance. 

\paragraph{Does stronger regularization reduce variance?} To mitigate performance variance, we investigated standard regularization techniques, including weight decay and dropout. As depicted in \Cref{fig:wd_dropout:a}, higher weight decay values (e.g., 0.1, 0.3) slightly enhance the likelihood of achieving effective length generalization. Nonetheless, non-trivial length generalization remains attainable with either very low (e.g., 1e-6) or high (e.g., 1.0) weight decay values, evidenced by approximately 80\% accuracy in 100 digit addition trained on 40-digit sequences. Conversely, \Cref{fig_app:wd_dropout} shows that substantial dropout values (e.g., 0.2) severely impair length generalization. Dropout rates of 0.0 or 0.1, however, do not show statistically significant improvements over their counterparts. Overall, while regularization can modestly decrease performance variability, it falls short in ensuring robust length generalization. The variance in performance is still significantly influenced by the randomness of weights initialization and the training data order (\Cref{fig:fire_all_seeds,fig:order_digit_len:a}).
}

%% file: figures/length_generalization.tex
\begin{figure*}[t]
\centering
\begin{minipage}{0.48\textwidth}
\centering
\begin{subfigure}[t]{1.0\columnwidth}
\includegraphics[width=1.0\columnwidth]{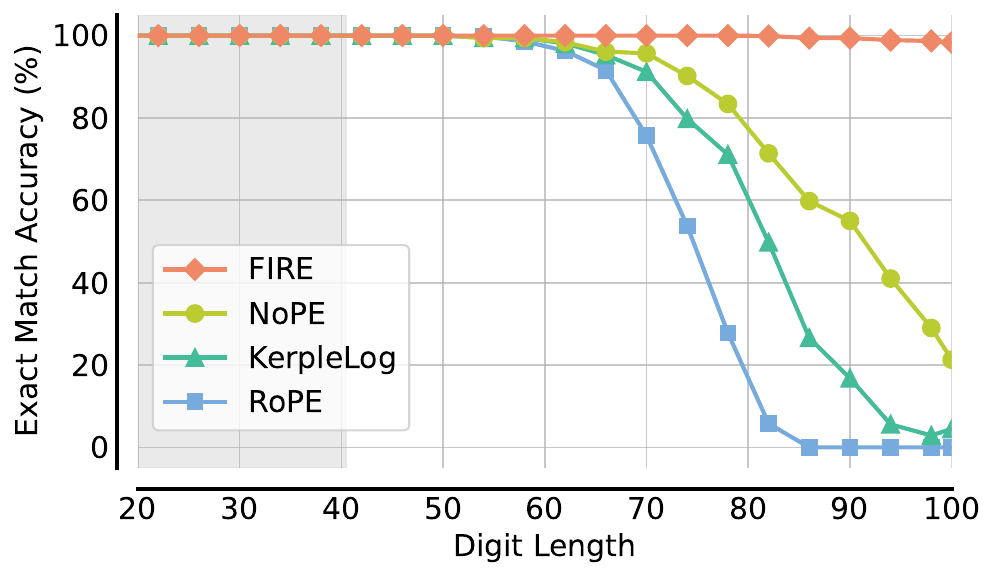}
\end{subfigure}
\caption{%
EM accuracy (best of 10 trials), trained exclusively on sequences of lengths 1 to 40, the best trials involving FIRE exhibit near-perfect generalization on 100-digit addition.}
\label{fig:length_generalization:a}
\end{minipage}
\hfill
\begin{minipage}{0.48\textwidth}
\centering
\begin{subfigure}[t]{1.0\columnwidth}
\includegraphics[width=1.0\columnwidth]{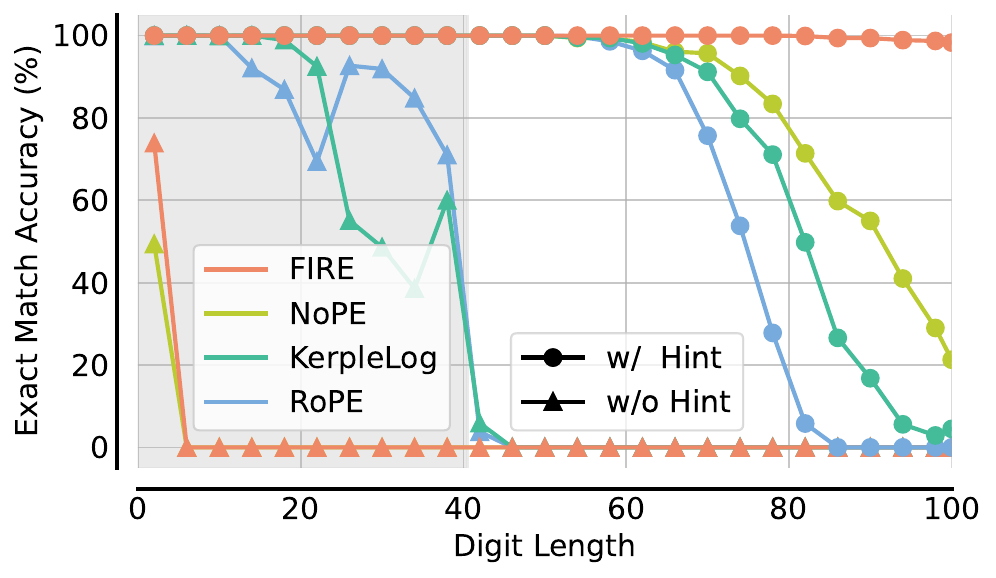}
\end{subfigure}
\caption{%
EM accuracy of models trained with and without index hints (best of 10 trials): Without index hints, all PE methods fail in generalization, both within and beyond trained lengths.}
\label{fig:em_index_hint:a}
\end{minipage}
\vspace{-2mm}
\end{figure*}

%% file: figures/index_hint.tex
\begin{figure*}[t]
\centering
\begin{minipage}{0.48\textwidth}
\centering
\includegraphics[width=1.0\columnwidth]{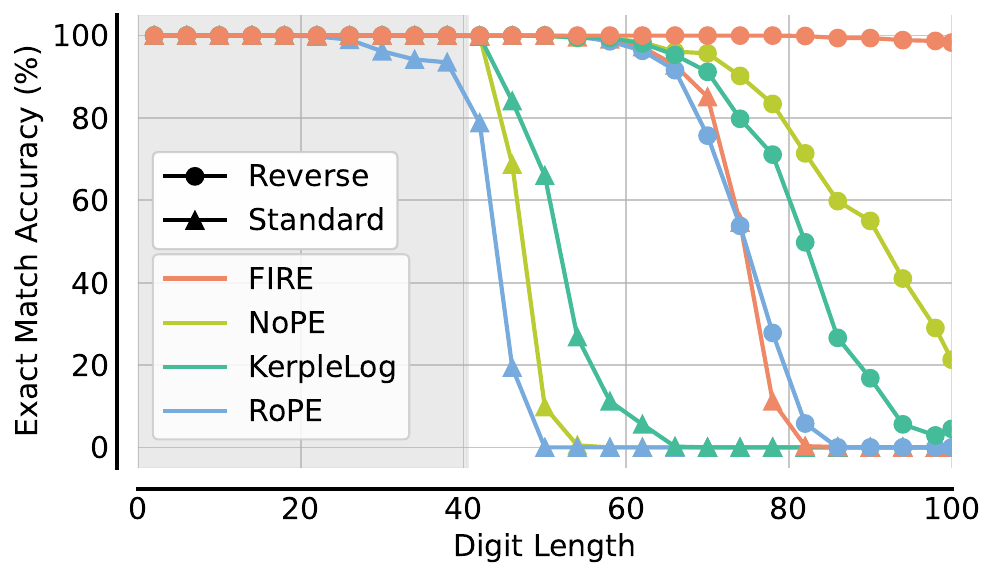}
\caption{%
EM accuracy of the standard vs. the reversed format: Consistently with prior studies, the reversed format excels over the standard format across all PEs. 
}
\label{fig:em_index_hint:b}
\end{minipage}
\hfill
\begin{minipage}{0.48\textwidth}
\centering
\includegraphics[width=1\columnwidth]{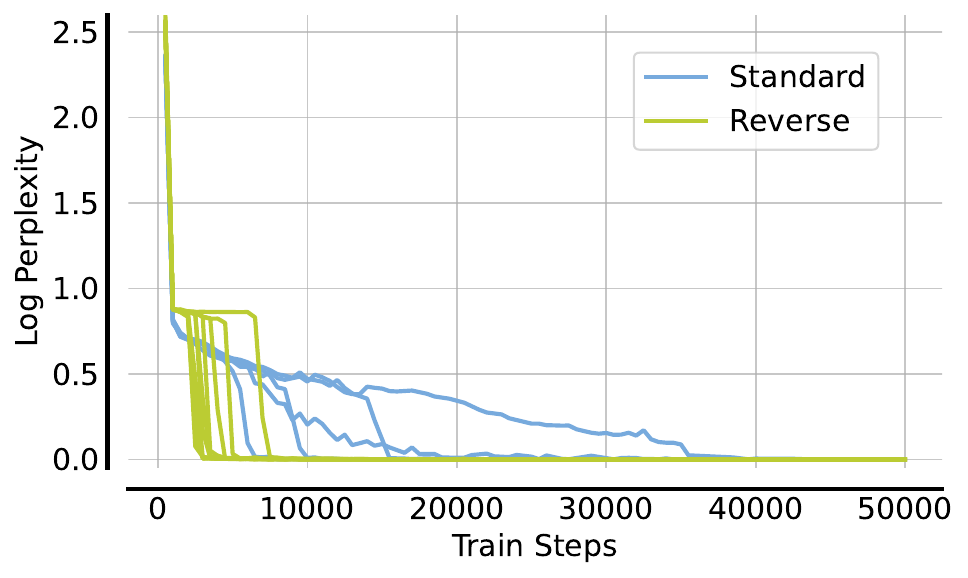}
\caption{%
The reversed format shows distinct grokking during training, unlike the gradual enhancement in the standard format. This phenomenon is observed across all PEs (\Cref{fig_app:logpplx_acc_step_4pe})
}
\label{fig:reverse_loss}
\end{minipage}
\end{figure*}

%% file: figures/rs_noisy.tex
\begin{figure*}[t]
\centering
\begin{minipage}[t]{0.45\textwidth}
\includegraphics[width=1\columnwidth]{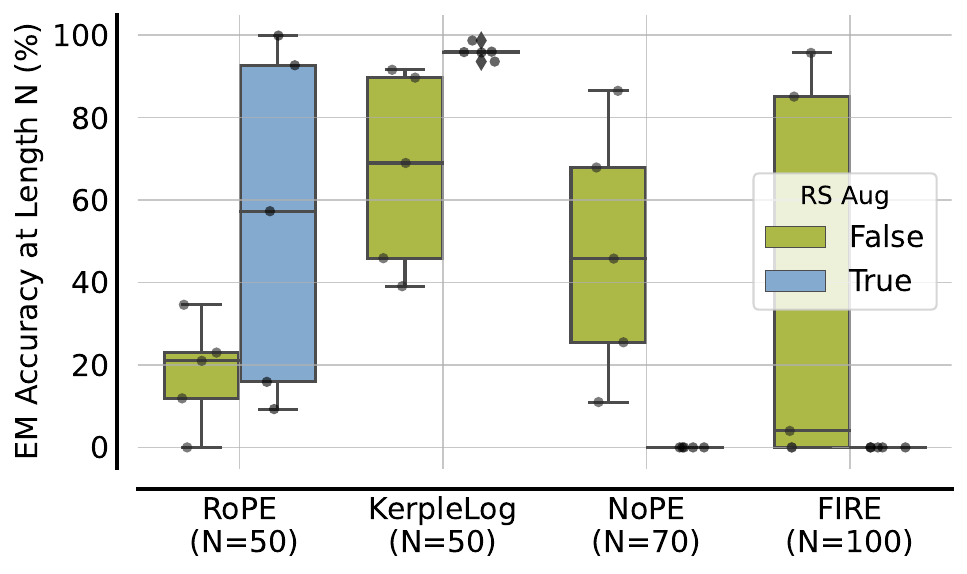}
\caption{%
Effects of Random Space Augmentation (RS Aug): Random space augmentation is beneficial for RoPE and KerpleLog; adverse for NoPE and FIRE.}
\label{fig:rs_noisy:a}
\end{minipage}
\hfill
\begin{minipage}[t]{0.25\textwidth}
\includegraphics[width=1\columnwidth]{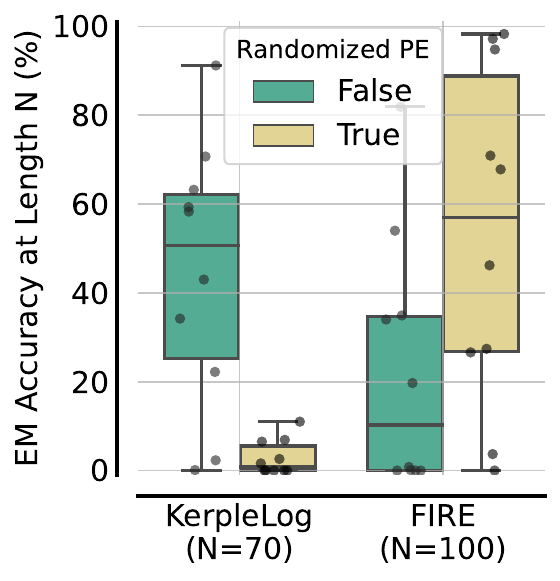}
\caption{%
Effects of Randomized PE: 
Randomized PE enhances FIRE but degrades KerpleLog}
\label{fig:rs_noisy:b}
\end{minipage}
\hfill
\begin{minipage}[t]{0.23\textwidth}
\includegraphics[width=1\columnwidth]{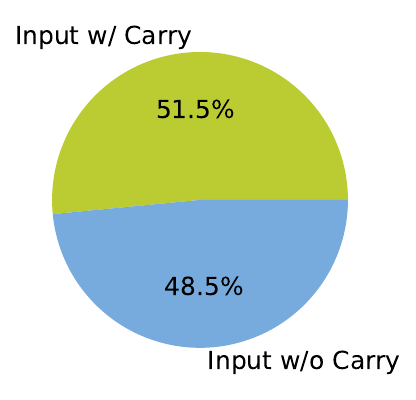}
\caption{%
Error Distribution: Errors appear almost equally with and without carry.}
\label{fig:rs_noisy:c}
\end{minipage}
\vspace{-2mm}
\end{figure*}

%% file: figures/robustness_figs.tex
\begin{figure*}[t]
\centering
\begin{minipage}{0.48\textwidth}
\centering
\includegraphics[width=1.0\columnwidth]{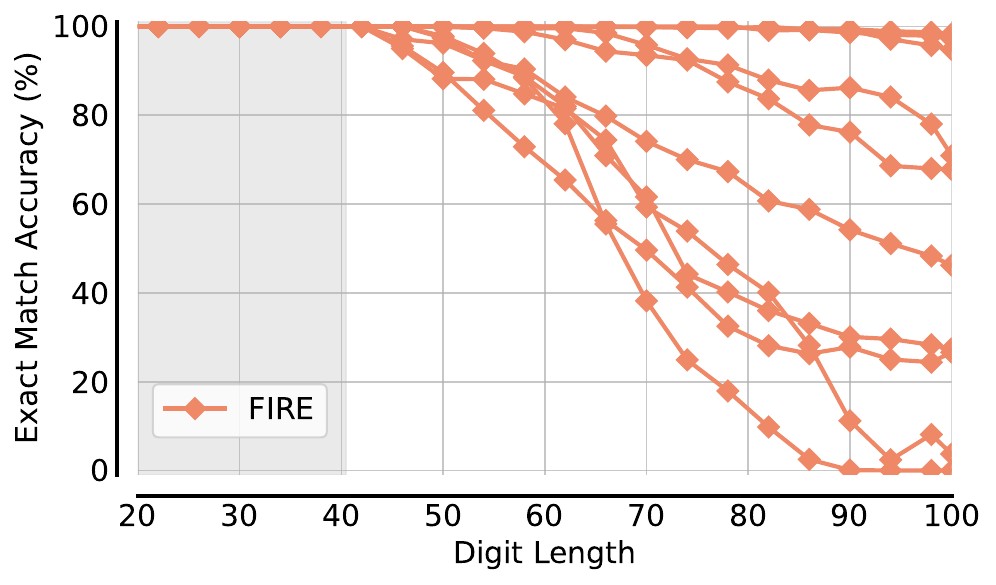}
\caption{
Exact match across 10 trials using FIRE. While transformers can achieve near-perfect accuracy in 100-digit addition, the variance across different random seeds is high. 
}
\label{fig:fire_all_seeds}
\end{minipage}
\hfill
\begin{minipage}{0.48\textwidth}
\centering
\includegraphics[width=1\columnwidth]{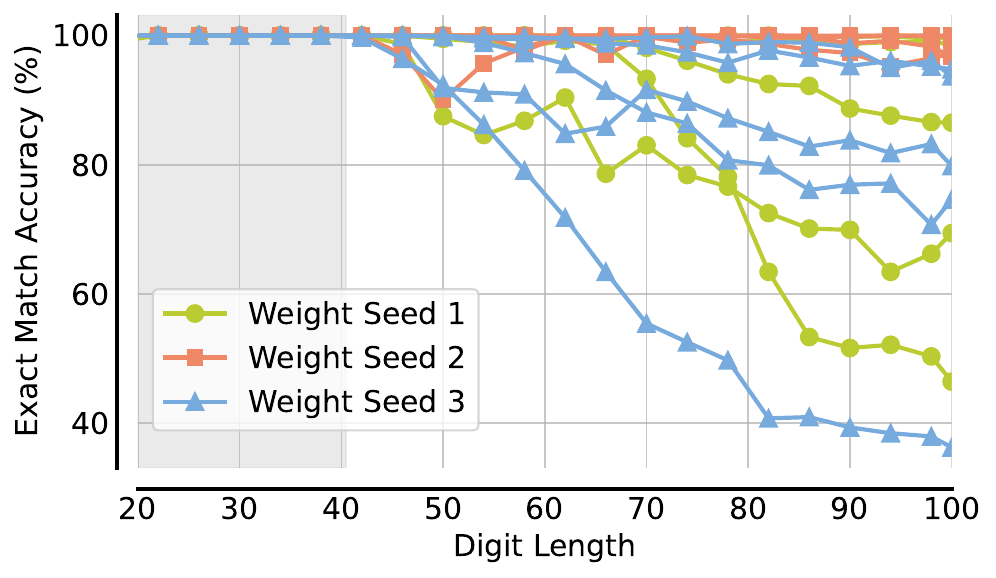}
\caption{
Effects of weight initialization and data input order:
15 models trained on a combination of three weight initialization seeds and five data input order seeds.
}
\label{fig:order_digit_len:a}
\end{minipage}
\end{figure*}

%% file: figures/order_digit_len.tex
\begin{figure*}[t]
\centering
\begin{minipage}{0.48\textwidth}
\centering
\includegraphics[width=1.0\columnwidth]{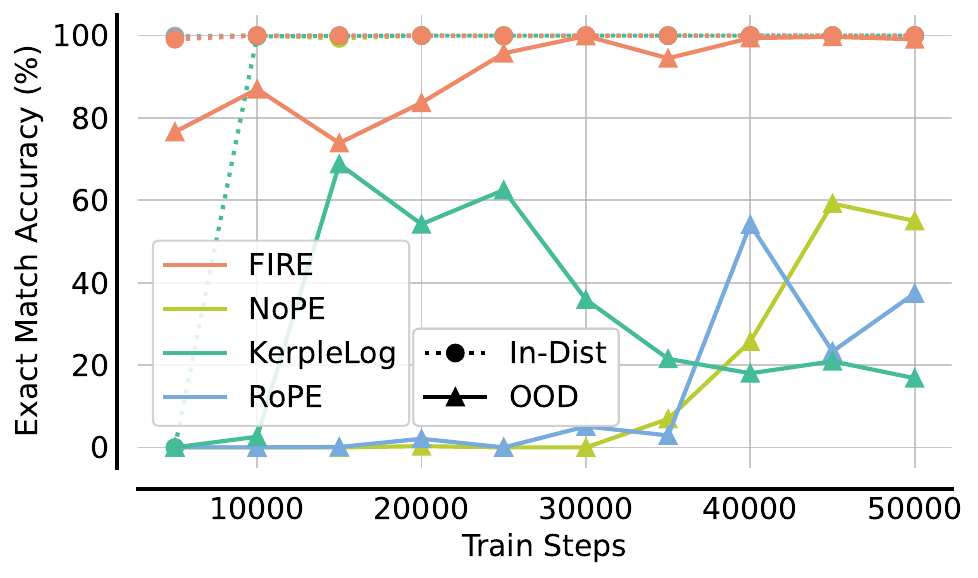}
\caption{
Comparison of In-Distribution (30-digit addition) and Out-of-Distribution Generalization (90-digit addition, except for RoPE at 70-digit addition).
}
\label{fig:in_vs_ood}
\end{minipage}
\hfill
\begin{minipage}{0.48\textwidth}
\includegraphics[width=1.0\columnwidth]{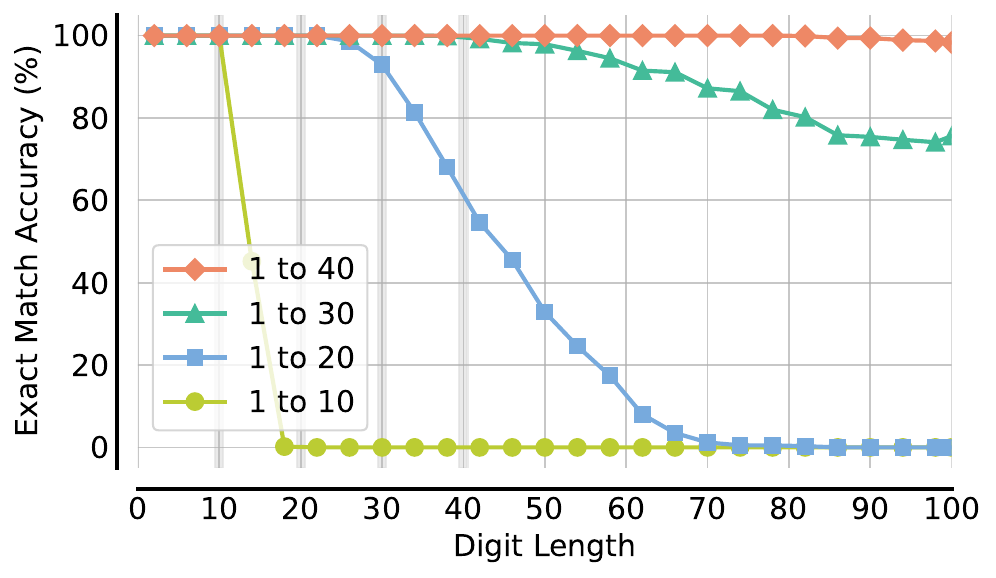}
\caption{
Different training lengths: Increasing the training length significantly improves length generalization in FIRE, achieving near-perfect accuracy at length 100.
}
\label{fig:order_digit_len:b}
\end{minipage}
\end{figure*}

%% file: figures/scaling.tex
\workshopexclude{
\begin{figure*}[t]
\centering
\begin{subfigure}[t]{0.32\textwidth}
\centering
\captionsetup{justification=centering}
\includegraphics[width=1.0\columnwidth]{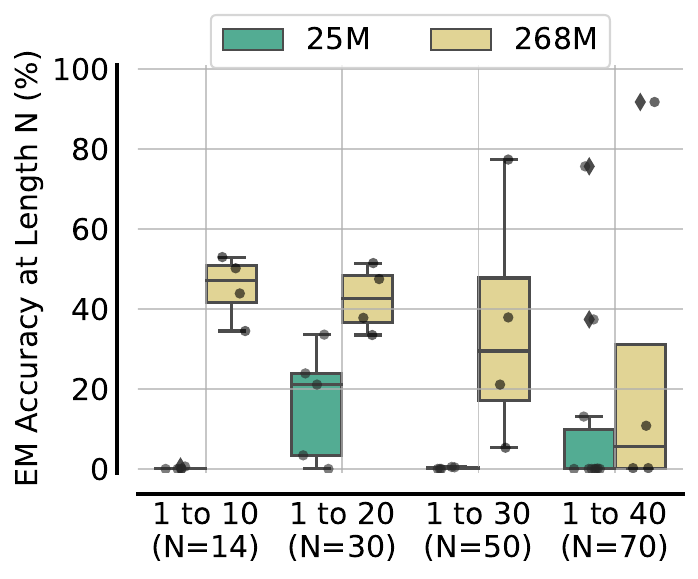}
\caption{RoPE}
\end{subfigure}
\begin{subfigure}[t]{0.32\textwidth}
\centering
\captionsetup{justification=centering}
\includegraphics[width=1.0\columnwidth]{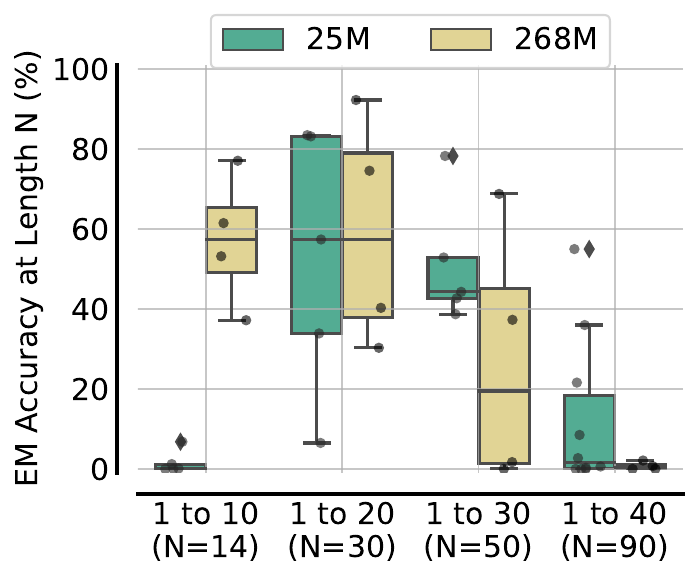}
\caption{NoPE}
\end{subfigure}
\begin{subfigure}[t]{0.32\textwidth}
\centering
\captionsetup{justification=centering}
\includegraphics[width=1.0\columnwidth]{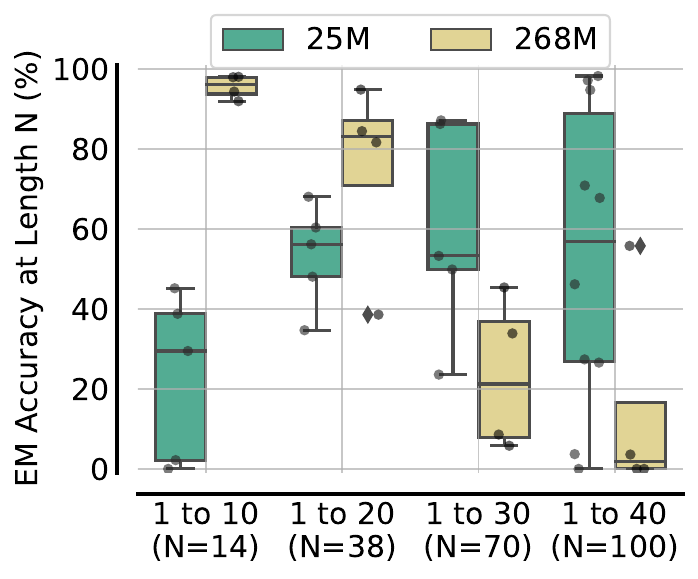}
\caption{FIRE}
\end{subfigure}
\caption{%
Scaling model size inconsistently affects length generalization performance. 
While consistently enhancing performance in shorter length regimes (1-10, 1-20) across four PEs, this trend does not hold for larger regimes (1-30, 1-40). For instance, larger models outperform smaller ones with RoPE and KerpleLog (\Cref{fig_app:scaling}), but underperform with NoPE and FIRE. Moreover, increasing model size doesn't noticeably decrease performance variance, suggesting size scaling isn't vital for length generalization.
}\label{fig:scaling}
\end{figure*}
}

\workshoponly{
\begin{figure*}[t]
\centering
\begin{subfigure}[t]{0.32\textwidth}
\centering
\includegraphics[width=1.0\columnwidth]{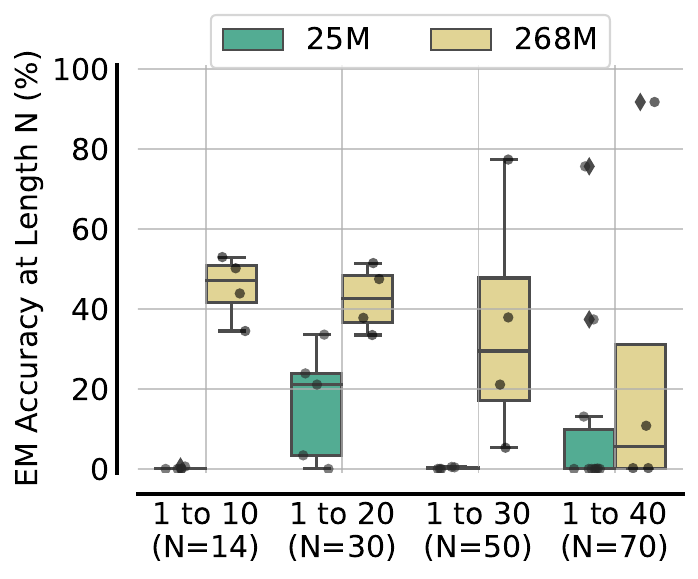}
\caption{RoPE}
\end{subfigure}
\begin{subfigure}[t]{0.32\textwidth}
\centering
\includegraphics[width=1.0\columnwidth]{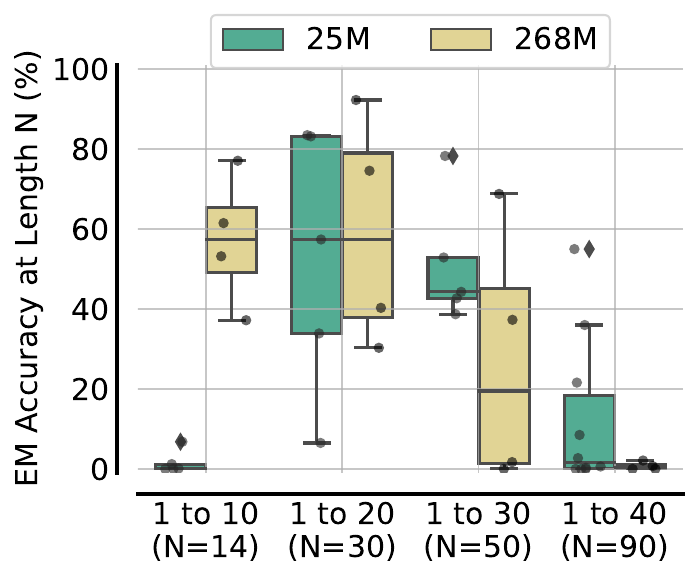}
\caption{NoPE}
\end{subfigure}
\begin{subfigure}[t]{0.32\textwidth}
\centering
\includegraphics[width=1.0\columnwidth]{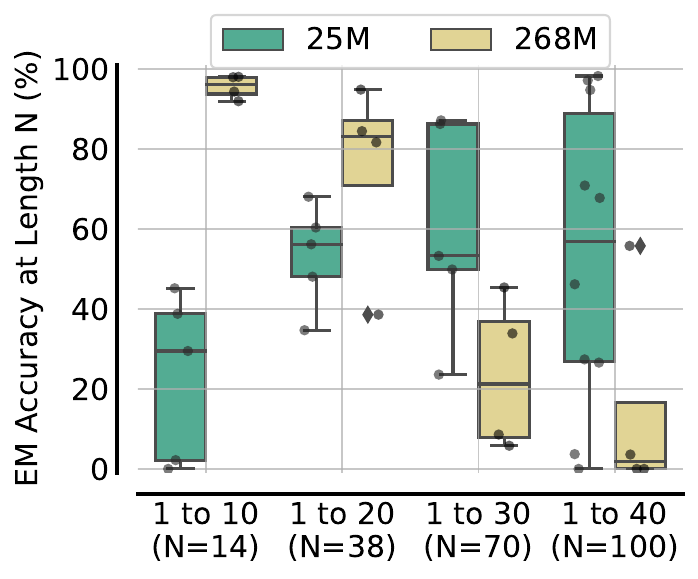}
\caption{FIRE}
\end{subfigure}
\caption{%
Scaling model size inconsistently affects length generalization performance. 
While consistently enhancing performance in shorter length regimes (1-10, 1-20) across four PEs, this trend does not hold for larger regimes (1-30, 1-40). For instance, larger models outperform smaller ones with RoPE and KerpleLog (\Cref{fig_app:scaling}), but underperform with NoPE and FIRE. Moreover, increasing model size doesn't noticeably decrease performance variance, suggesting size scaling isn't vital for length generalization.
}\label{fig:scaling}
\end{figure*}
}

%% file: figures/hyperparmeter_sensitivity.tex
\begin{figure*}[t]
\centering
\begin{minipage}{0.61\textwidth}
\centering
\includegraphics[width=1\columnwidth]
{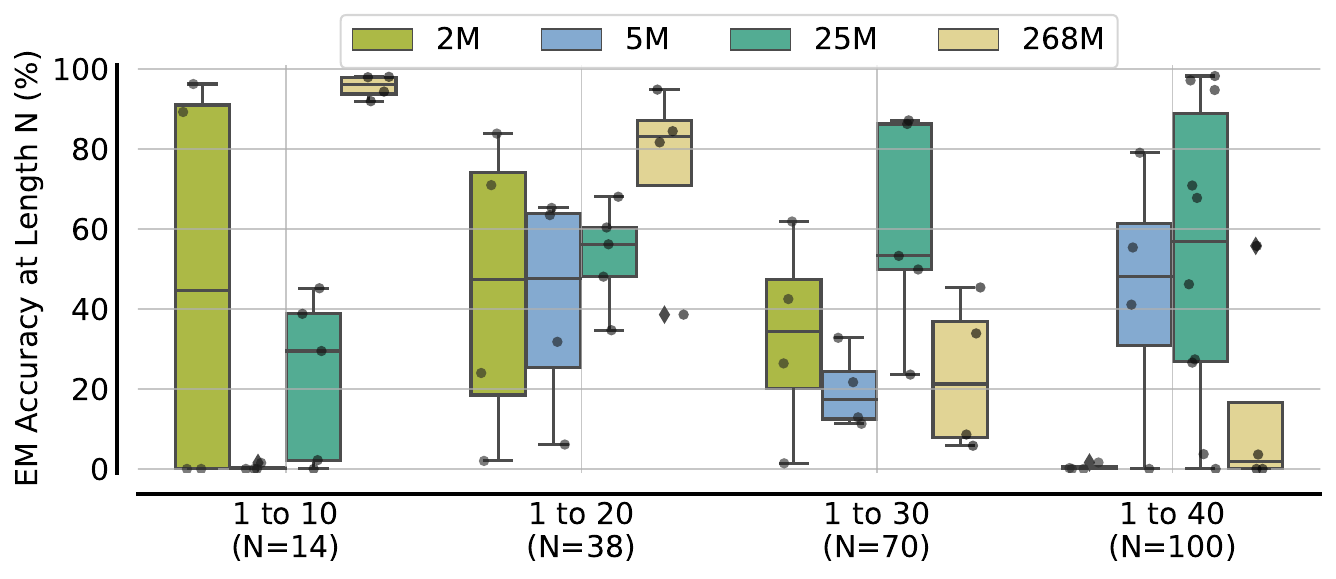}
\caption{Effect of different model sizes with FIRE  as the position encoding.}
\label{fig:scaling:b}
\end{minipage}
\hfill
\begin{minipage}{0.36\textwidth}
\includegraphics[width=1\columnwidth]{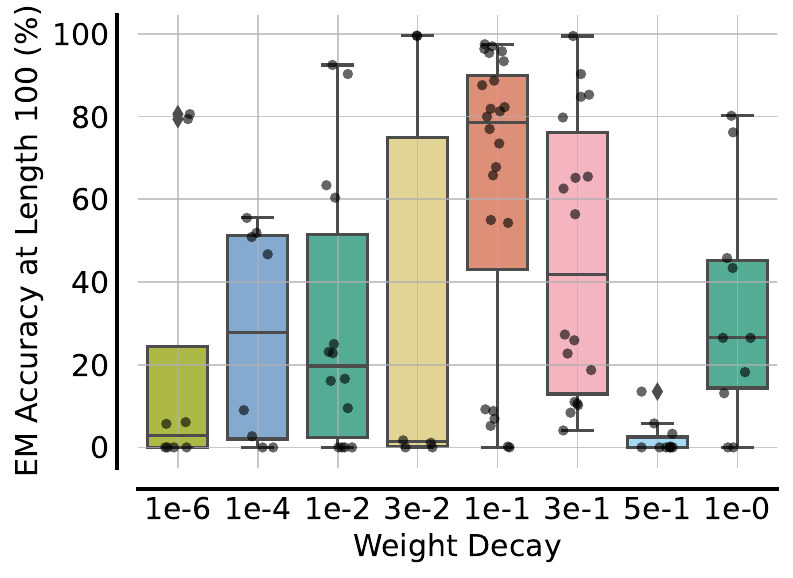}
\caption{Effect of weight decay with FIRE as the position encoding.
}
\label{fig:wd_dropout:a}
\end{minipage}
\end{figure*}

%% file: 5-related_work.tex
\section{Related Work}

Length generalization remains a significant challenge in neural networks, underscored by substantial research \citep{graves2016hybrid, hupkes2020compositionality, schwarzschild2021can, zhang2022unveiling, deletang2023neural, dziri2023faith}. Despite their advanced reasoning capabilities, Transformer-based large language models (LLMs) \citep{thoppilan2022lamda, chowdhery2023palm} struggle with processing sequences beyond their training scope \cite{anil2022exploring}. Enhancements in length generalization, especially in the addition task, primarily focus on two areas: refining positional encoding and optimizing data format.

\paragraph{Position Encoding for Length Generalization}
The inability of Transformers to extrapolate to longer sequences has been primarily attributed to Position Encoding (PE) \cite{shaw2018self}. Various studies have suggested alternatives, such as relative positional encodings, which focus on the relative distances between tokens \citep{dai2019transformer}, the implementation of randomized position encoding \citep{ruoss2023randomized}, or the adoption of weighted attention mechanisms in place of position embeddings \citep{press2022train, raffel2020exploring, chi2022kerple, li2023functional}. These approaches have shown promise in natural language processing (NLP). However, \citet{kazemnejad2023impact} found that omitting position encoding entirely yields better results for algorithmic tasks. In contrast, our experiments indicate that an effectively designed PE, such as the FIRE, is crucial for achieving optimal length generalization (\Cref{fig:length_generalization:a}). Moreover, we show that a synergistic approach to consider both PE and data design markedly enhances length generalization capabilities.

\paragraph{Data format for Length Generalization}
A range of heuristic-based data formatting methods have been introduced, particularly for pretrained LLMs. These methods, including the use of scratchpads and the chain of thoughts approach, aim to facilitate arithmetic learning either through in-context learning or fine-tuning \cite{anil2022exploring, zhou2022teaching}. Conversely, there is a body of research focused on Transformers trained from scratch. 
This research indicates that employing techniques such as reversed formatting and scratch pads can significantly boost length generalization performance \cite{shen2023positional, lee2023teaching}. 
Furthermore, it has been observed that both the data distribution and the sampling strategies can profoundly influence generalization \cite{lee2023teaching}. 
\citet{awasthi2023improving} further demonstrates the benefits of incorporating a simpler auxiliary task (e.g., identifying the successor element) in supporting the primary task (e.g., sorting). In contrast, \citet{jelassi2023length} finds that train set priming enables length generalization for a encoder-only Transformer model. 
In contrast, our good length generalization performance achieved with naive random sampling approach suggesting that sophisticated data sampling might be redundant.

%% file: appendix/positional_encoding.tex
\section{Positional Encoding}
\subsection{Additive Relative Positional Encoding (RPE)}\label{app:additive_rpe}
\citet{shaw2018self} pioneered additive RPE by integrating position encodings into the attention layer's key, and optionally the value, rather than the input. This concept was further simplified in T5~\citep{raffel2020exploring}, where the vector representations of relative positions are simplified to scalar biases added to pre-softmax attention logits. Subsequent advancements in additive RPE, aimed at enhancing length generalization and computational efficiency, include notable methods like Alibi \citep{press2022train}, Kerple \citep{chi2022kerple}, and FIRE \citep{li2023functional}. A commonality among these methods is the unified computation formula for pre-softmax attention logits, as outlined by \citet{li2023functional}:
\begin{equation}
    \label{eq:rpe-attn-mat}
    \mA_{\mathrm{RPE}}(\mX) = \mX \mW_Q(\mX \mW_K)^{\top}+\mB,
\end{equation}
where the bias matrix $\mB\in\R^{n\times n}$ is induced by the \textbf{position encoding function} $b: \sN^{*2}\to\R$, has its $(i,j)$-th entry defined as $b(i,j)$.
Variations in $b$'s formulations and parameterizations give rise to diverse RPE variants.

\begin{itemize}
    \item T5~\citep{raffel2020exploring}: T5's RPE segments relative distances into distinct buckets with a logarithmic scale, each associated with a unique parameter. With $K+1$ buckets and a pre-defined distance $L_1$, the attention bias is calculated as (assuming $K+1$ is even)
    \begin{equation}\label{eq:logbinning-t5-rpe}
    b(i,j)=\begin{cases}
        r_{i-j} & 0\leq i-j <  \frac{K+1}{2} \\
        r_{\frac{K+1}{2} +  \lfloor\frac{K+1}{2} \log\left(\frac{2(i-j)}{K+1}\right) / \log\left(\frac{2L_1}{K+1}\right) \rfloor}    & \frac{K+1}{2}\leq i-j< L_1 \\
        r_K & i-j\geq L_1
        \end{cases}.
    \end{equation}
    \item Alibi~\citep{press2022train}: $b(i,j) = -r|i-j|$, where $r>0$ is a hyper-parameter.
    \item Kerple~\citep{chi2022kerple}: $b(i,j) = -r_1\log(1+r_2|i-j|)$ (logarithmic variant) or $-r_1|i-j|^{r_2}$ (power variant), where $r_1, r_2>0$ are learnable scalars.
    \item FIRE~\citep{li2023functional}: $b(i,j) = f_{\theta}\left(\frac{\psi(i-j)}{\psi(\max\{L, i\})}\right)$, where $f_{\theta}:\R\to\R$ is a learnable MLP parameterized by $\theta$, $\psi: \sN \to \sR_{+}$ is monotonically increasing and $L>0$ is a learnable scalar.
\end{itemize}

%% file: appendix/implementation_details.tex
\clearpage
\newpage
\section{Implementation Details}\label{app:imp_details}

\subsection{Data Generation}\label{app:data_generation}
As shown in \Cref{fig:pe_data_format}, we adopt the reversed format with index hints as our default data format. During training, we randomly sample a consecutive index hints from a pre-defined ordered index set with 102 distinct symbols, thereby enhancing the learning of hint sequences and their order. At inference, the same hint sampling strategy is applied to questions, prompting the model for answers. 

To generate addition examples, we opt for a naive random sampling approach instead of structured data sampling \cite{lee2023teaching}, as our analysis indicates that carry operations are not a major hindrance to length generalization (See \Cref{fig:rs_noisy:c}). Our approach involves uniformly selecting the number's length from 1 to the maximum training length, followed by independent sampling of two operands based on this length, with an additional zero padding to accommodate potential carry-induced extra digits. For training, datasets comprising 30M, 40M, 60M, and 120M examples are generated for number lengths 1-40, 1-30, 1-20, and 1-10, respectively. In contrast, the test set consists of 1,000 examples per digit length.

\subsection{Training Details}\label{app:training}
Our base model, following \citet{zhou2023algorithms}, is a 25M parameter Transformer featuring 6 blocks, a 512 hidden size, a feedforward layer with a hidden dimension of 2048 using GeGLU activation \citep{shazeer2020glu}, and an 8-head attention mechanism. We also adopt RMSNorm, integrating both PreNorm and PostNorm layers, following the Primer architecture \citep{so2021primer}. Additionally, our preliminary investigations underscore the significance of employing causal language modeling when applying the index hint technique. Conversely, attempts to leverage prefix language modeling paired with bidirectional attention in model inputs consistently falter in length generalization. Our three other model variants with size [2M, 5M, 268M] consist of [2, 4, 16] blocks, a [256, 256, 1024] hidden size, a feedforward layer with a hidden dimension of [1024, 1024, 4096], and a [4, 4, 16]-head attention mechanism, respectively. 

In our implementation of FIRE \cite{li2023functional}, we employ layerwise sharing of attention bias across all attention blocks to enhance training efficiency. The paraterization of FIRE consists of a 2-layer MLP with a 32-unit hidden layer, utilizing ReLU activation.

We use the AdamW optimizer \citep{loshchilov2017decoupled} to train the model with a weight decay value of 0.1 and dropout rate of 0.0. The learning rate schedule incorporates an initial 500-step linear warm-up, followed by a cosine decay, starting at 3e-4. We train the model with sequence packing, a batch size of 128, and a sequence length of 2048, over 50,000 steps. We use greedy decoding to generate the model output during evaluation. We summarize the hyperparameters in \Cref{tab:hparams_training}.

\begin{table}[h]
    \centering
    \caption{Hyperparameters Summary for Length Generalization}
    \begin{tabularx}{0.75\textwidth}{ll}
        \toprule
        \textbf{Hyperparameter} & \textbf{Value} \\
        \midrule
        Language Model Type & Causal \\
        Activation Functions & GeGLU \\
        Normalization Layer & RMSNorm \\
        Normalization Type & PreNorm and PostNorm \\
        \midrule
        Optimizer & AdamW \\
        Training Steps & 50,000 \\
        Batch size & 128 \\
        Weight Decay & 0.1 \\
        Dropout & 0.0 \\
        Learning Rate~(LR) & 0.0003\\
        LR Warmup Steps & 500 \\
        LR Cooldown (Begin, End) & (500, 50,000) \\
        Warmup Schedule & Linear (from 0 to LR)\\
        Cooldown Schedule & Cosine Decay (from LR to 0.1LR)\\
        Training Sequence Length & 2048 \\
        Evaluation & Greedy \\
        \bottomrule
    \end{tabularx}
    \label{tab:hparams_training}
\end{table}

%% file: appendix/additional_results.tex
\clearpage
\newpage

\section{Additional Results}
\subsection{Training Loss and Sequence Exact Match Accuracy of Reverse Format with Index Hint trained up to 40}\label{sec:loss_acc_reverse_index_hint}
\begin{figure}[!h]
\centering
\begin{subfigure}[t]{0.45\textwidth}
\centering
\includegraphics[width=1.0\columnwidth]{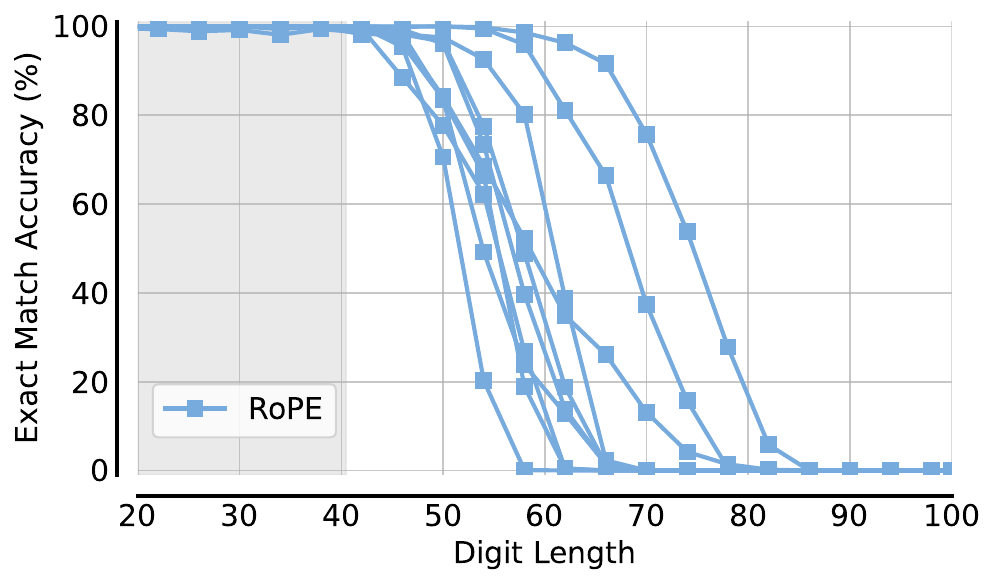}
\end{subfigure}
\begin{subfigure}[t]{0.45\textwidth}
\centering
\includegraphics[width=1.0\columnwidth]{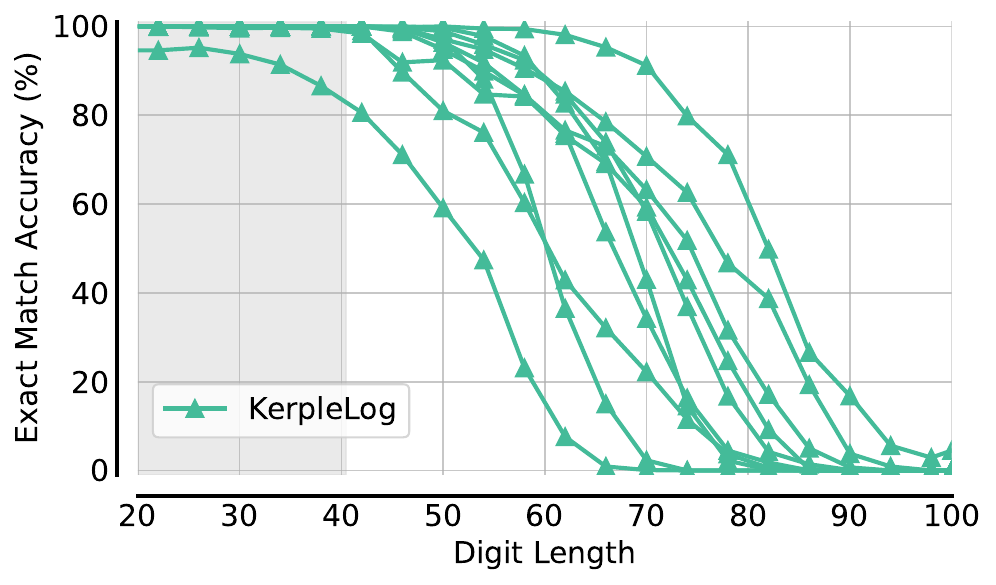}
\end{subfigure}
\begin{subfigure}[t]{0.45\textwidth}
\centering
\includegraphics[width=1.0\columnwidth]{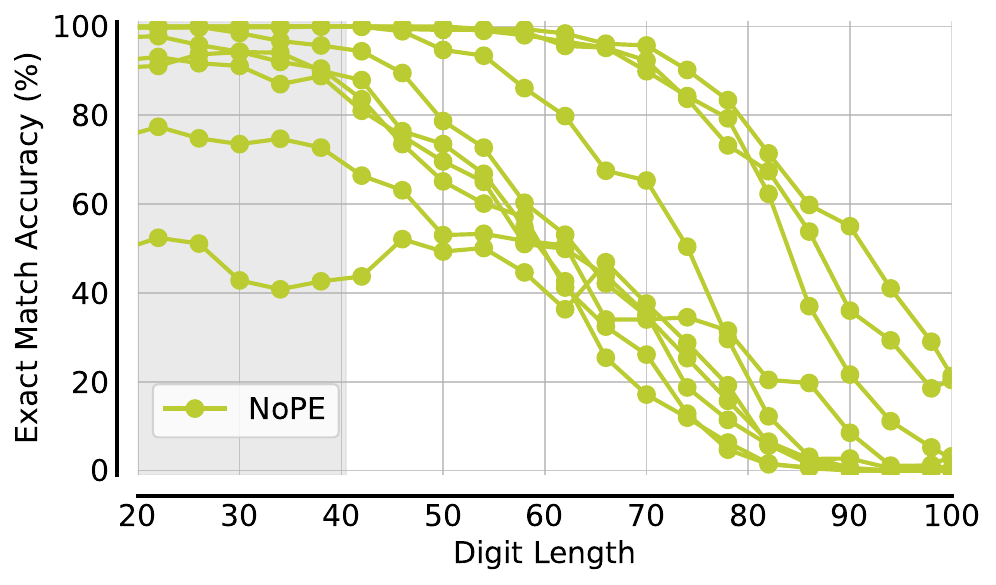}
\end{subfigure}
\begin{subfigure}[t]{0.45\textwidth}
\centering
\includegraphics[width=1.0\columnwidth]{figures/files/seed_em_tlen40_fire}
\end{subfigure}
\vspace{-2mm}
\caption{Exact match accuracy on 20 to 100 digit addition of all 10 trials trained on up to 40-digit addition with index hint and reverse format using four different position encodings.
}\label{fig_app:seed_acc_tlen40}
\end{figure}

\begin{figure}[!h]
\centering
\begin{subfigure}[t]{0.45\textwidth}
\centering
\includegraphics[width=1.0\columnwidth]{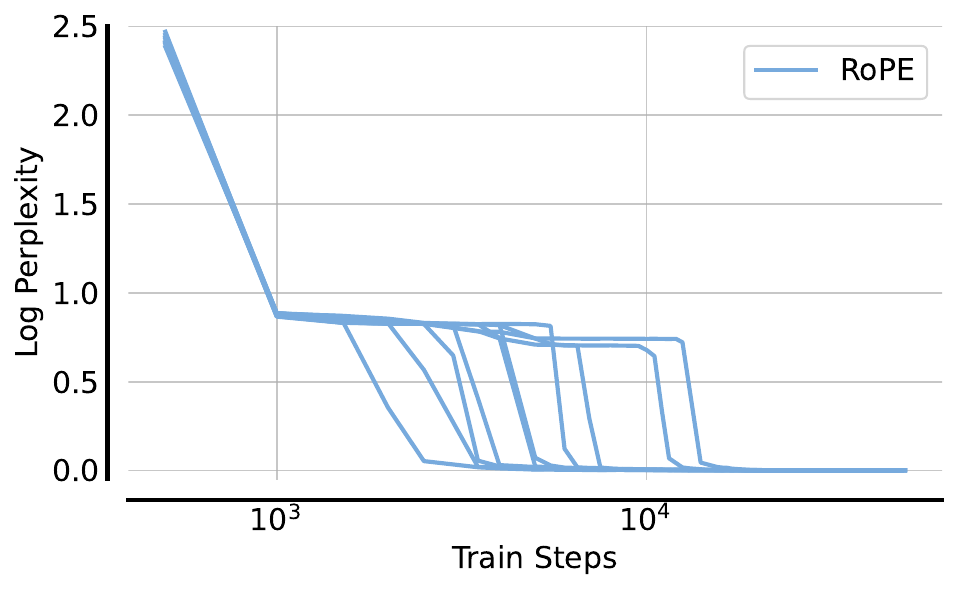}
\end{subfigure}
\begin{subfigure}[t]{0.45\textwidth}
\centering
\includegraphics[width=1.0\columnwidth]{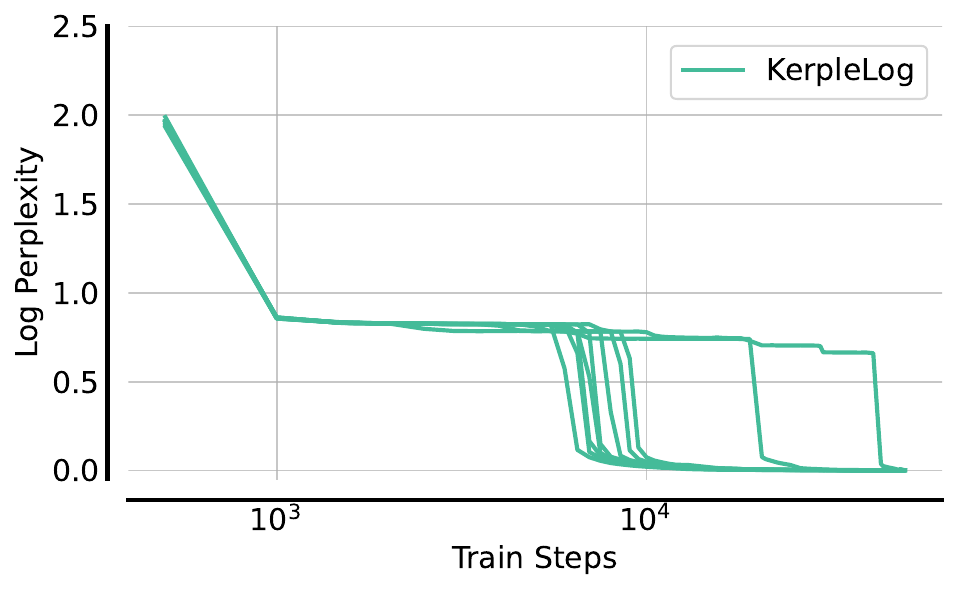}
\end{subfigure}
\begin{subfigure}[t]{0.45\textwidth}
\centering
\includegraphics[width=1.0\columnwidth]{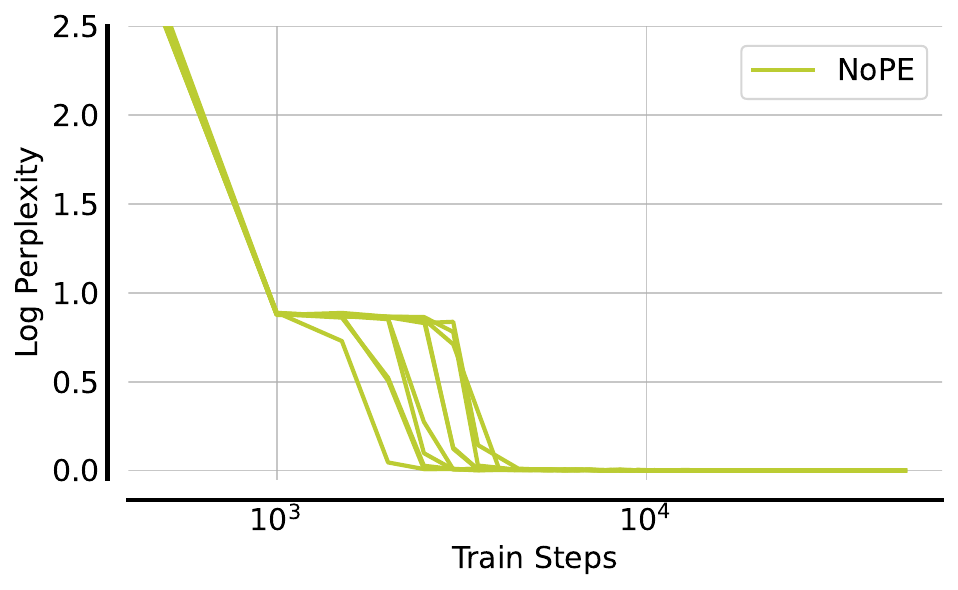}
\end{subfigure}
\begin{subfigure}[t]{0.45\textwidth}
\centering
\includegraphics[width=1.0\columnwidth]{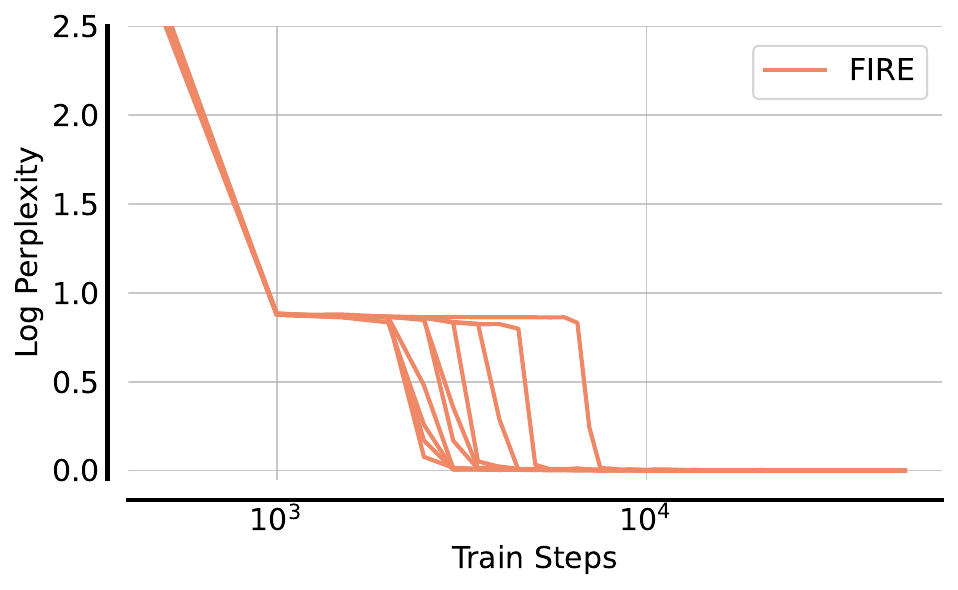}
\end{subfigure}
\vspace{-2mm}
\caption{Training loss over 10 trials in reverse formats. Despite similar nearly 0 log perplexity losses across runs after 10K training steps, different runs exhibit very different length generalization.
}\label{fig_app:logpplx_logx_all}
\end{figure}

\input{figures/appendix/grokking_app}

\clearpage
\newpage
\subsection{The evolution of EM Accuracy during training in reverse format using 4 PEs}
\label{sec:acc_vs_steps_reverse_index_hint}
\begin{figure}[!h]
\centering
\begin{subfigure}[t]{0.45\textwidth}
\centering
\includegraphics[width=1.0\columnwidth]{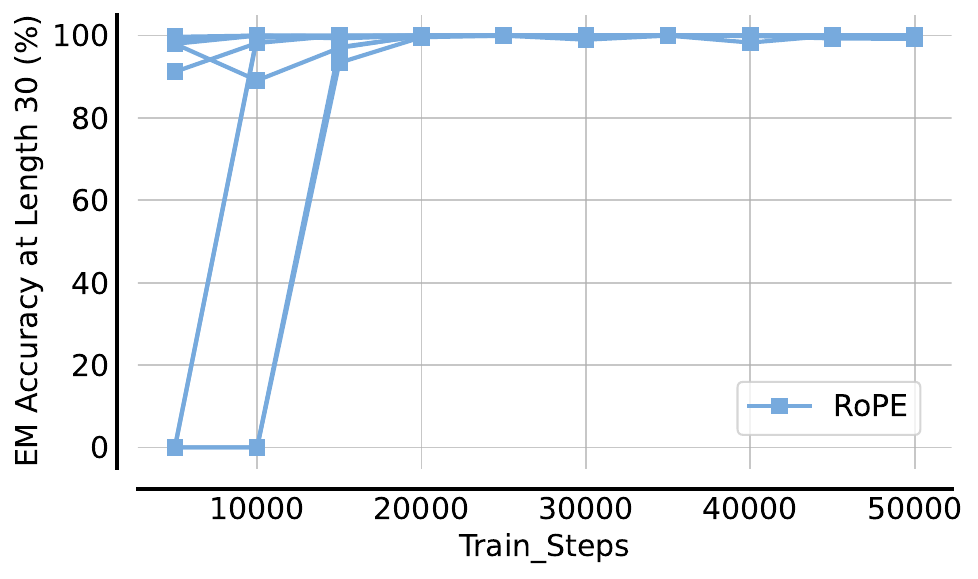}
\end{subfigure}
\begin{subfigure}[t]{0.45\textwidth}
\centering
\includegraphics[width=1.0\columnwidth]{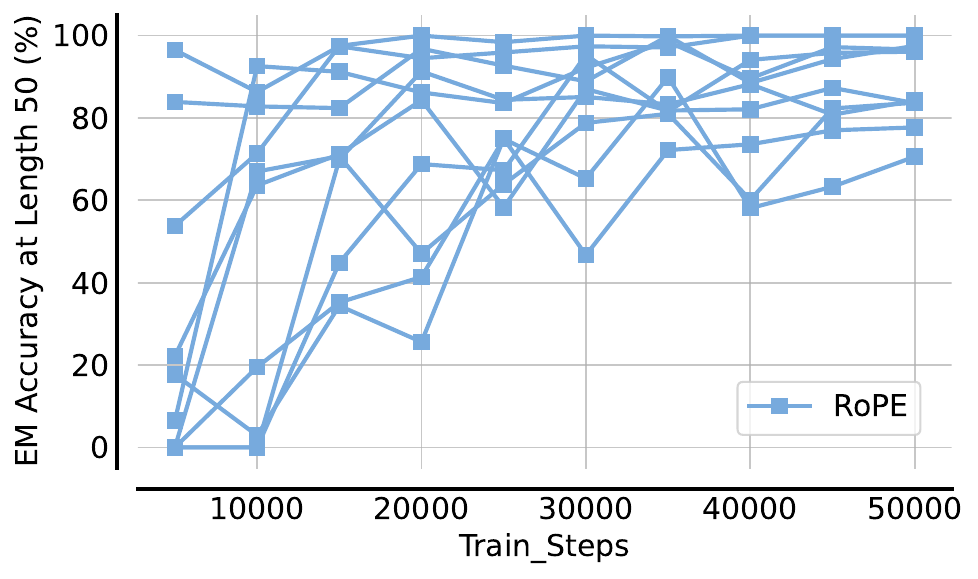}
\end{subfigure}
\begin{subfigure}[t]{0.45\textwidth}
\centering
\includegraphics[width=1.0\columnwidth]{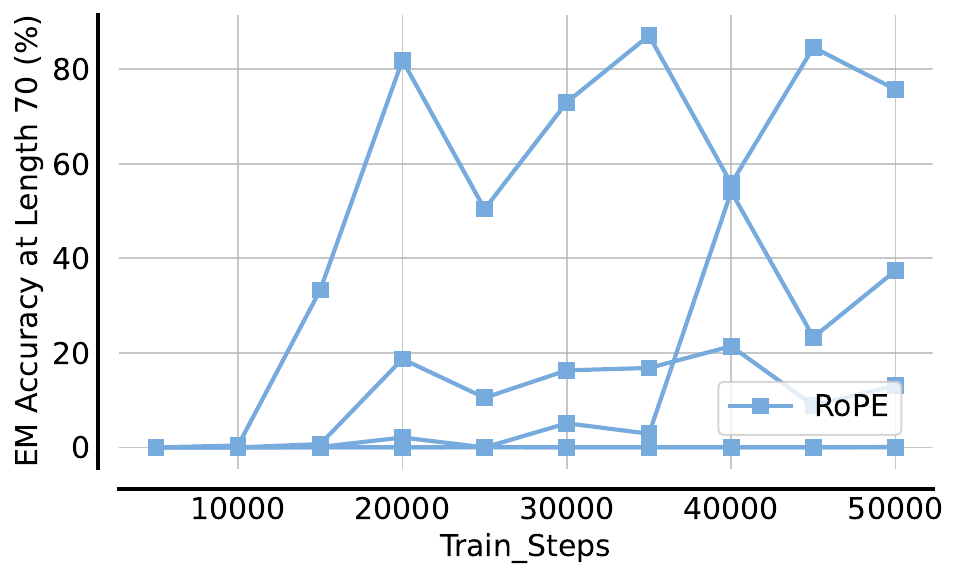}
\end{subfigure}
\begin{subfigure}[t]{0.45\textwidth}
\centering
\includegraphics[width=1.0\columnwidth]{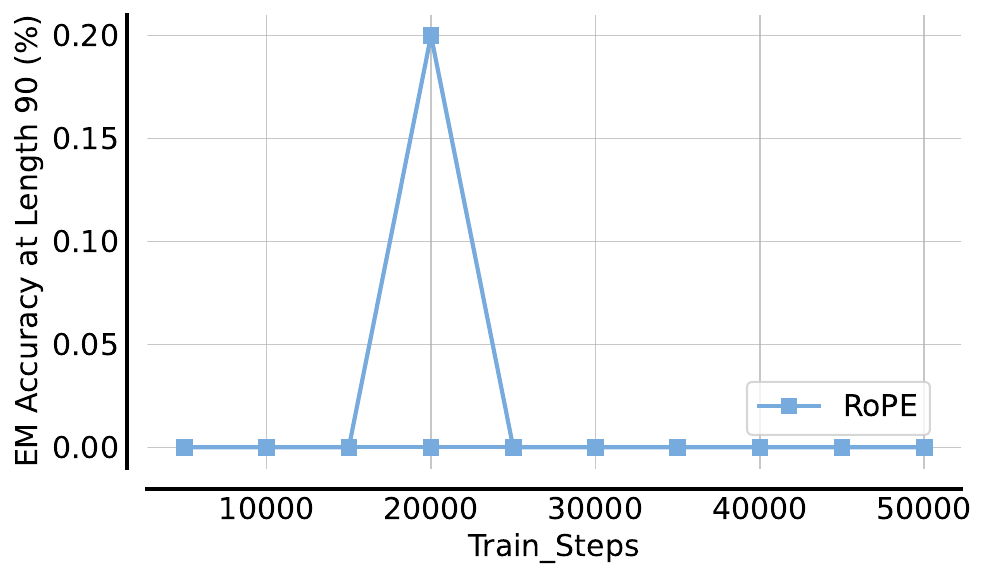}
\end{subfigure}
\vspace{-2mm}
\caption{\small Exact match accuracy on $[30, 50, 70, 90]$ digit addition of all 10 trials trained on up to 40-digit addition with index hint and reverse format using RoPE.
}\label{fig_app:acc_steps_rope}
\end{figure}

\begin{figure}[!h]
\centering
\begin{subfigure}[t]{0.45\textwidth}
\centering
\includegraphics[width=1.0\columnwidth]{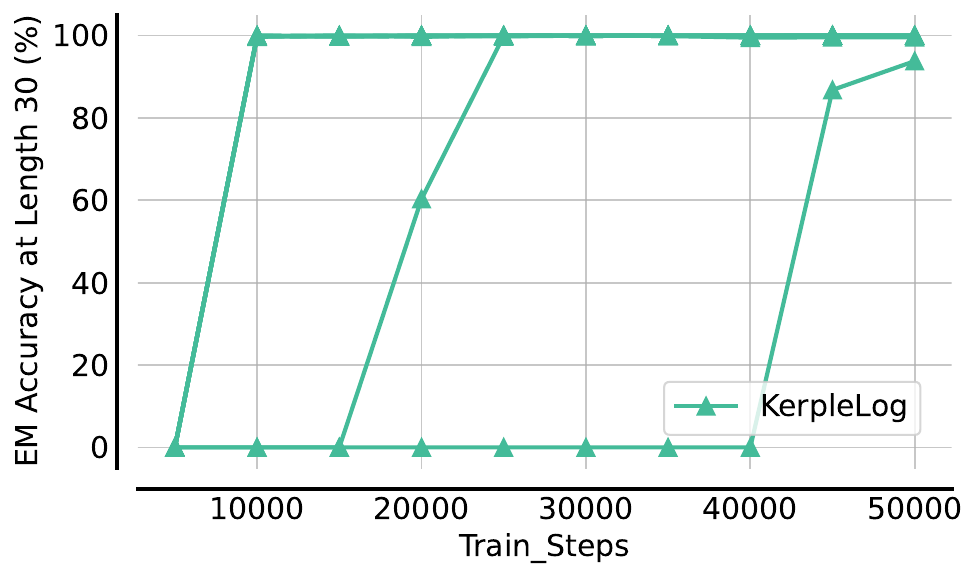}
\end{subfigure}
\begin{subfigure}[t]{0.45\textwidth}
\centering
\includegraphics[width=1.0\columnwidth]{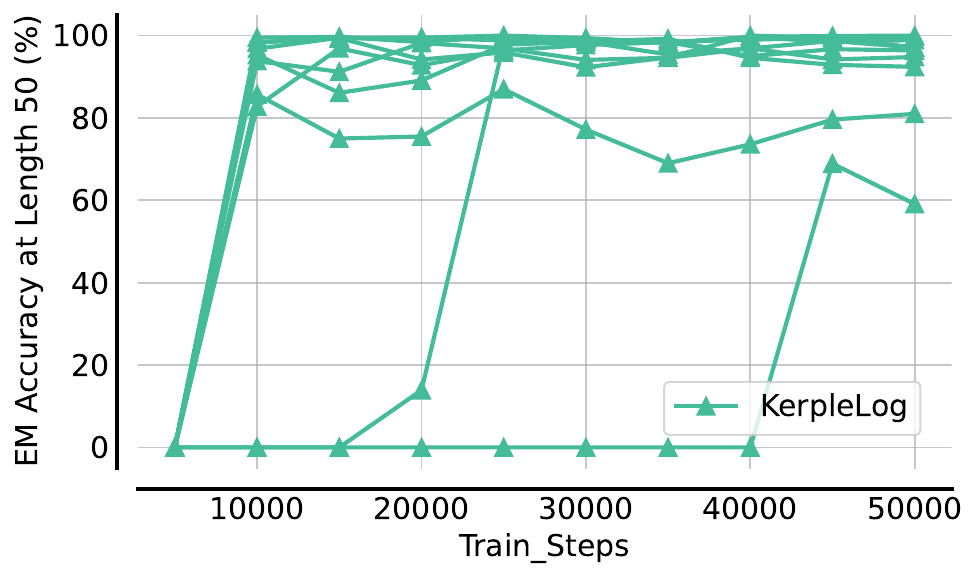}
\end{subfigure}
\begin{subfigure}[t]{0.45\textwidth}
\centering
\includegraphics[width=1.0\columnwidth]{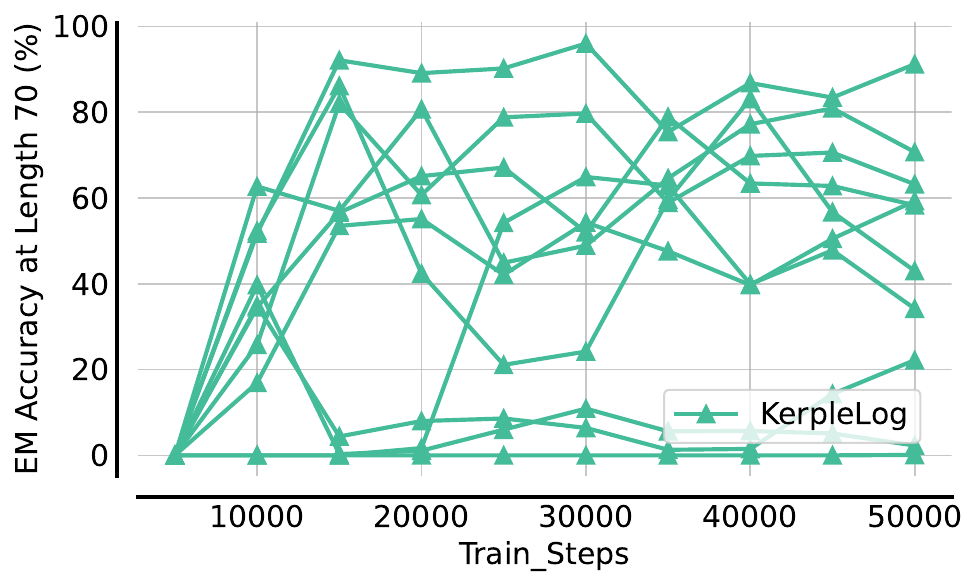}
\end{subfigure}
\begin{subfigure}[t]{0.45\textwidth}
\centering
\includegraphics[width=1.0\columnwidth]{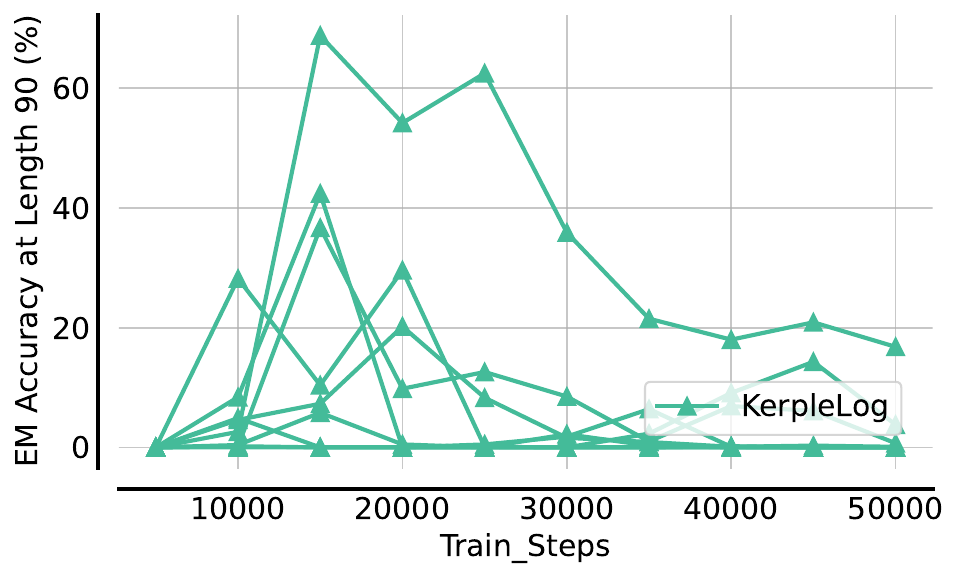}
\end{subfigure}
\vspace{-2mm}
\caption{\small Exact match accuracy on $[30, 50, 70, 90]$ digit addition of all 10 trials trained on up to 40-digit addition with index hint and reverse format using KerpleLog.
}\label{fig_app:acc_steps_kerple}
\end{figure}

\begin{figure}[!h]
\centering
\begin{subfigure}[t]{0.45\textwidth}
\centering
\includegraphics[width=1.0\columnwidth]{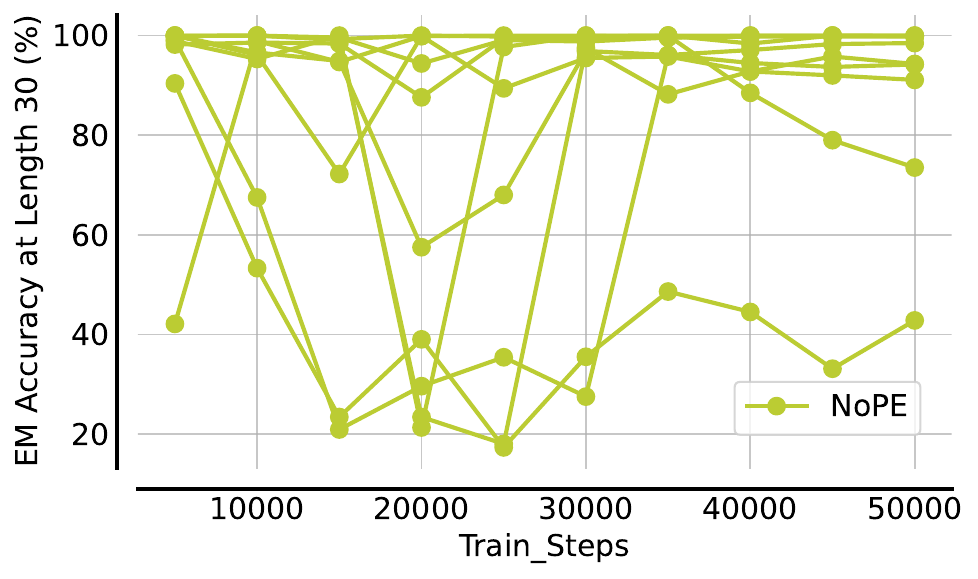}
\end{subfigure}
\begin{subfigure}[t]{0.45\textwidth}
\centering
\includegraphics[width=1.0\columnwidth]{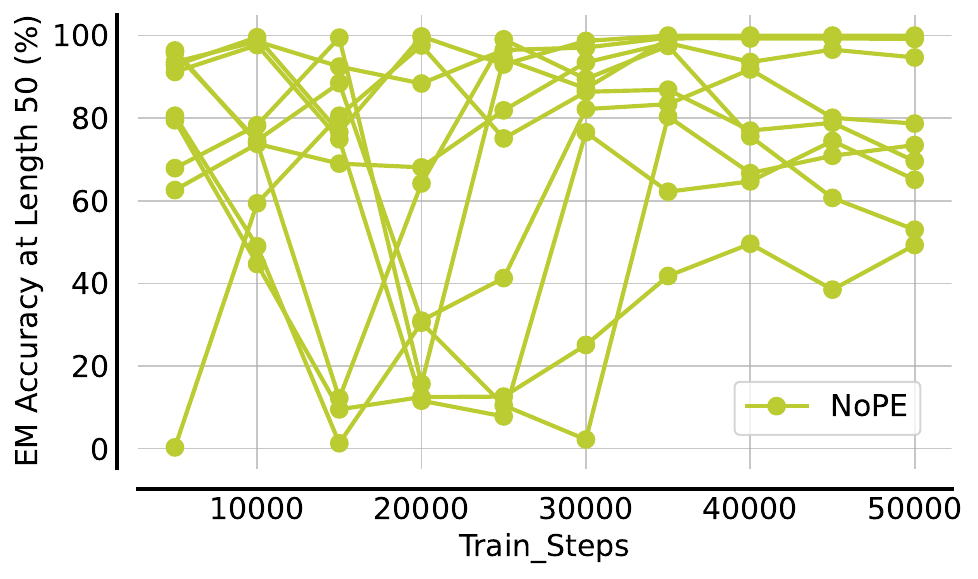}
\end{subfigure}
\begin{subfigure}[t]{0.45\textwidth}
\centering
\includegraphics[width=1.0\columnwidth]{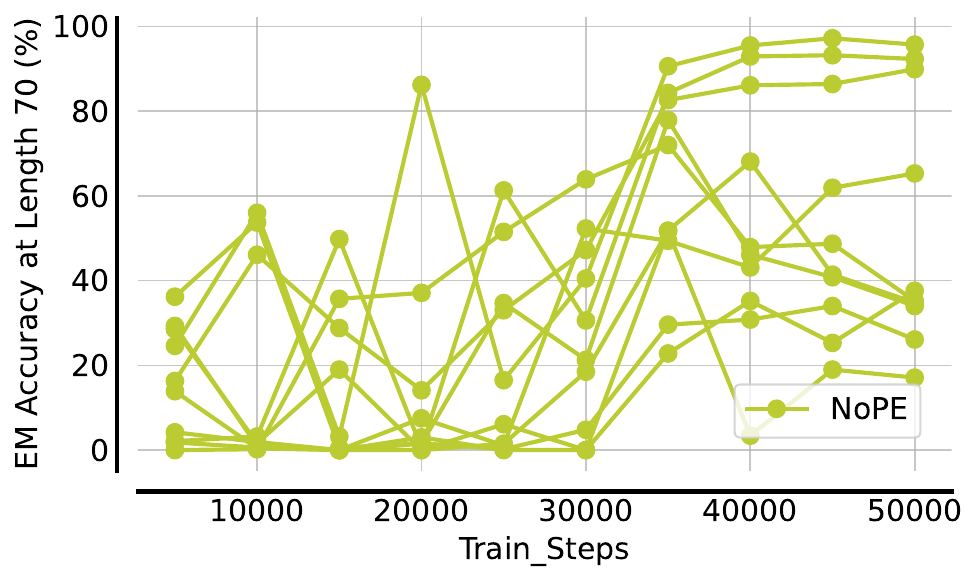}
\end{subfigure}
\begin{subfigure}[t]{0.45\textwidth}
\centering
\includegraphics[width=1.0\columnwidth]{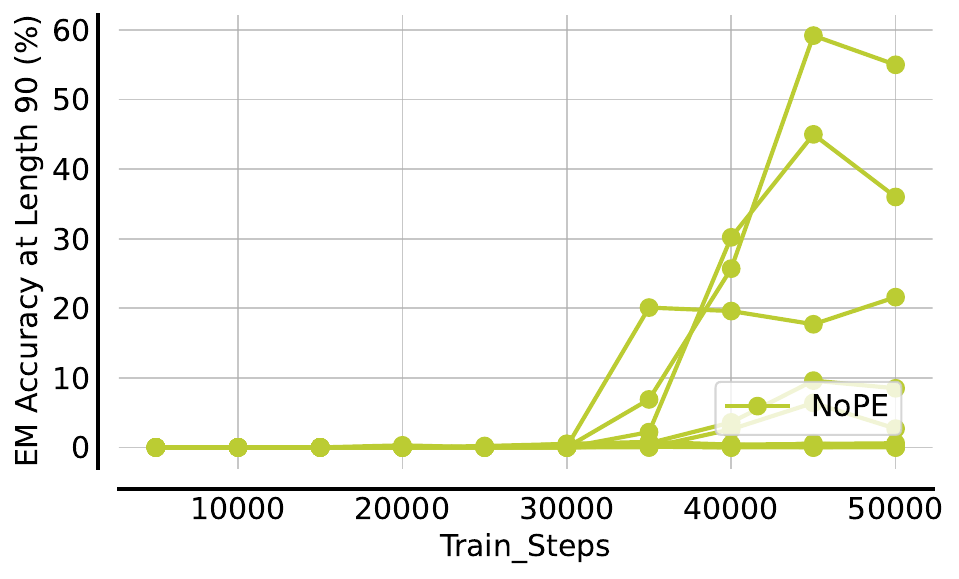}
\end{subfigure}
\vspace{-2mm}
\caption{\small Exact match accuracy on $[30, 50, 70, 90]$ digit addition of all 10 trials trained on up to 40-digit addition with index hint and reverse format using NoPE.
}\label{fig_app:acc_steps_nope}
\end{figure}

\begin{figure}[!h]
\centering
\begin{subfigure}[t]{0.45\textwidth}
\centering
\includegraphics[width=1.0\columnwidth]{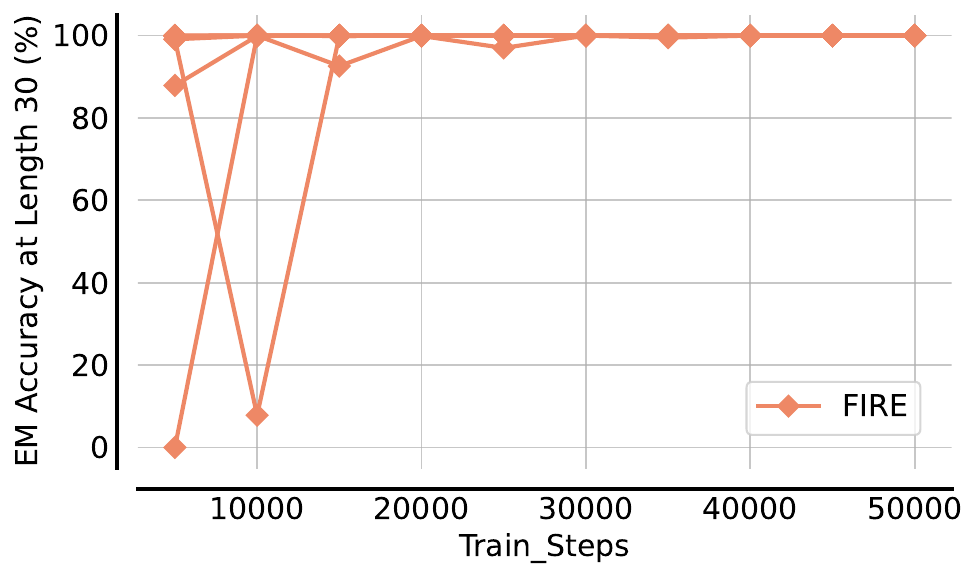}
\end{subfigure}
\begin{subfigure}[t]{0.45\textwidth}
\centering
\includegraphics[width=1.0\columnwidth]{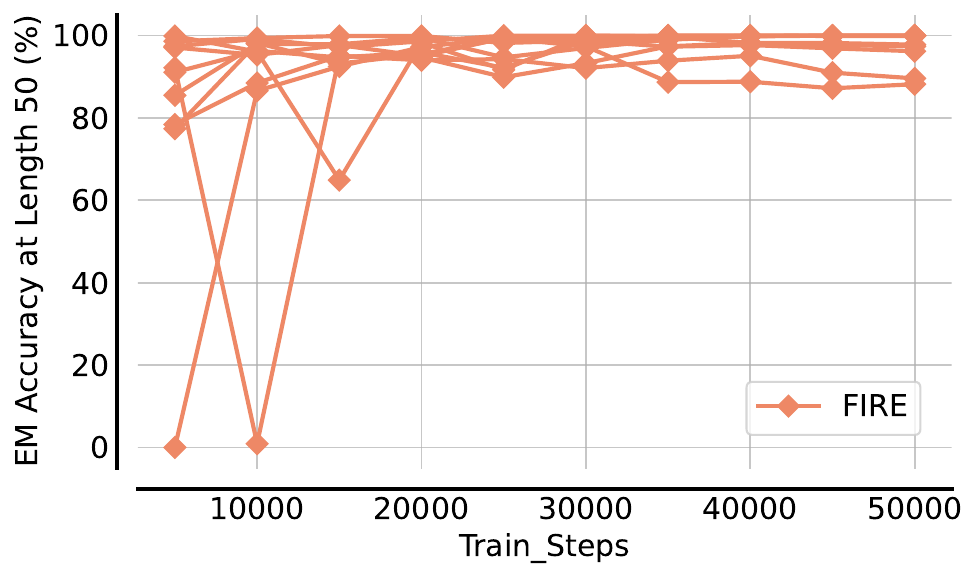}
\end{subfigure}
\begin{subfigure}[t]{0.45\textwidth}
\centering
\includegraphics[width=1.0\columnwidth]{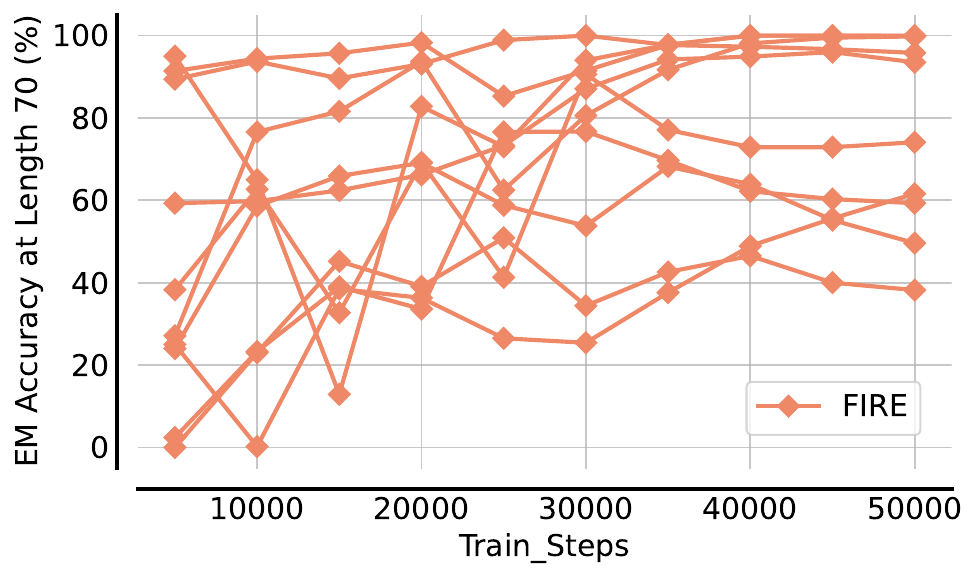}
\end{subfigure}
\begin{subfigure}[t]{0.45\textwidth}
\centering
\includegraphics[width=1.0\columnwidth]{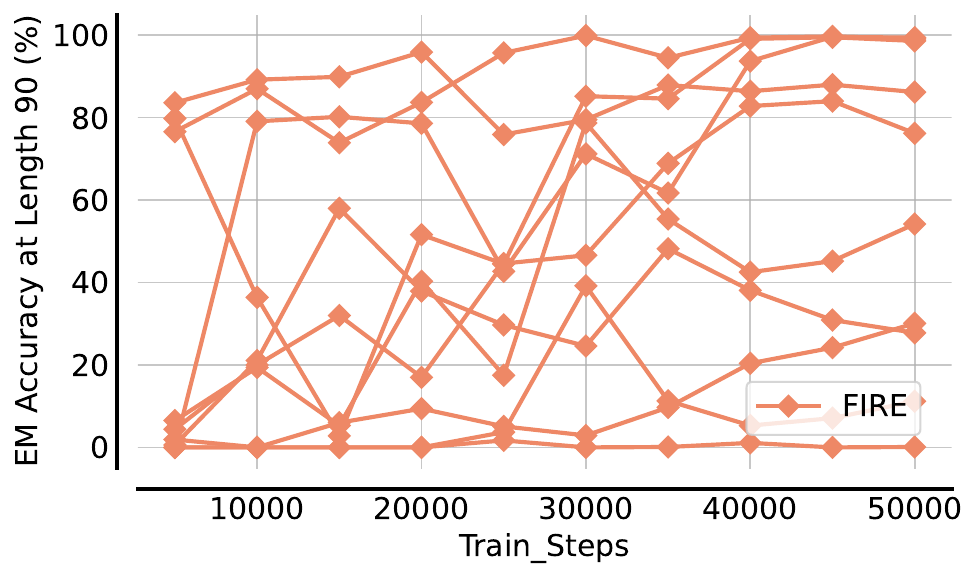}
\end{subfigure}
\vspace{-2mm}
\caption{\small Exact match accuracy on $[30, 50, 70, 90]$ digit addition of all 10 trials trained on up to 40-digit addition with index hint and reverse format using FIRE.
}\label{fig_app:acc_steps_fire}
\end{figure}

\clearpage
\newpage
\subsection{Effect of Index Hint}
\begin{figure}[!h]
\centering
\begin{subfigure}[t]{0.45\textwidth}
\centering
\includegraphics[width=1.0\columnwidth]{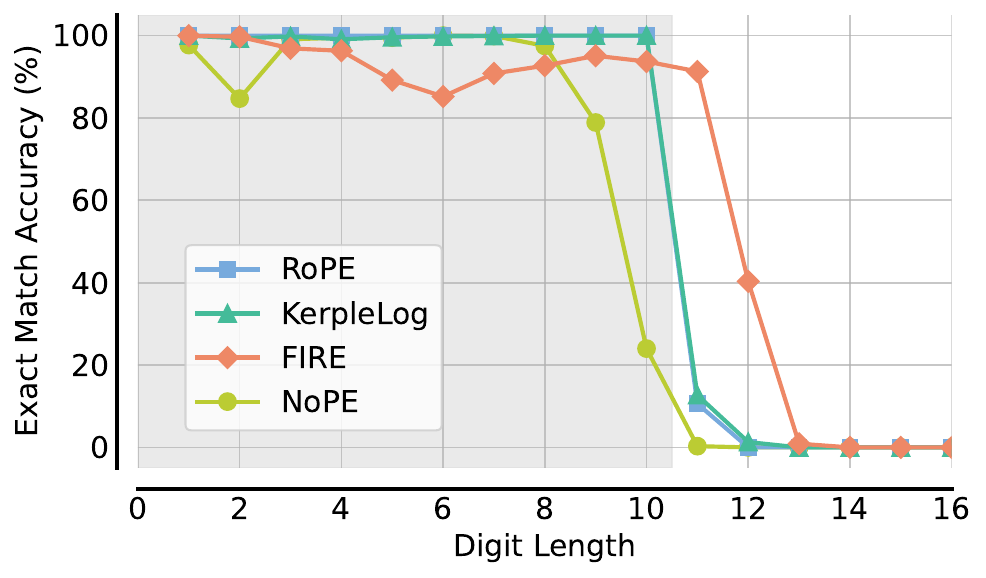}
\caption{Models trained up to 10-digit addition}
\label{fig_app:em_no_index_hint:a}
\end{subfigure}
\begin{subfigure}[t]{0.45\textwidth}
\centering
\includegraphics[width=1.0\columnwidth]{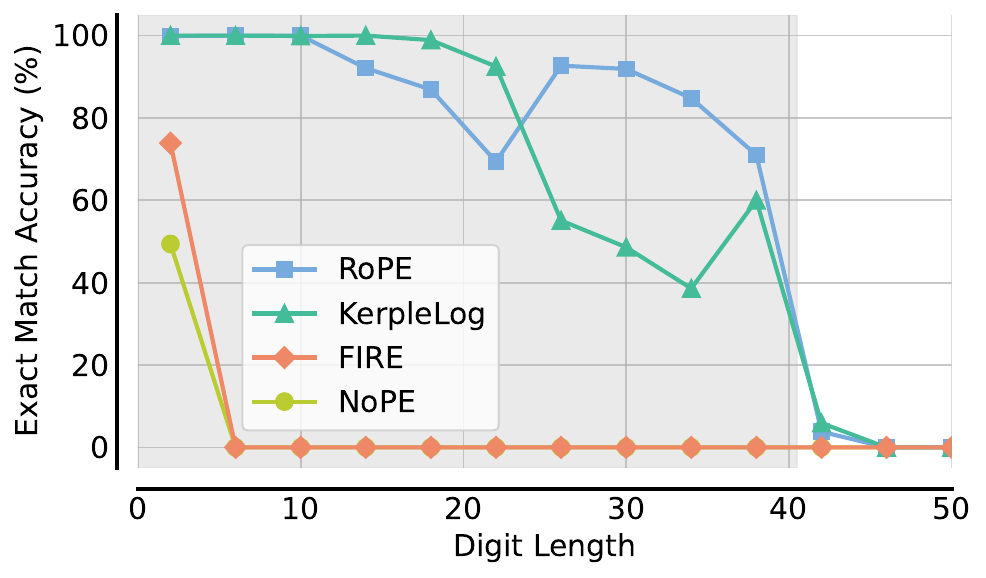}
\caption{Models trained up to 40-digit addition}
\label{fig_app:em_no_index_hint:b}
\end{subfigure}
\vspace{-2mm}
\caption{\small Best sequence exact match accuracy over five trials without index hint, trained upto length 10. All position encoding methods fail to generalize beyond trivial lengths and struggle with in-distribution generalization, highlighting the crucial role of index hints in length generalization. See the performance of each run in \Cref{sec:loss_acc_reverse_wo_index_hint_10} and \Cref{sec:loss_acc_reverse_wo_index_hint} for trained up to 10-digit and 40-digit addition.
}\label{fig_app:em_no_index_hint}
\end{figure}

\subsection{Source of Variance}
\begin{figure}[!h]
\centering
\begin{subfigure}[t]{0.45\textwidth}
\centering
\includegraphics[width=1.0\columnwidth]{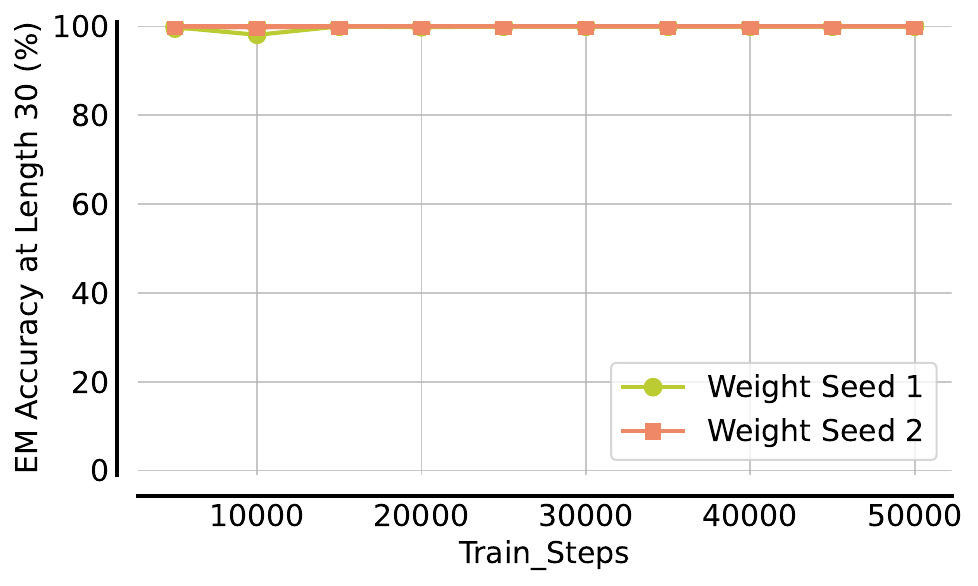}
\end{subfigure}
\begin{subfigure}[t]{0.45\textwidth}
\centering
\includegraphics[width=1.0\columnwidth]{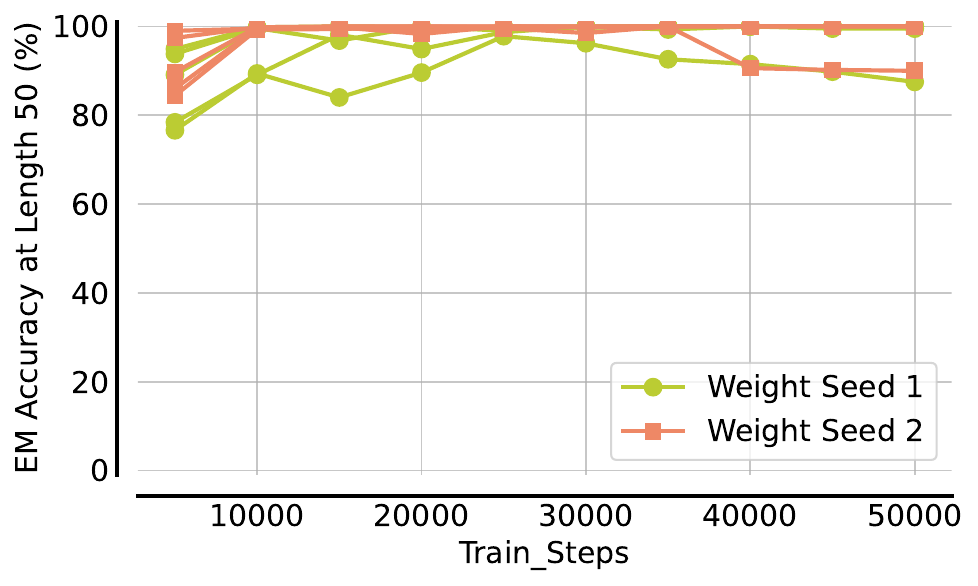}
\end{subfigure}
\begin{subfigure}[t]{0.45\textwidth}
\centering
\includegraphics[width=1.0\columnwidth]{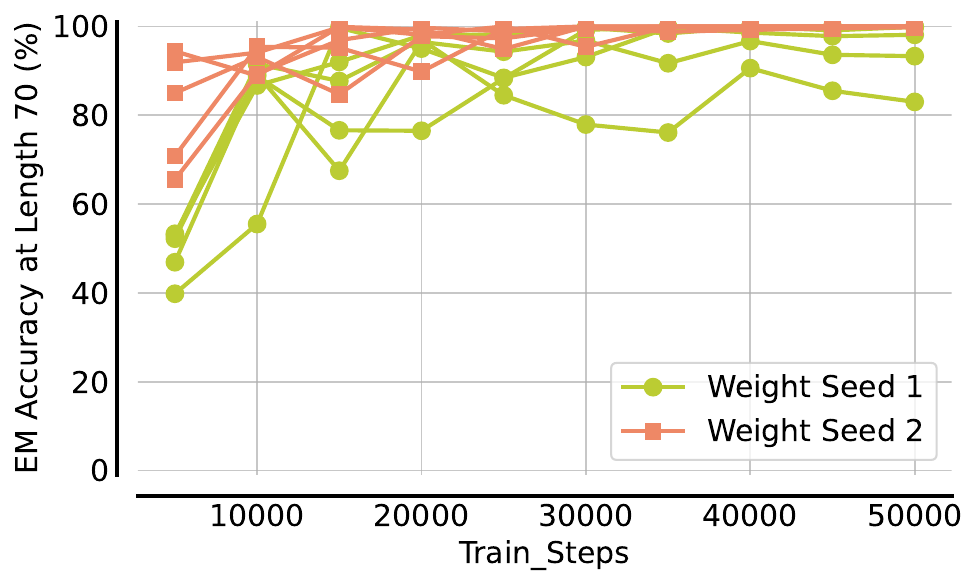}
\end{subfigure}
\begin{subfigure}[t]{0.45\textwidth}
\centering
\includegraphics[width=1.0\columnwidth]{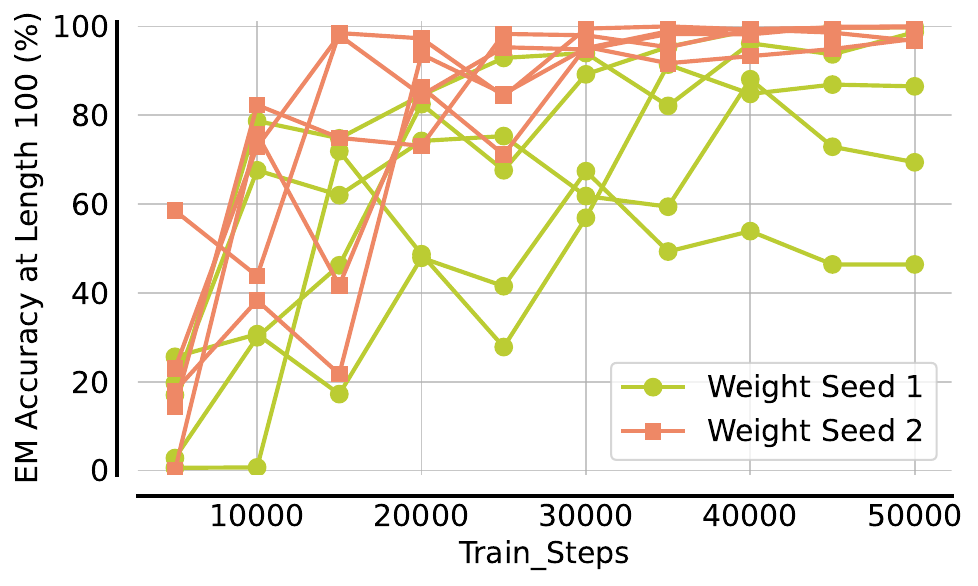}
\end{subfigure}
\vspace{-2mm}
\caption{\small Exact match accuracy on $[30, 50, 70, 100]$ digit addition of all 10 trials (2 weight seeds x 5 data seeds) trained on up to 40-digit addition with index hint and reverse format using FIRE.
}\label{fig_app:acc_steps_fire_order}
\end{figure}

\input{figures/appendix/error_analysis_app}

\input{figures/appendix/scaling}

\input{figures/appendix/hyperparameter_sensitivity_app}

\input{figures/appendix/loss_acc}

%% file: figures/appendix/grokking_app.tex
\clearpage
\newpage
\subsection{Training Loss and Next-token Prediction Accuracy of Standard and Reverse Format}
\begin{figure}[!h]
\centering
\begin{subfigure}[t]{0.45\textwidth}
\centering
\includegraphics[width=1.0\columnwidth]{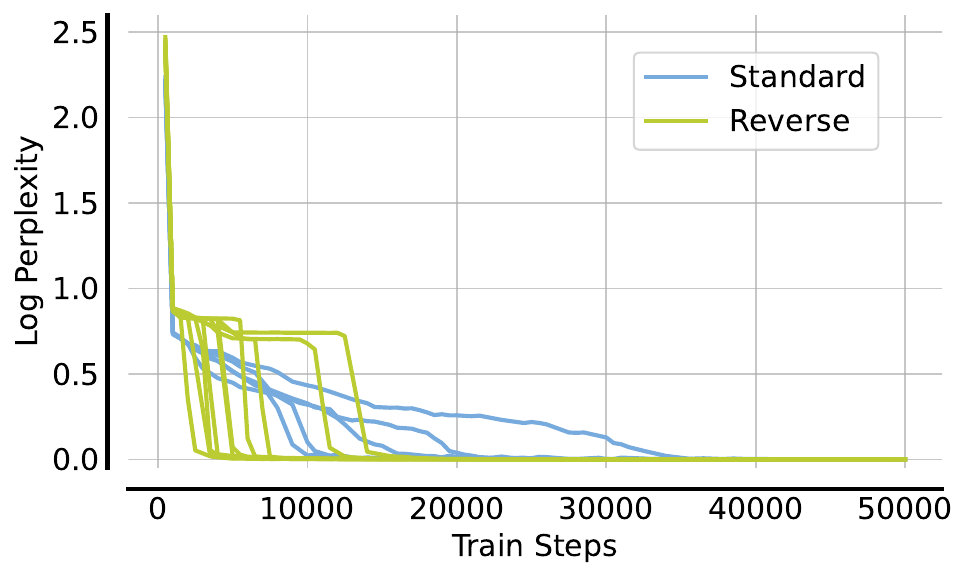}
\captionsetup{justification=centering}
\caption{Training Loss using RoPE}
\end{subfigure}
\begin{subfigure}[t]{0.45\textwidth}
\centering
\includegraphics[width=1.0\columnwidth]{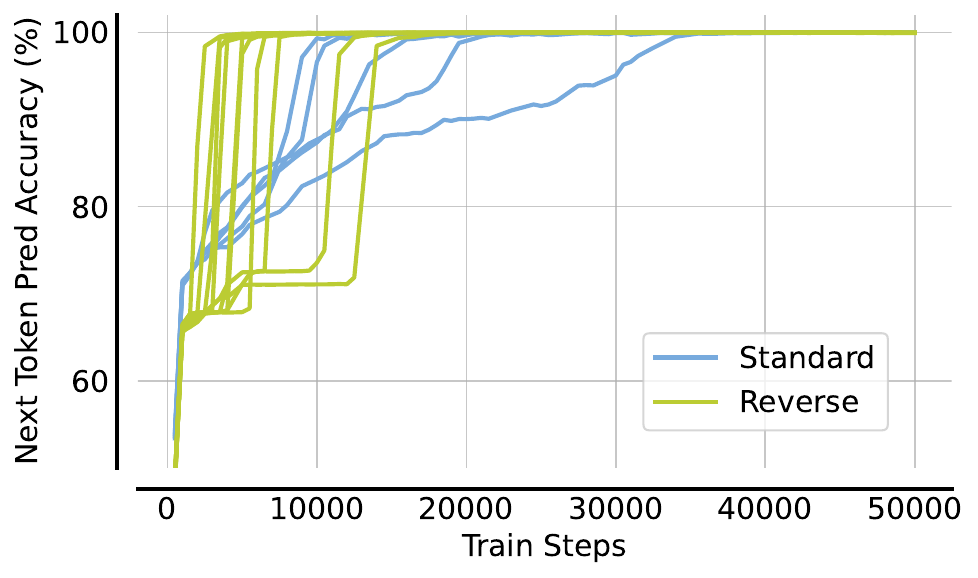}
\captionsetup{justification=centering}
\caption{Next-token Prediction Accuracy using RoPE}
\end{subfigure}
\begin{subfigure}[t]{0.45\textwidth}
\centering
\includegraphics[width=1.0\columnwidth]{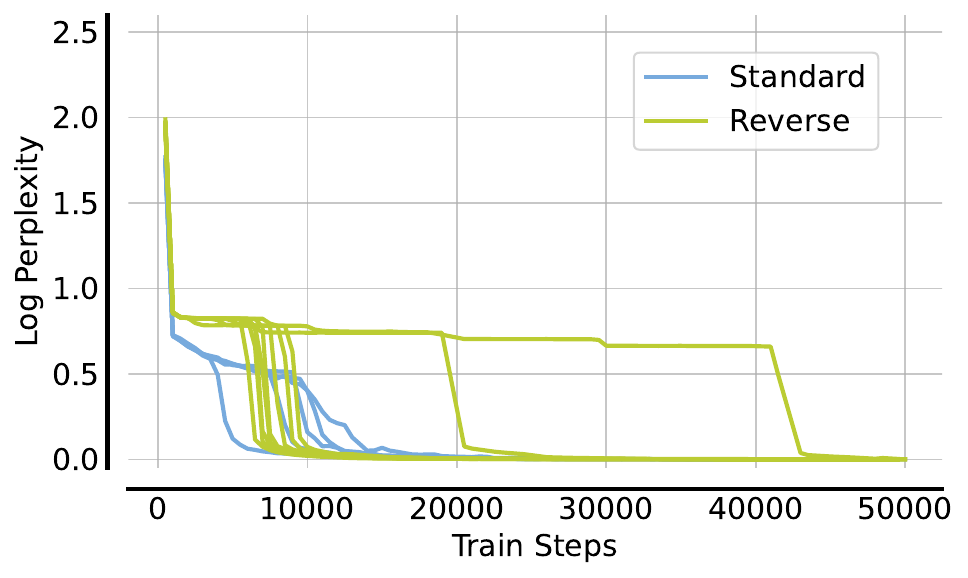}
\captionsetup{justification=centering}
\caption{Training Loss using KerpleLog}
\end{subfigure}
\begin{subfigure}[t]{0.45\textwidth}
\centering
\includegraphics[width=1.0\columnwidth]{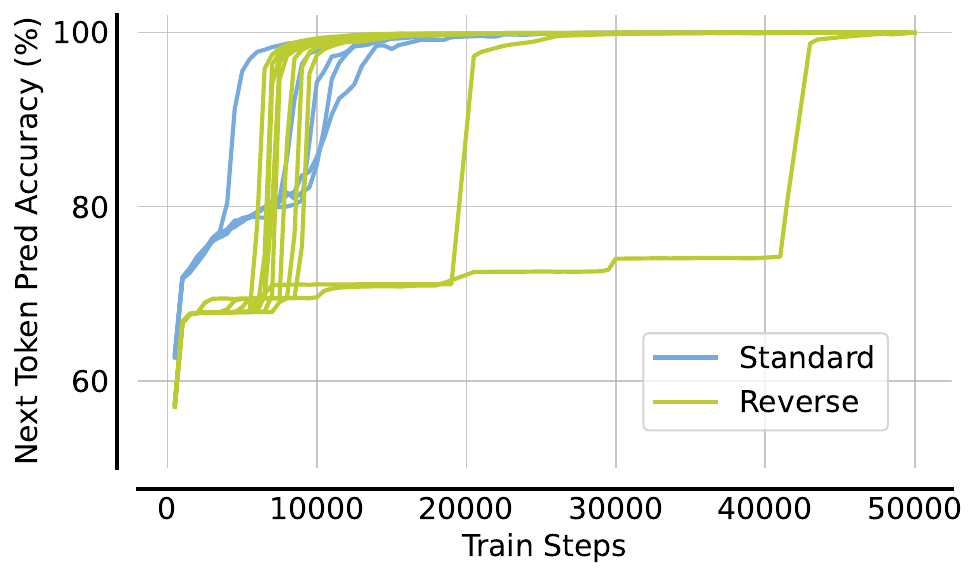}
\captionsetup{justification=centering}
\caption{Next-token Prediction Accuracy using KerpleLog}
\end{subfigure}
\begin{subfigure}[t]{0.45\textwidth}
\centering
\includegraphics[width=1.0\columnwidth]{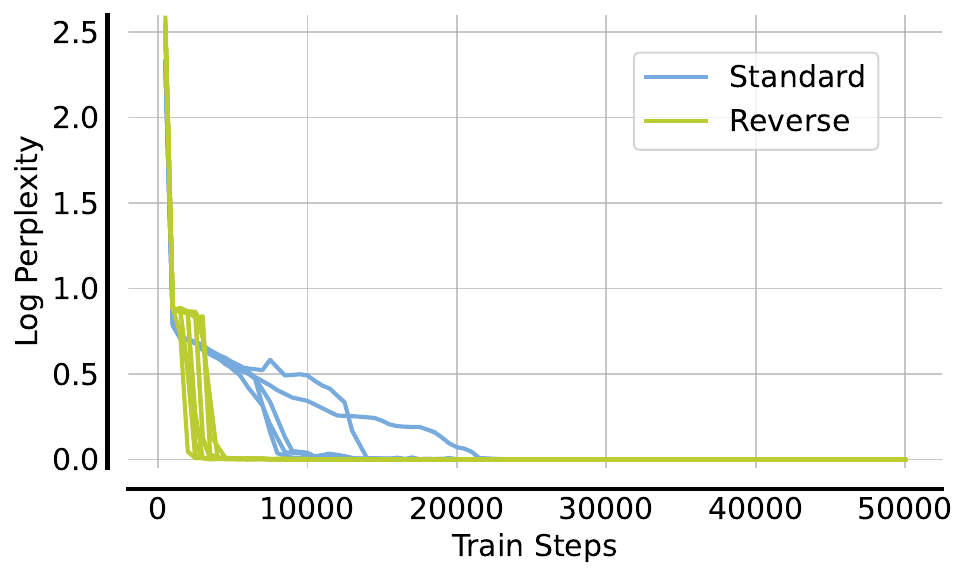}
\captionsetup{justification=centering}
\caption{Training Loss using NoPE}
\end{subfigure}
\begin{subfigure}[t]{0.45\textwidth}
\centering
\includegraphics[width=1.0\columnwidth]{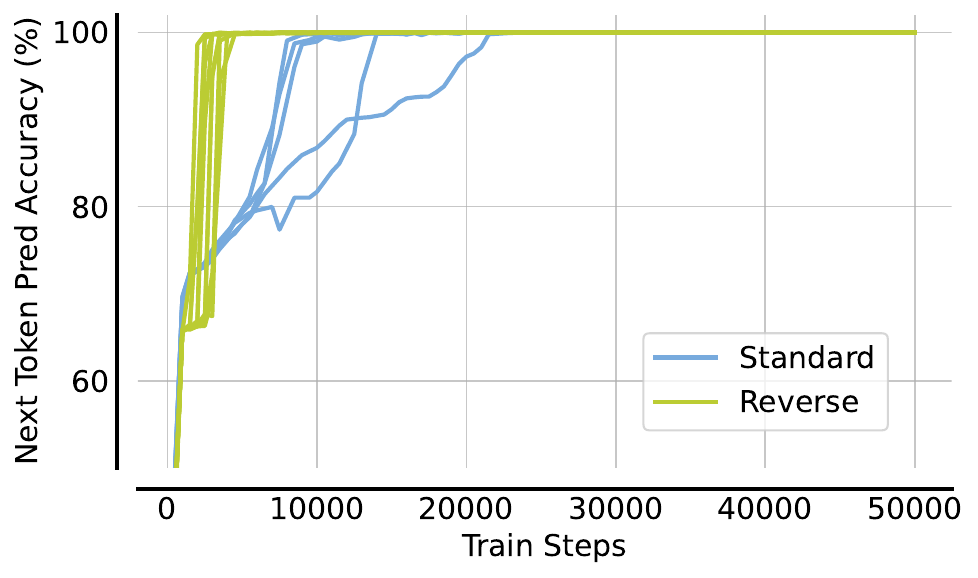}
\captionsetup{justification=centering}
\caption{Next-token Prediction Accuracy using NoPE}
\end{subfigure}
\begin{subfigure}[t]{0.45\textwidth}
\centering
\includegraphics[width=1.0\columnwidth]{figures/files/grokking_contrast_logpplx_fire.pdf}
\captionsetup{justification=centering}
\caption{Training Loss using FIRE}
\end{subfigure}
\begin{subfigure}[t]{0.45\textwidth}
\centering
\includegraphics[width=1.0\columnwidth]{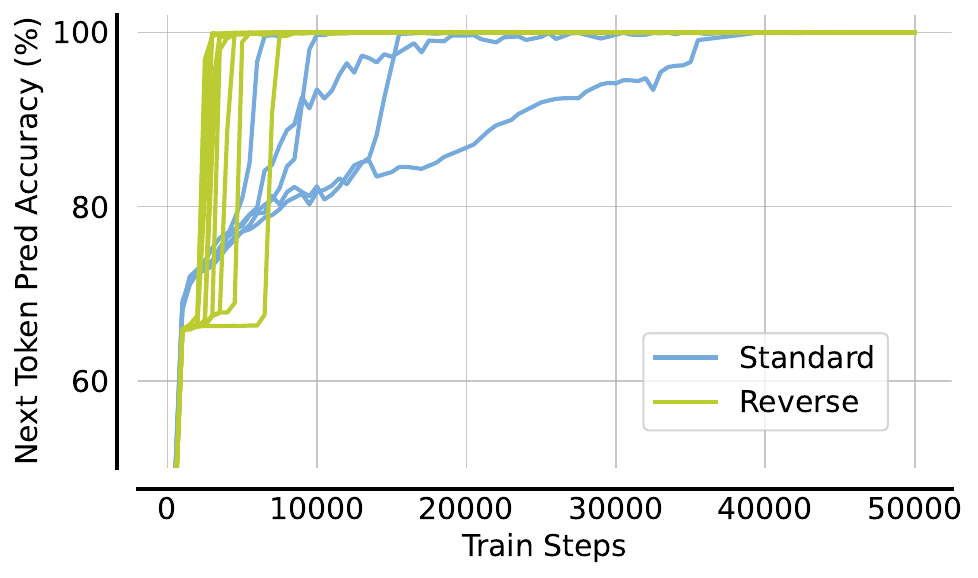}
\captionsetup{justification=centering}
\caption{Next-token Prediction Accuracy using FIRE}
\end{subfigure}
\vspace{-2mm}
\caption{\small Training log perplexity and next-token prediction accuracy over 10 trials in standard versus reverse formats using RoPE, KerpleLog, NoPE and FIRE. Reverse format shows distinct grokking during training, unlike the gradual enhancement in standard format. 
}\label{fig_app:logpplx_acc_step_4pe}
\end{figure}

%% file: figures/appendix/error_analysis_app.tex
\clearpage
\newpage
\subsection{Error Analysis}

\begin{figure}[!h]
\centering
\includegraphics[width=0.47\columnwidth]{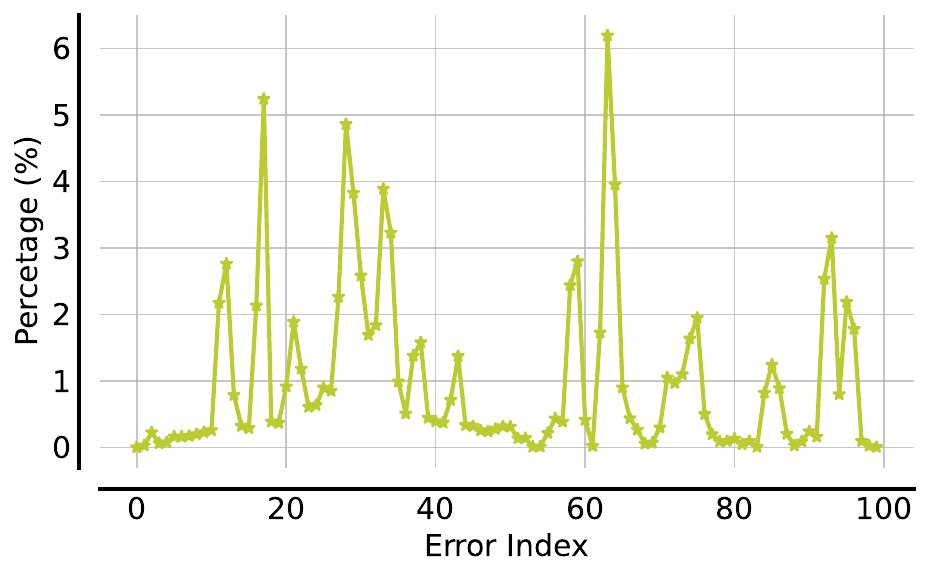}
\includegraphics[width=0.47\columnwidth]{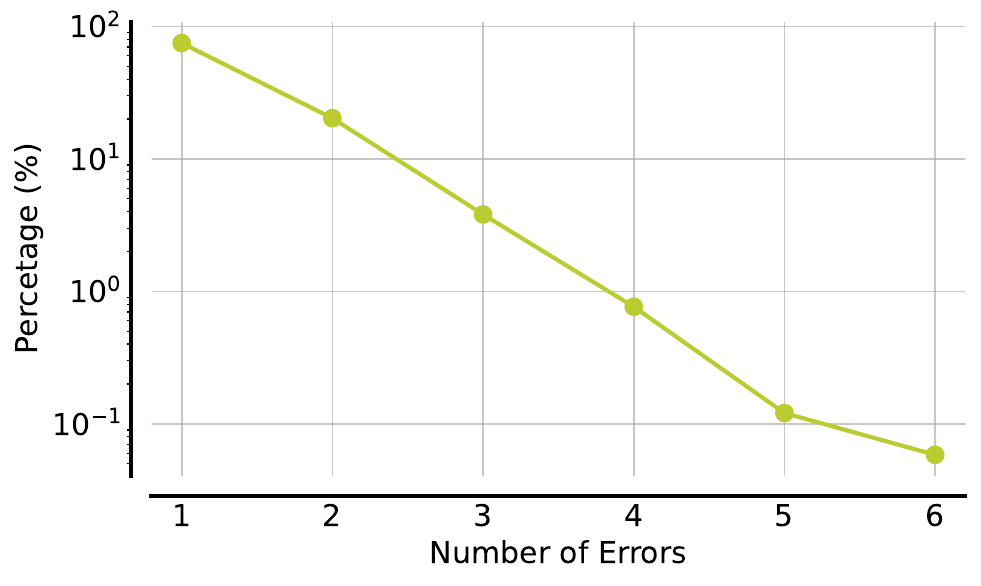}
\vspace{-2mm}
\caption{\small 
(Left) Average error position distribution over 10 runs, showing a broad error spread across all positions. Specific checkpoints exhibit a propensity for errors at certain positions (refer to Figure~\ref{fig_app:error_pos_hist}). 
(Right) Notably, in successful generalizations, more than 90\% of errors are confined to single-digit inaccuracies, exhibiting an exponential distribution.
}\label{fig_app:error_analysis}
\end{figure}

\begin{figure}[!h]
\centering
\includegraphics[width=1.0\columnwidth]{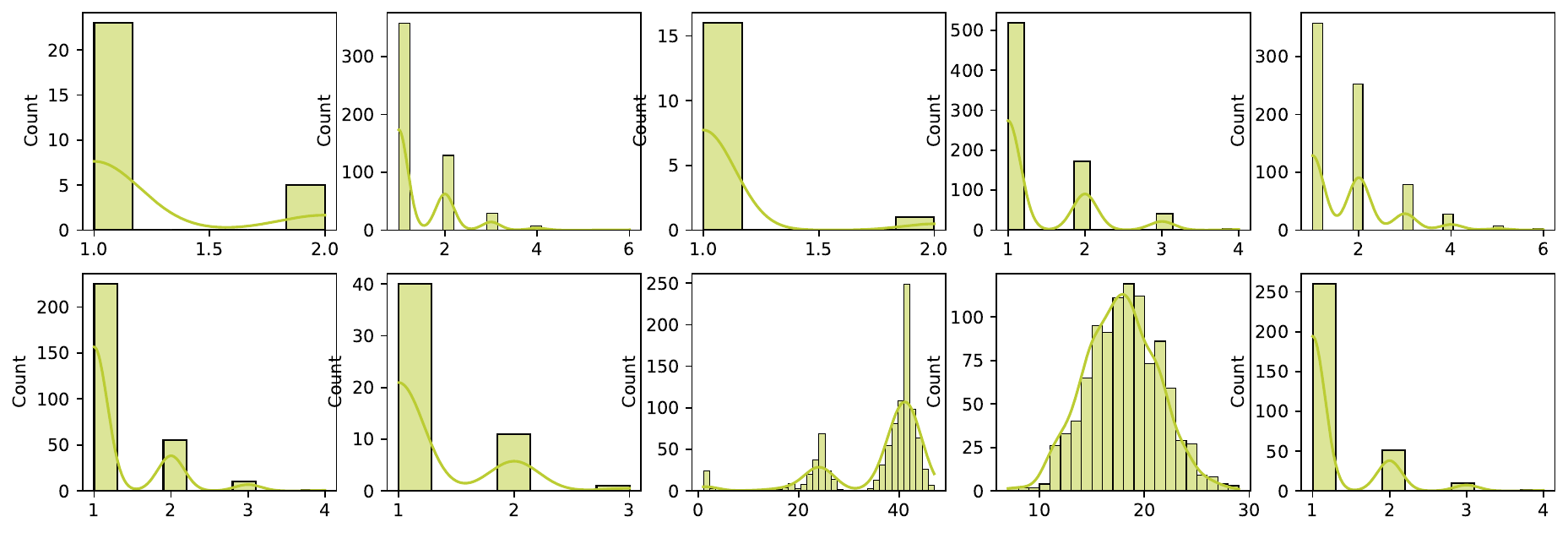}
\captionsetup{justification=centering}
\vspace{-2mm}
\caption{\small Error count distribution
}\label{fig_app:error_count_hist}
\end{figure}

\begin{figure}[!h]
\centering
\includegraphics[width=1.0\columnwidth]{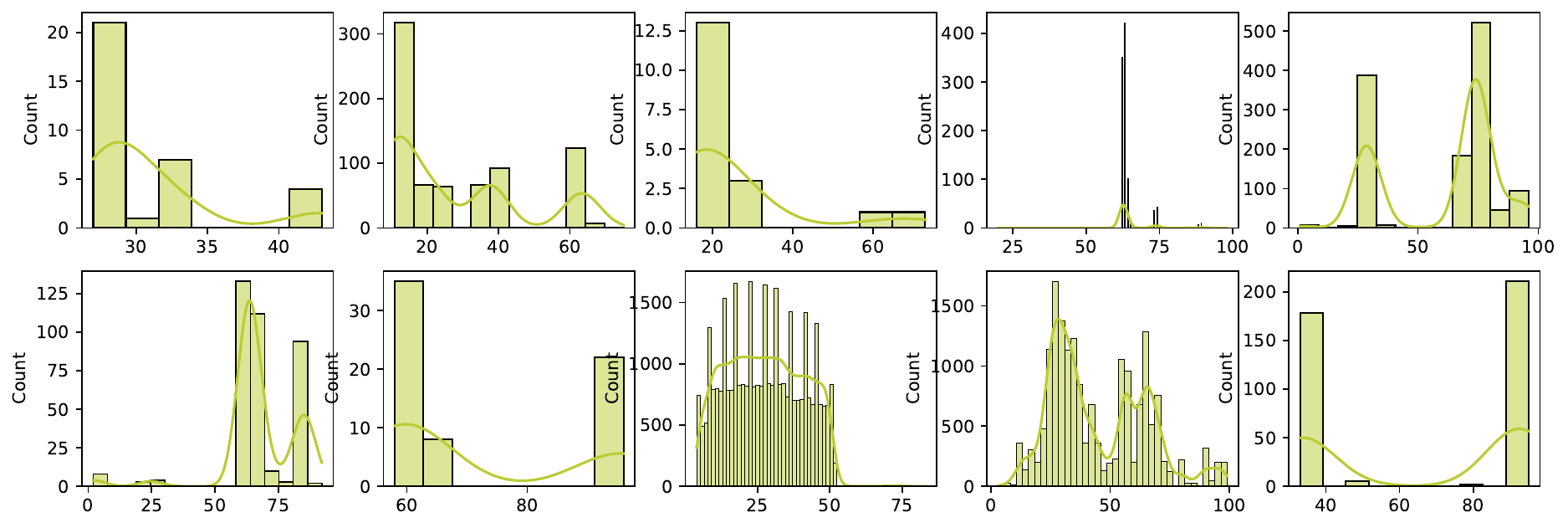}
\captionsetup{justification=centering}
\vspace{-2mm}
\caption{\small Error position distribution (FIRE)
}\label{fig_app:error_pos_hist}
\end{figure}

%% file: figures/appendix/scaling.tex
\clearpage
\newpage
\subsection{Training Digit Length Effect}
\begin{figure}[!h]
\centering
\begin{subfigure}[t]{0.45\textwidth}
\centering
\includegraphics[width=1.0\columnwidth]{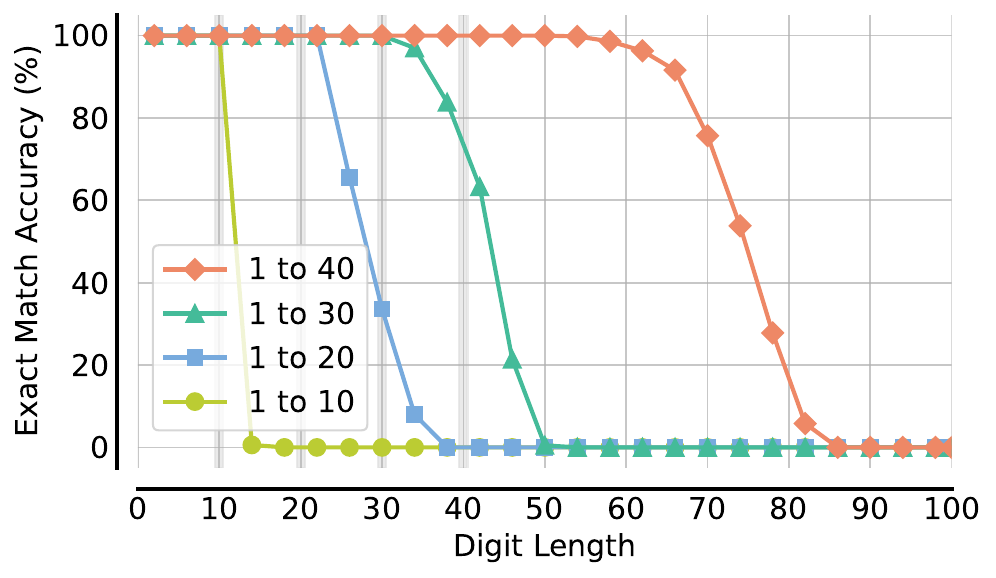}
\captionsetup{justification=centering}
\caption{EM accuracy of a 25M model using RoPE}
\end{subfigure}
\begin{subfigure}[t]{0.45\textwidth}
\centering
\includegraphics[width=1.0\columnwidth]{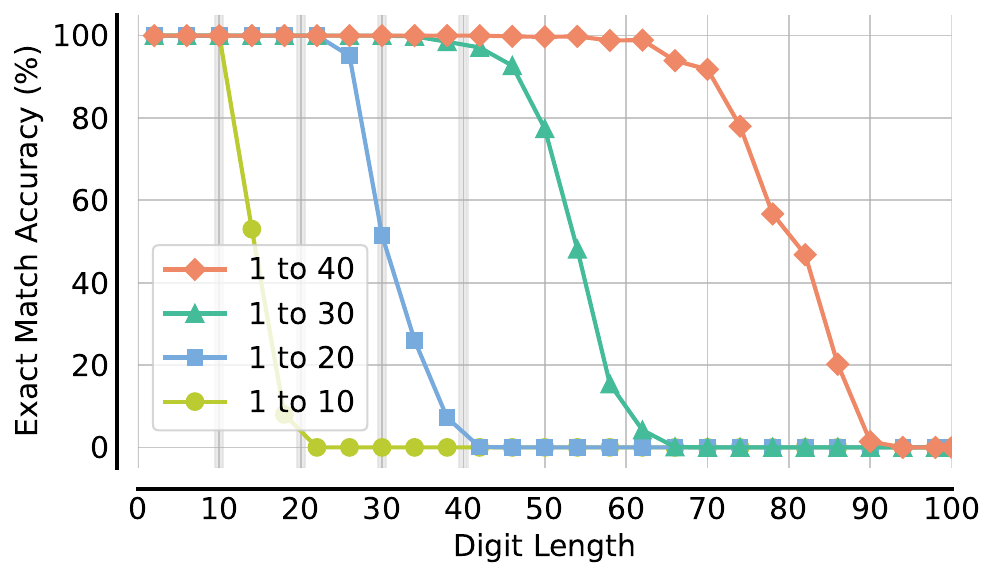}
\captionsetup{justification=centering}
\caption{EM accuracy of a 268M model using RoPE}
\end{subfigure}
\begin{subfigure}[t]{0.45\textwidth}
\centering
\includegraphics[width=1.0\columnwidth]{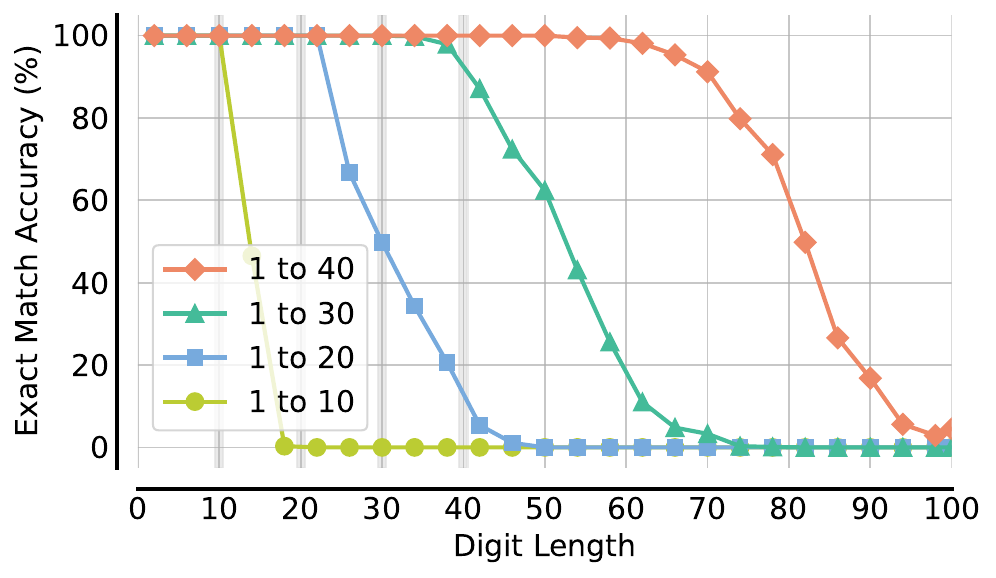}
\captionsetup{justification=centering}
\caption{EM accuracy of a 25M model using KerpleLog}
\end{subfigure}
\begin{subfigure}[t]{0.45\textwidth}
\centering
\includegraphics[width=1.0\columnwidth]{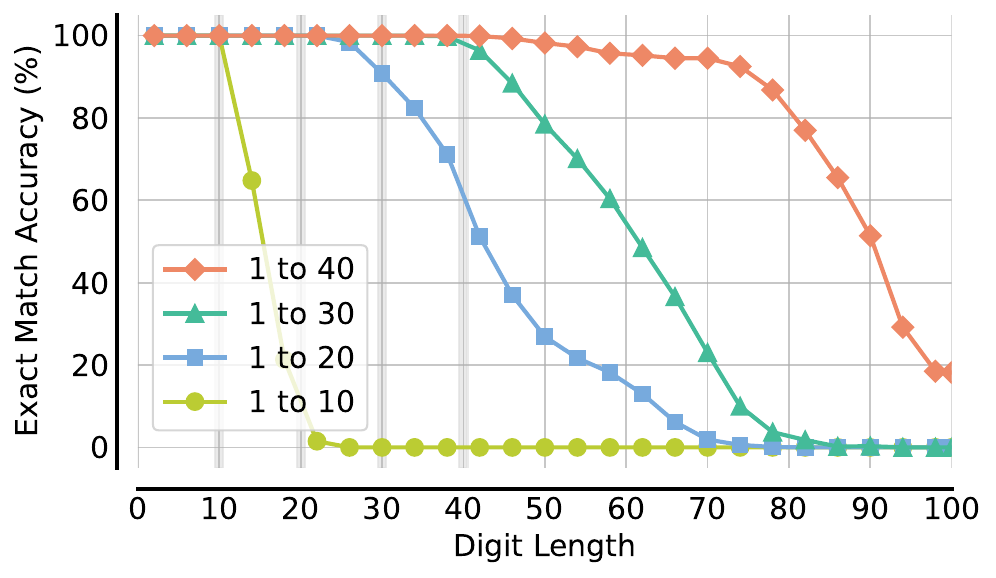}
\captionsetup{justification=centering}
\caption{EM accuracy of a 268M model using KerpleLog}
\end{subfigure}
\begin{subfigure}[t]{0.45\textwidth}
\centering
\includegraphics[width=1.0\columnwidth]{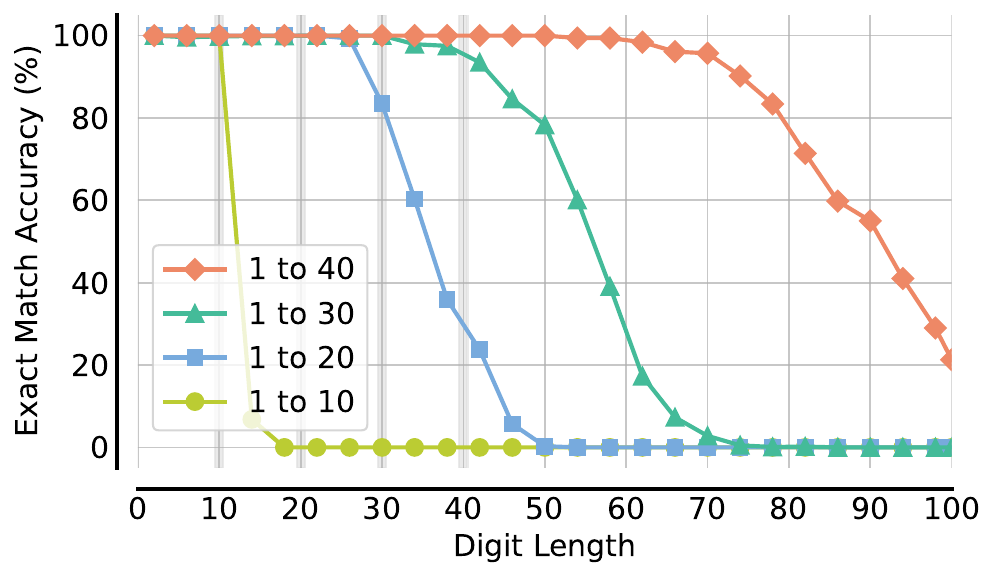}
\captionsetup{justification=centering}
\caption{EM accuracy of a 25M model using NoPE}
\end{subfigure}
\begin{subfigure}[t]{0.45\textwidth}
\centering
\includegraphics[width=1.0\columnwidth]{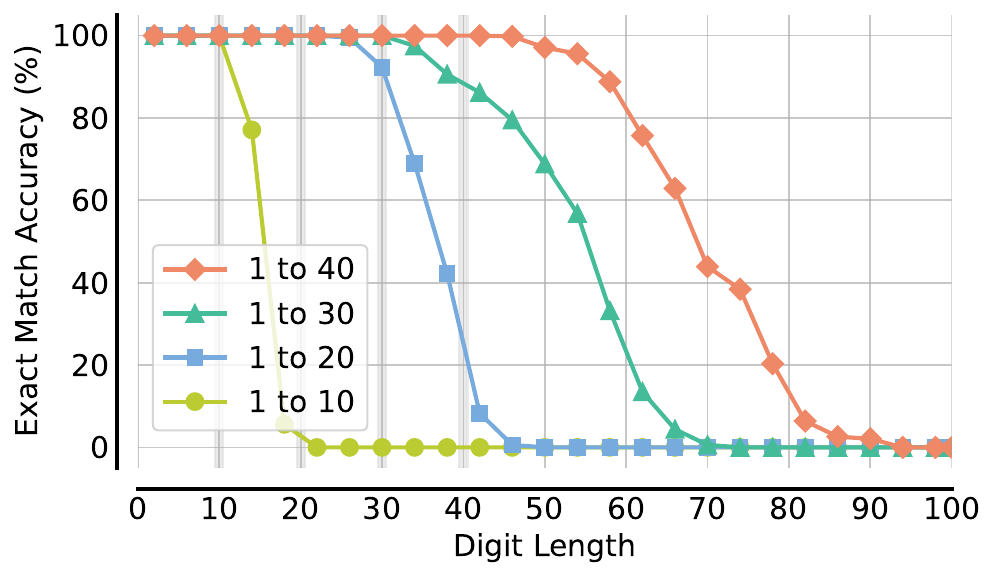}
\captionsetup{justification=centering}
\caption{EM accuracy of a 268M model using NoPE}
\end{subfigure}
\begin{subfigure}[t]{0.45\textwidth}
\centering
\includegraphics[width=1.0\columnwidth]{figures/files/em_tlen_fire_25m_max.pdf}
\captionsetup{justification=centering}
\caption{EM accuracy of a 25M model using FIRE}
\end{subfigure}
\begin{subfigure}[t]{0.45\textwidth}
\centering
\includegraphics[width=1.0\columnwidth]{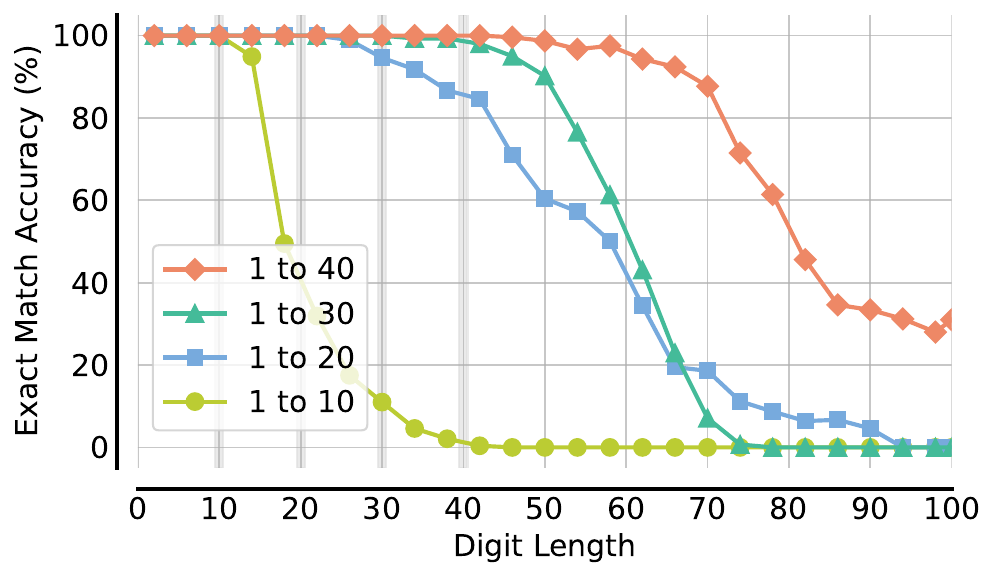}
\captionsetup{justification=centering}
\caption{EM accuracy of a 268M model using FIRE}
\end{subfigure}
\vspace{-2mm}
\caption{\small Best sequence exact match accuracy of 5 trials with two model sizes (i.e., 25M and 268M), trained on up to 10, 20, 30 and 40 digit length using 4 PEs. 
}\label{fig_app:scale_4pe}
\end{figure}

\clearpage
\newpage
\subsection{Model Size Effect}
\begin{figure}[!h]
\centering
\begin{subfigure}[t]{0.45\textwidth}
\centering
\includegraphics[width=1.0\columnwidth]{figures/files/scale_model_size_box_plot_rope.pdf}
\captionsetup{justification=centering}
\caption{RoPE}
\end{subfigure}
\begin{subfigure}[t]{0.45\textwidth}
\centering
\includegraphics[width=1.0\columnwidth]{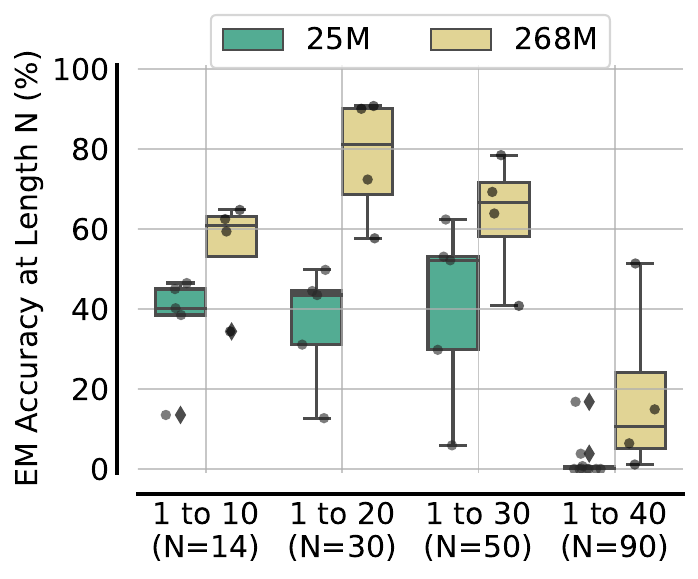}
\captionsetup{justification=centering}
\caption{KerpleLog}
\end{subfigure}
\begin{subfigure}[t]{0.45\textwidth}
\centering
\includegraphics[width=1.0\columnwidth]{figures/files/scale_model_size_box_plot_nope.pdf}
\captionsetup{justification=centering}
\caption{NoPE}
\end{subfigure}
\begin{subfigure}[t]{0.45\textwidth}
\centering
\includegraphics[width=1.0\columnwidth]{figures/files/scale_model_size_box_plot_fire.pdf}
\captionsetup{justification=centering}
\caption{FIRE}
\end{subfigure}
\vspace{-2mm}
\caption{%
Scaling model size inconsistently affects length generalization performance. While consistently enhancing performance in shorter length regimes (1-10, 1-20) across four position encodings, this trend does not hold for larger regimes (1-30, 1-40). For instance, larger models outperform smaller ones with RoPE and KerpleLog encodings, but underperform with NoPE and FIRE. Moreover, increasing model size doesn't noticeably decrease performance variance, suggesting size scaling isn't vital for length generalization.
}\label{fig_app:scaling}
\end{figure}

\clearpage
\newpage
\subsection{FIRE Related Scaling}
\begin{figure}[!h]
\centering
\begin{subfigure}[t]{0.45\textwidth}
\centering
\includegraphics[width=1.0\columnwidth]{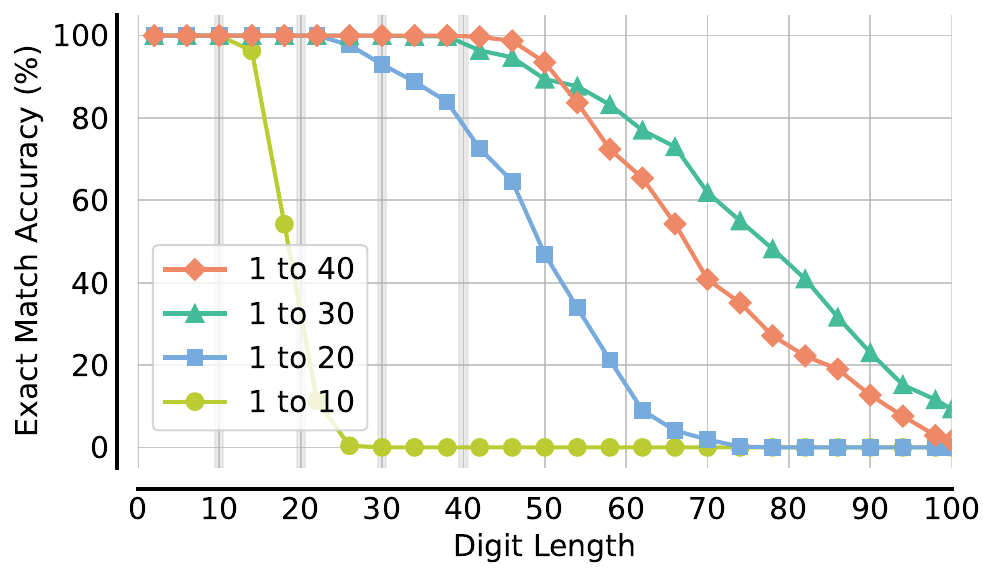}
\caption{EM accuracy of a 2M model using FIRE}
\end{subfigure}
\begin{subfigure}[t]{0.45\textwidth}
\centering
\includegraphics[width=1.0\columnwidth]{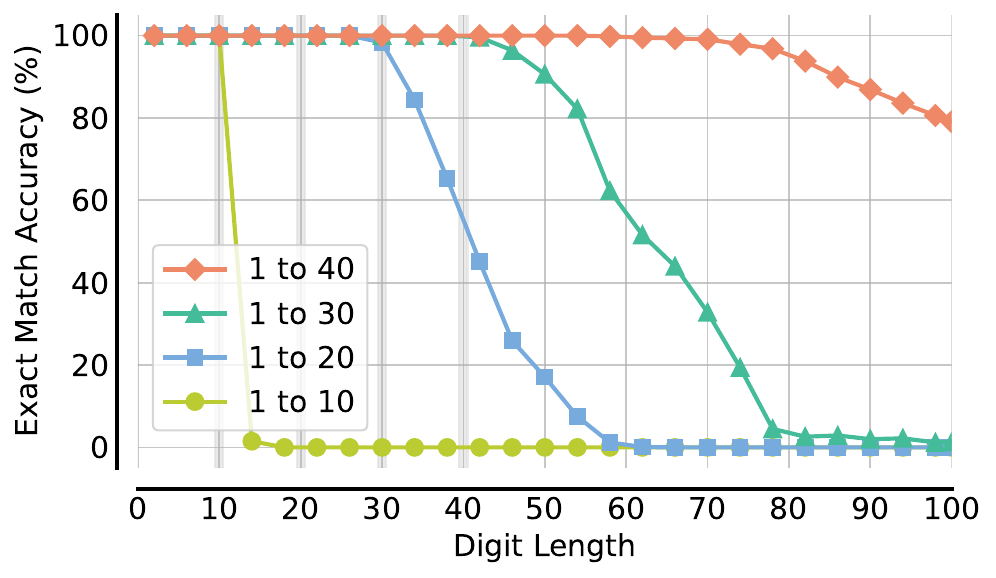}
\caption{EM accuracy of a 5M model using FIRE}
\end{subfigure}
\begin{subfigure}[t]{0.45\textwidth}
\centering
\includegraphics[width=1.0\columnwidth]{figures/files/em_tlen_fire_25m_max.pdf}
\caption{EM accuracy of a 25M model using FIRE}
\end{subfigure}
\begin{subfigure}[t]{0.45\textwidth}
\centering
\includegraphics[width=1.0\columnwidth]{figures/files/em_tlen_fire_268m_max.pdf}
\caption{EM accuracy of a 268M model using FIRE}
\end{subfigure}
\vspace{-2mm}
\caption{\small Best sequence exact match accuracy of 5 trials with four model sizes (i.e., 2M, 5M, 25M and 268M), trained on up to 10, 20, 30 and 40 digit length using \textbf{FIRE}.
}\label{fig_app:scale_fire}
\end{figure}

%% file: figures/appendix/hyperparameter_sensitivity_app.tex
\clearpage
\newpage
\subsection{Hyperparameter Study}\label{sec:hyperparam}

\begin{figure}[!h]
\centering
\includegraphics[width=0.5\columnwidth]{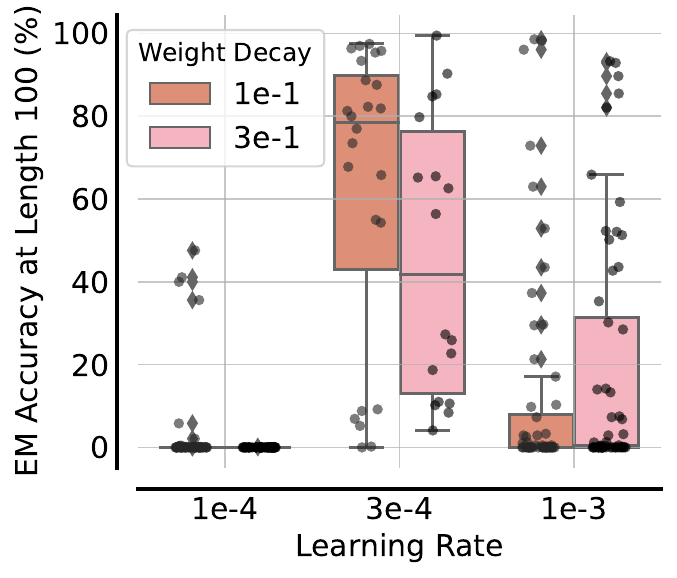}
\caption{Sequence exact match accuracy for test digit length 100, trained on digit lengths 1-40. 3e-4 seems to be the optimal learning rate.
}\label{fig_app:wd_lr}
\end{figure}

\begin{figure}[!h]
\centering
\includegraphics[width=0.9\columnwidth]{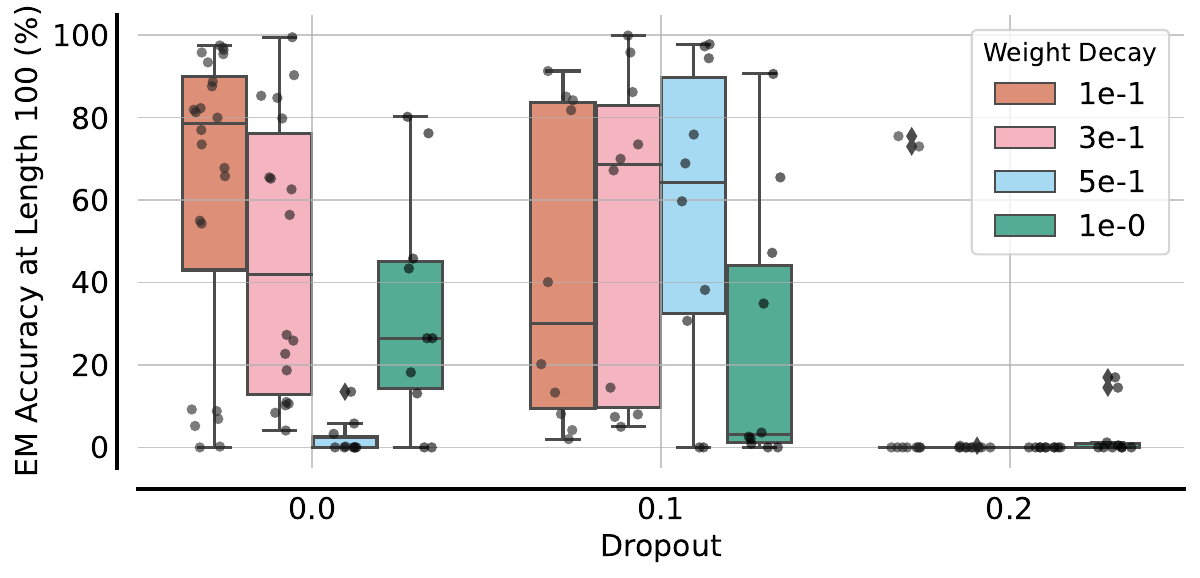}
\vspace{-2mm}
\caption{Sequence exact match accuracy for test digit length 100, trained on digit lengths 1-40. A higher dropout rate markedly impedes length generalization, whereas a lower rate shows negligible impact.
}\label{fig_app:wd_dropout}
\end{figure}

%% file: figures/appendix/loss_acc.tex
\clearpage
\newpage
\section{Training Loss and Sequence Exact Match Accuracy}\label{sec:loss_acc}

\subsection{Reverse Format without Index Hint trained up to 40-digit addition}\label{sec:loss_acc_reverse_wo_index_hint}
\begin{figure}[!h]
\centering
\begin{subfigure}[t]{0.45\textwidth}
\centering
\includegraphics[width=1.0\columnwidth]{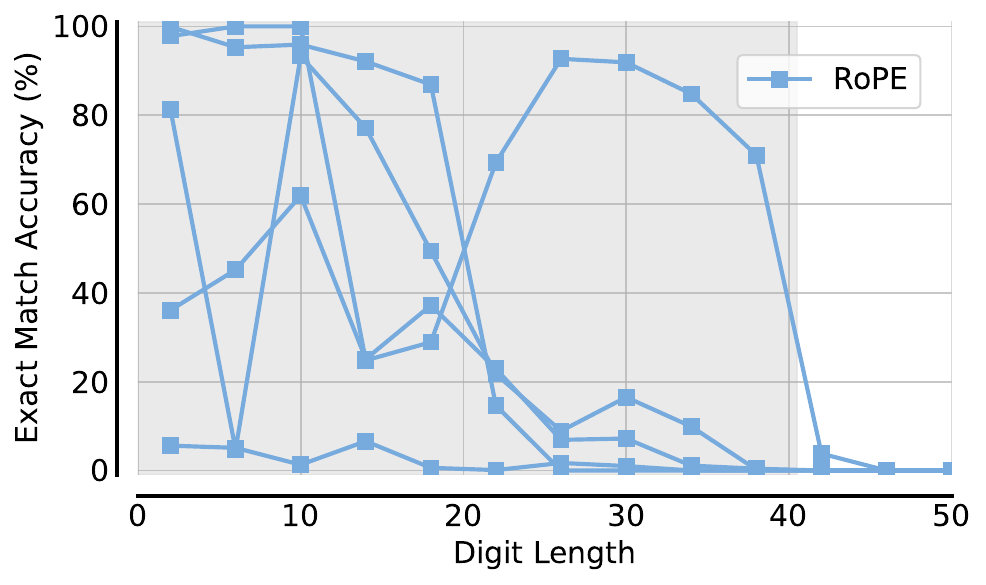}
\end{subfigure}
\begin{subfigure}[t]{0.45\textwidth}
\centering
\includegraphics[width=1.0\columnwidth]{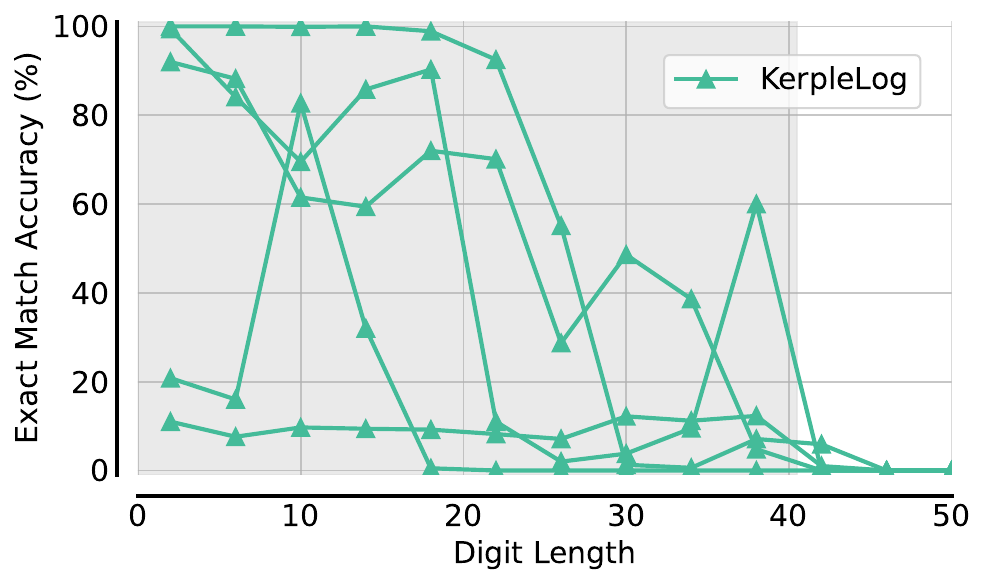}
\end{subfigure}
\begin{subfigure}[t]{0.45\textwidth}
\centering
\includegraphics[width=1.0\columnwidth]{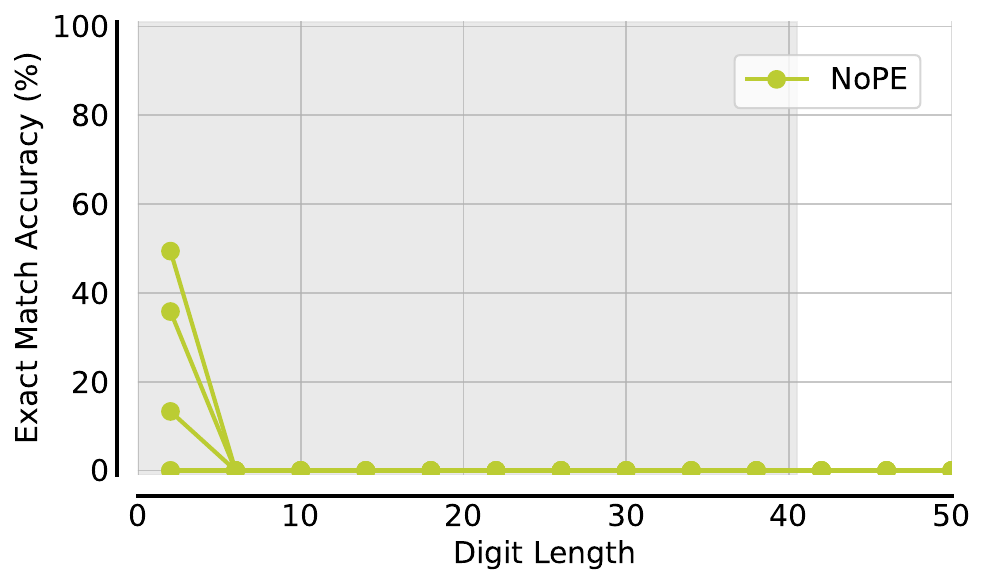}
\end{subfigure}
\begin{subfigure}[t]{0.45\textwidth}
\centering
\includegraphics[width=1.0\columnwidth]{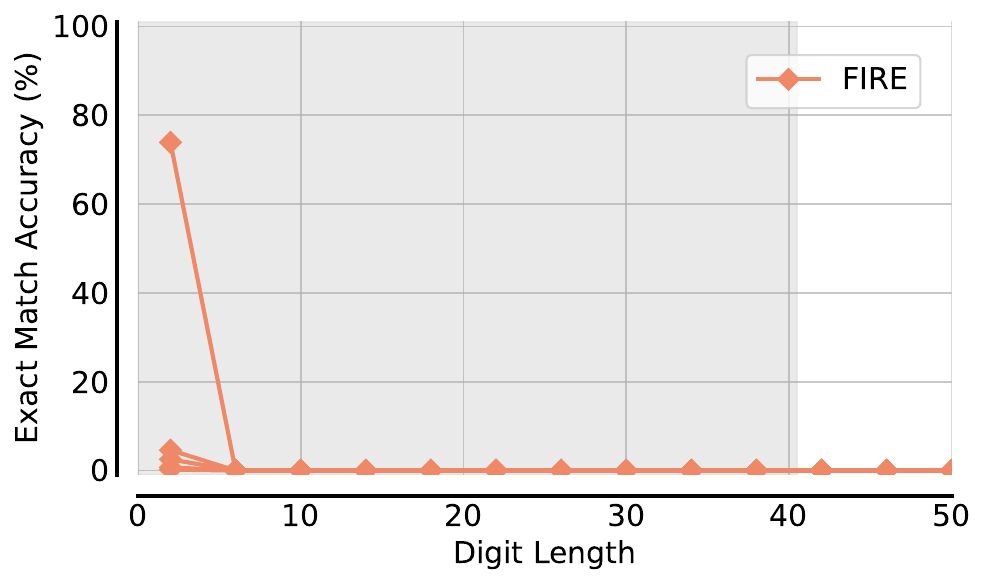}
\end{subfigure}
\vspace{-2mm}
\caption{Exact match accuracy on 20 to 100 digit addition of all 10 trials trained on up to 40-digit addition with index hint and reverse format using four different position encodings.
}\label{fig_app:seed_acc_tlen40_nopos}
\vspace{-5mm}
\end{figure}

\begin{figure}[!h]
\centering
\begin{subfigure}[t]{0.45\textwidth}
\centering
\includegraphics[width=1.0\columnwidth]{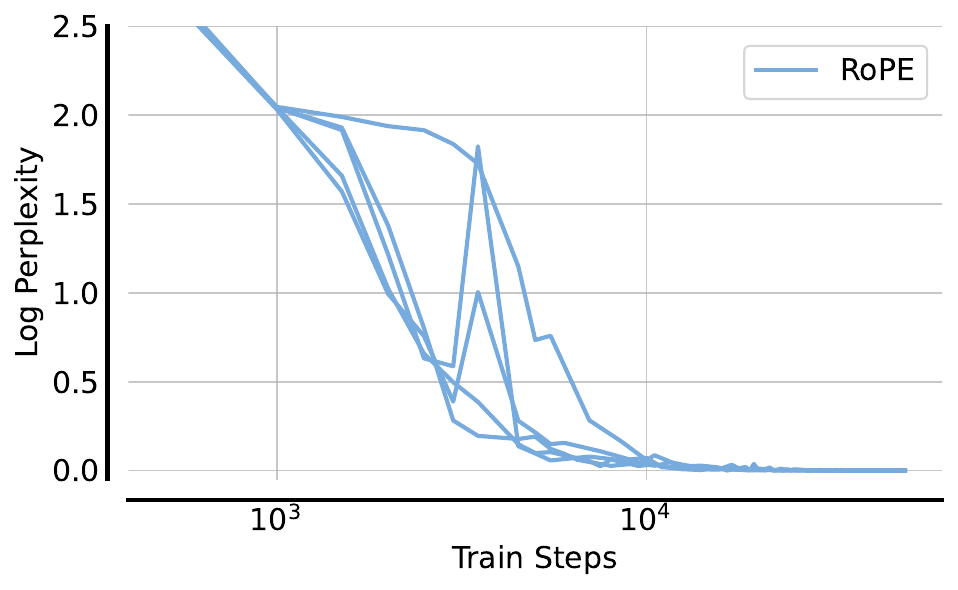}
\end{subfigure}
\begin{subfigure}[t]{0.45\textwidth}
\centering
\includegraphics[width=1.0\columnwidth]{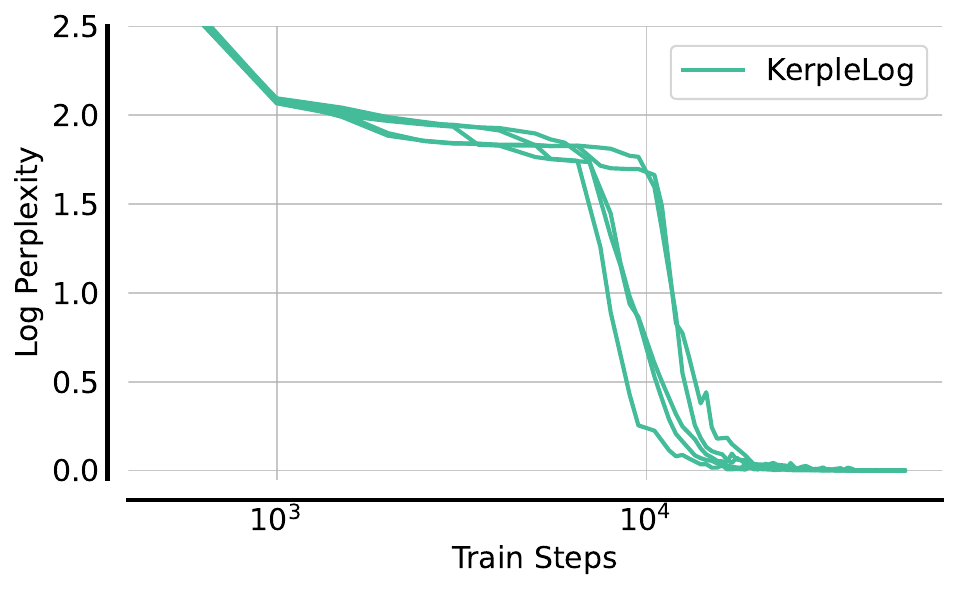}
\end{subfigure}
\begin{subfigure}[t]{0.45\textwidth}
\centering
\includegraphics[width=1.0\columnwidth]{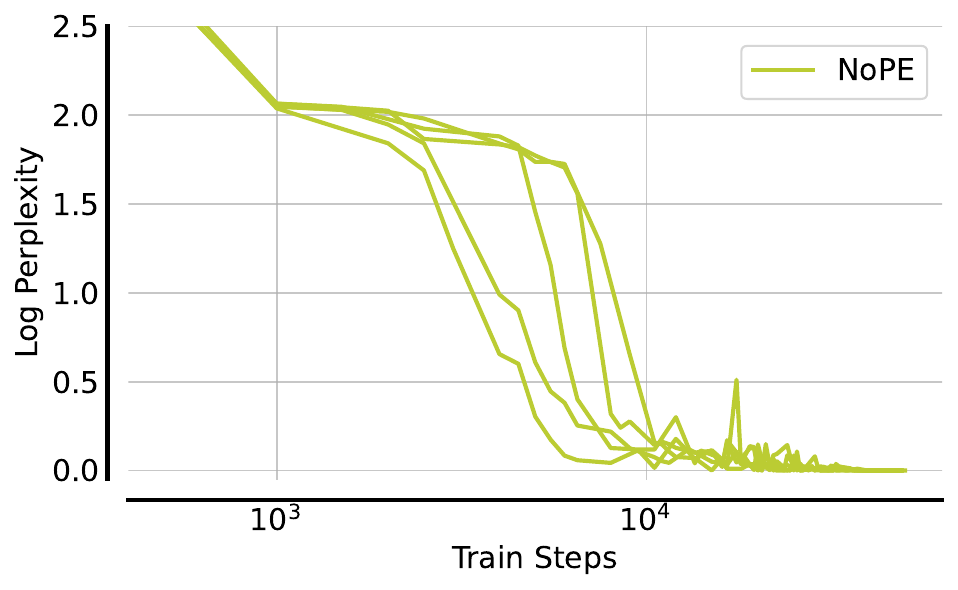}
\end{subfigure}
\begin{subfigure}[t]{0.45\textwidth}
\centering
\includegraphics[width=1.0\columnwidth]{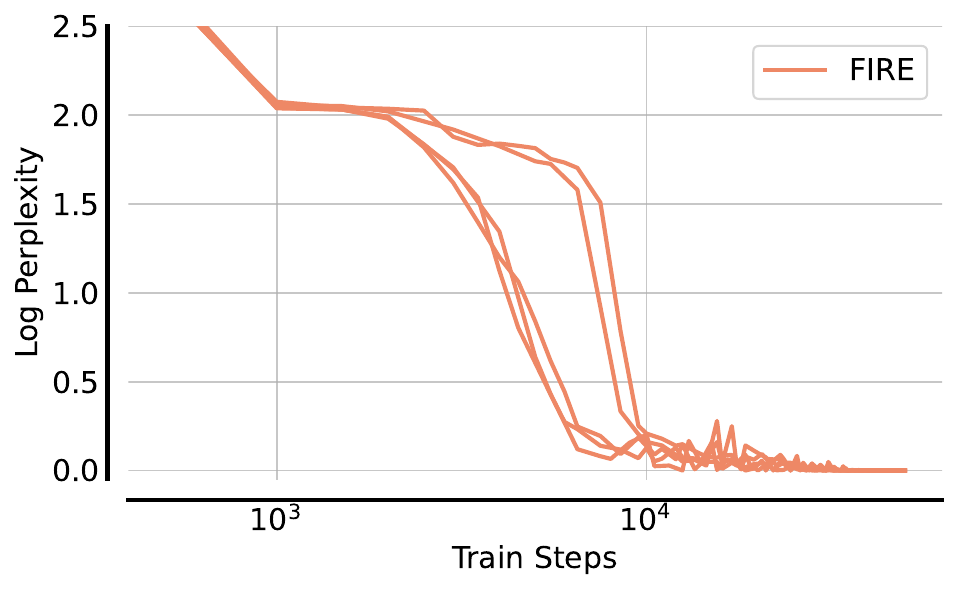}
\end{subfigure}
\vspace{-2mm}
\caption{Training loss over 10 trials in reverse formats. Despite similar nearly 0 log perplexity losses across runs after 10K training steps, different runs exhibit very different length generalization.
}\label{fig_app:logpplx_logx_all_nopos_tlen40}
\end{figure}

\clearpage
\newpage
\subsection{Reverse Format without Index Hint trained up to 10-digit addition}\label{sec:loss_acc_reverse_wo_index_hint_10}
\begin{figure}[!h]
\centering
\begin{subfigure}[t]{0.45\textwidth}
\centering
\includegraphics[width=1.0\columnwidth]{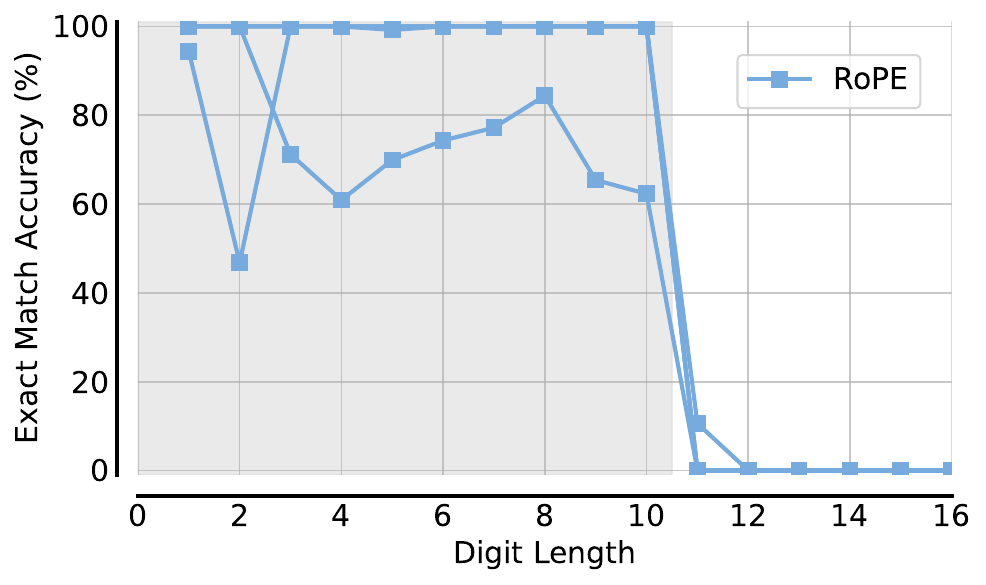}
\end{subfigure}
\begin{subfigure}[t]{0.45\textwidth}
\centering
\includegraphics[width=1.0\columnwidth]{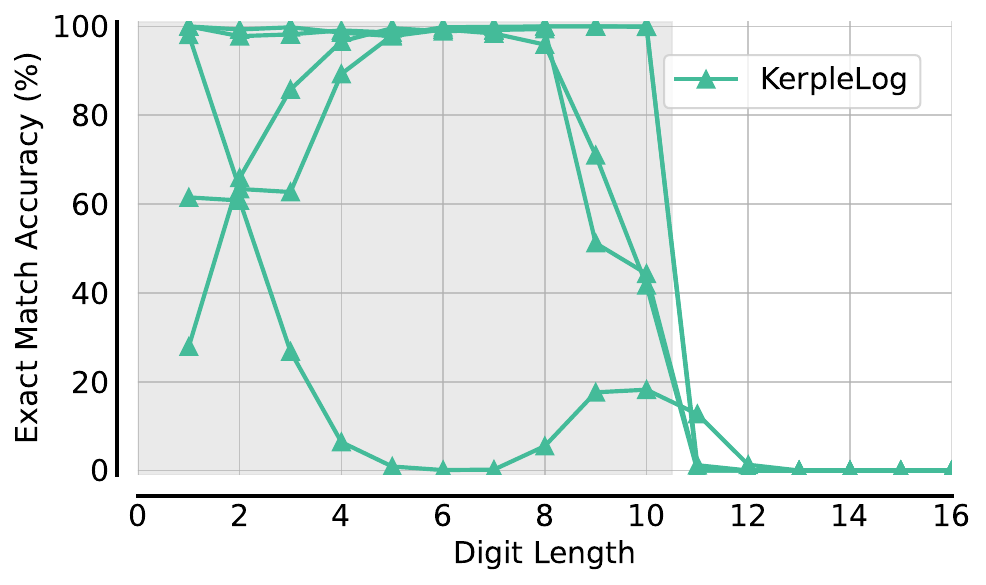}
\end{subfigure}
\begin{subfigure}[t]{0.45\textwidth}
\centering
\includegraphics[width=1.0\columnwidth]{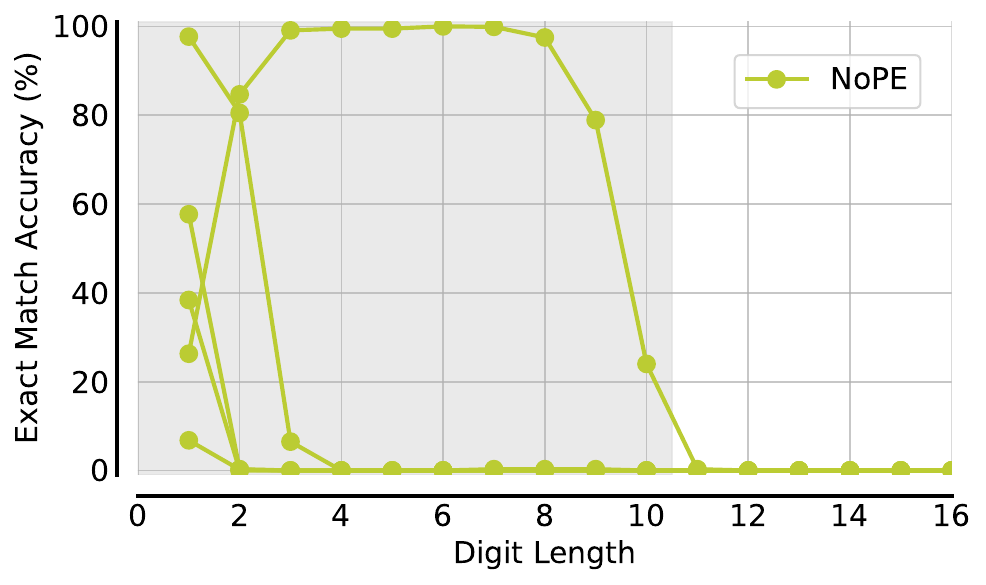}
\end{subfigure}
\begin{subfigure}[t]{0.45\textwidth}
\centering
\includegraphics[width=1.0\columnwidth]{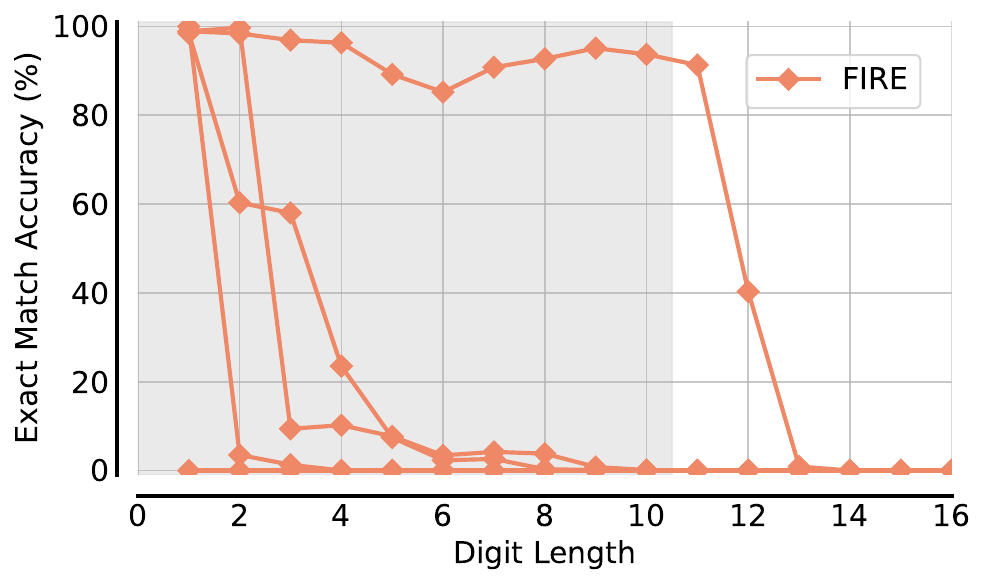}
\end{subfigure}
\vspace{-2mm}
\caption{Exact match accuracy on 20 to 100 digit addition of all 10 trials trained on up to 10-digit addition with index hint and reverse format using four different position encodings.
}\label{fig_app:seed_acc_tlen10_nopos}
\end{figure}

\begin{figure}[!h]
\centering
\begin{subfigure}[t]{0.45\textwidth}
\centering
\includegraphics[width=1.0\columnwidth]{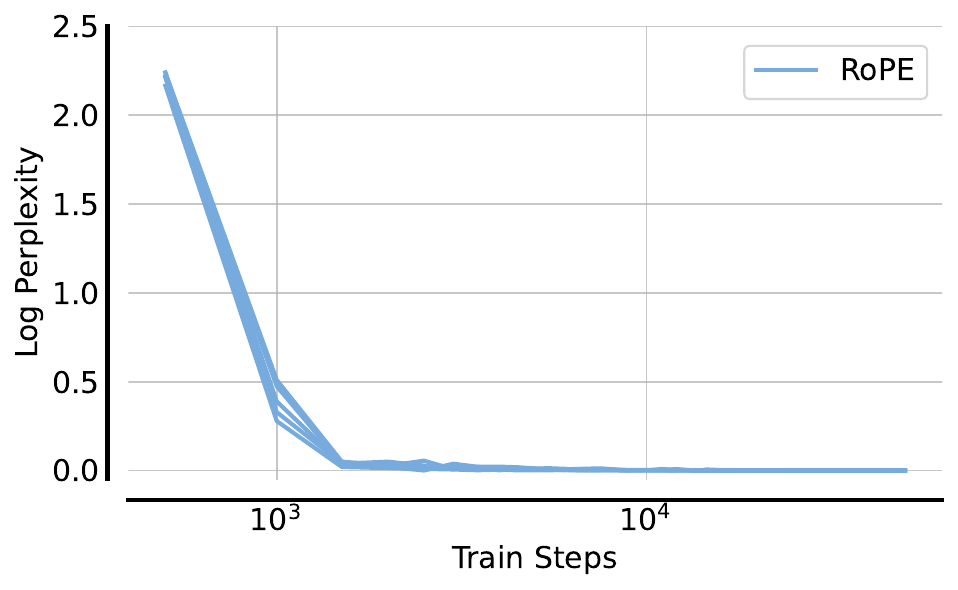}
\end{subfigure}
\begin{subfigure}[t]{0.45\textwidth}
\centering
\includegraphics[width=1.0\columnwidth]{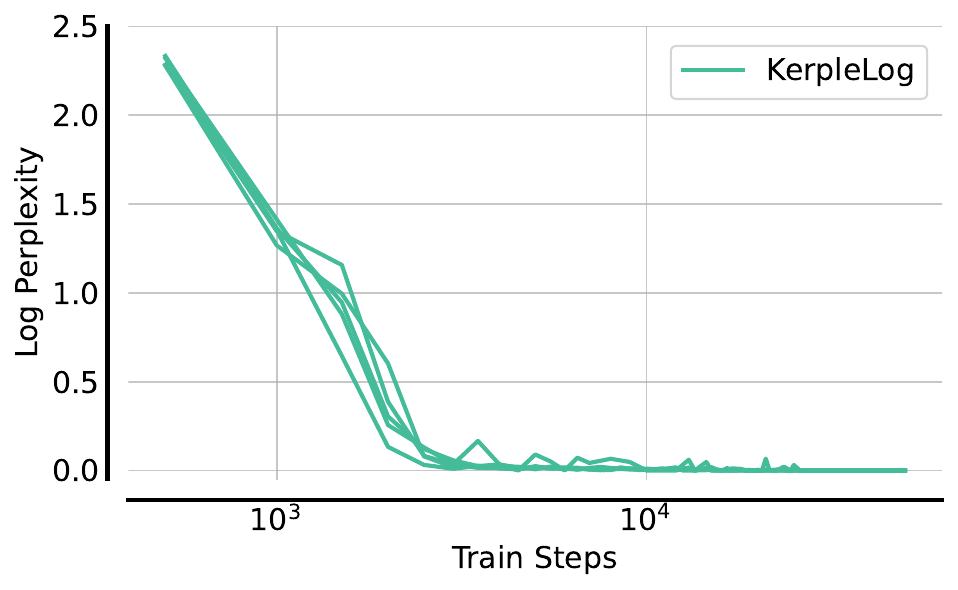}
\end{subfigure}
\begin{subfigure}[t]{0.45\textwidth}
\centering
\includegraphics[width=1.0\columnwidth]{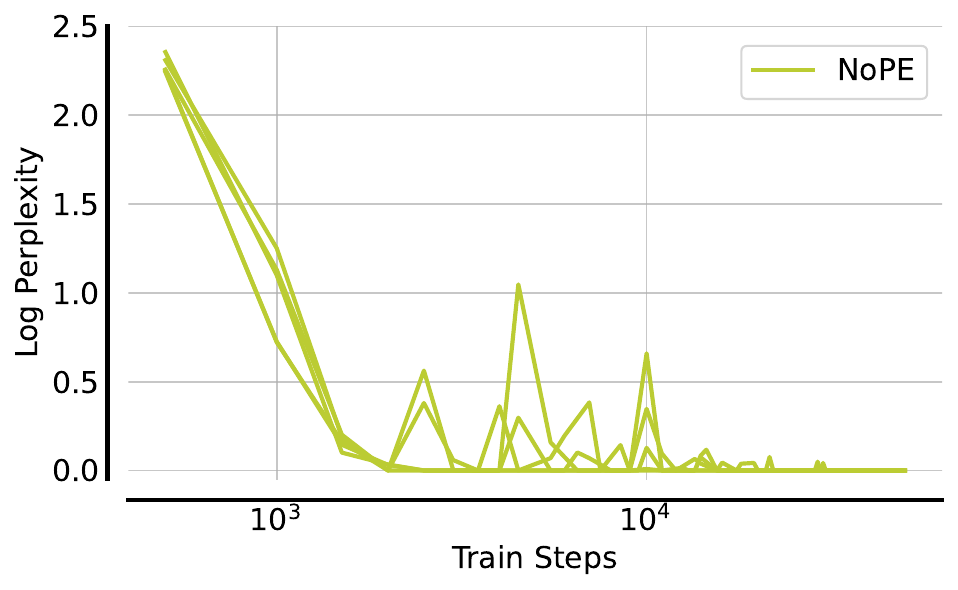}
\end{subfigure}
\begin{subfigure}[t]{0.45\textwidth}
\centering
\includegraphics[width=1.0\columnwidth]{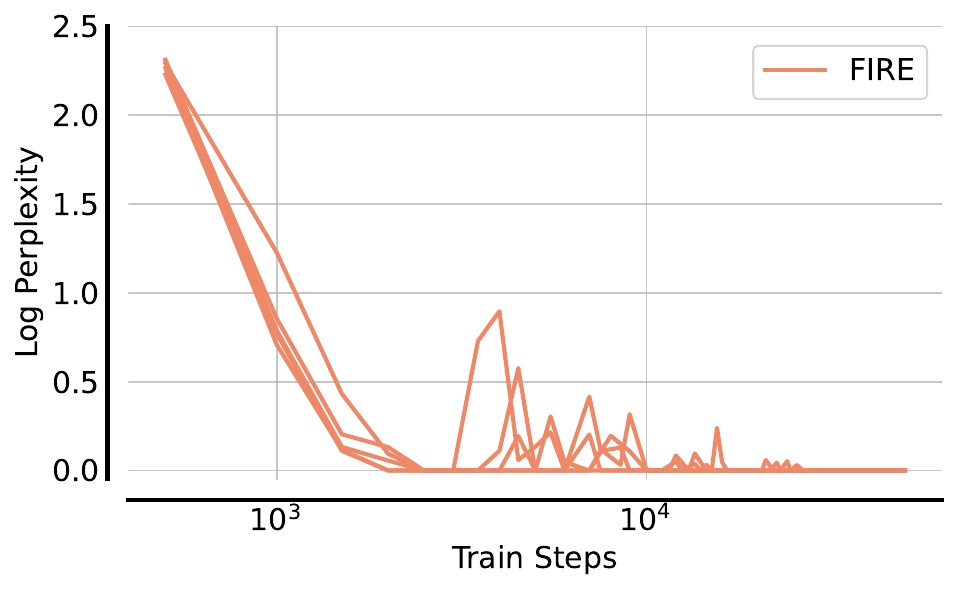}
\end{subfigure}
\vspace{-2mm}
\caption{Training loss over 10 trials in reverse formats. Despite similar nearly 0 log perplexity losses across runs after 10K training steps, different runs exhibit very different length generalization.
}\label{fig_app:logpplx_logx_all_nopos_tlen10}
\end{figure}

\clearpage
\newpage
\subsection{Standard Format with Index Hint trained up to 40}\label{sec:loss_acc_standard_index_hint}
\begin{figure}[!h]
\centering
\begin{subfigure}[t]{0.45\textwidth}
\centering
\includegraphics[width=1.0\columnwidth]{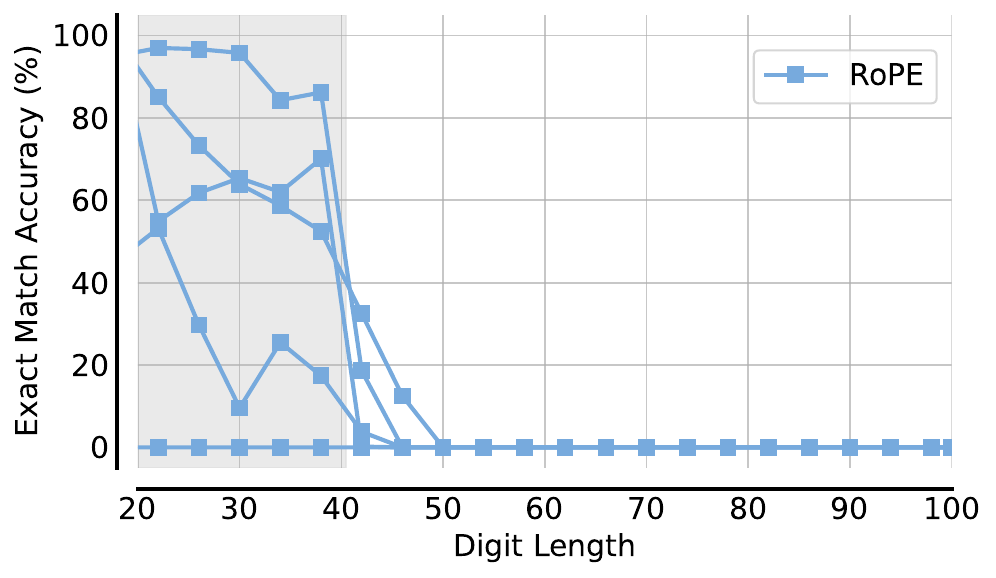}
\end{subfigure}
\begin{subfigure}[t]{0.45\textwidth}
\centering
\includegraphics[width=1.0\columnwidth]{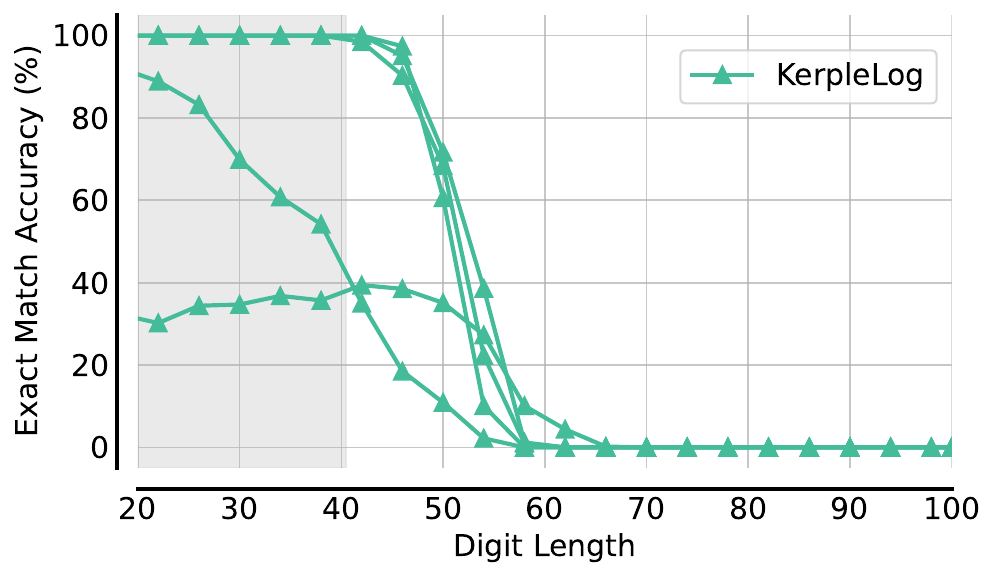}
\end{subfigure}
\begin{subfigure}[t]{0.45\textwidth}
\centering
\includegraphics[width=1.0\columnwidth]{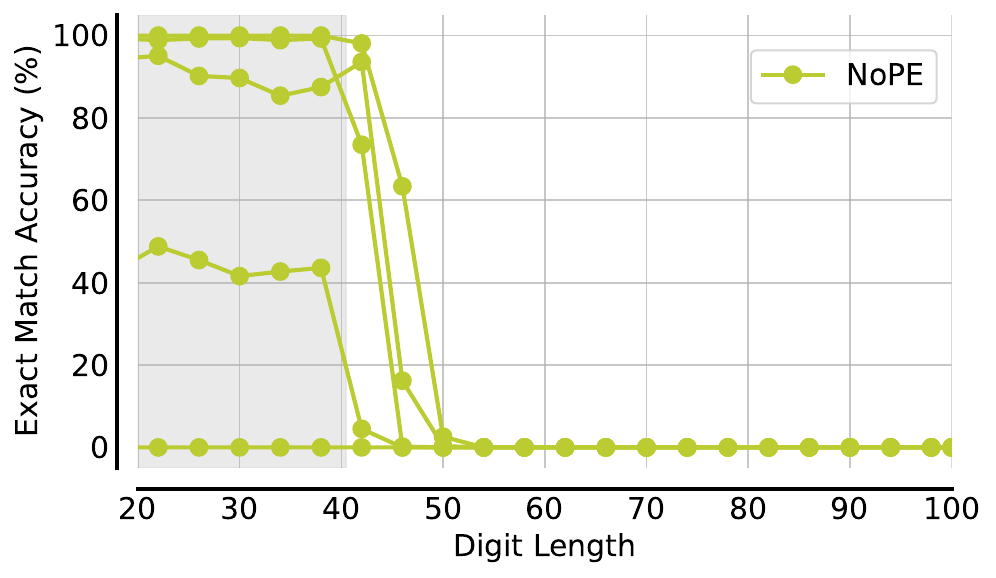}
\end{subfigure}
\begin{subfigure}[t]{0.45\textwidth}
\centering
\includegraphics[width=1.0\columnwidth]{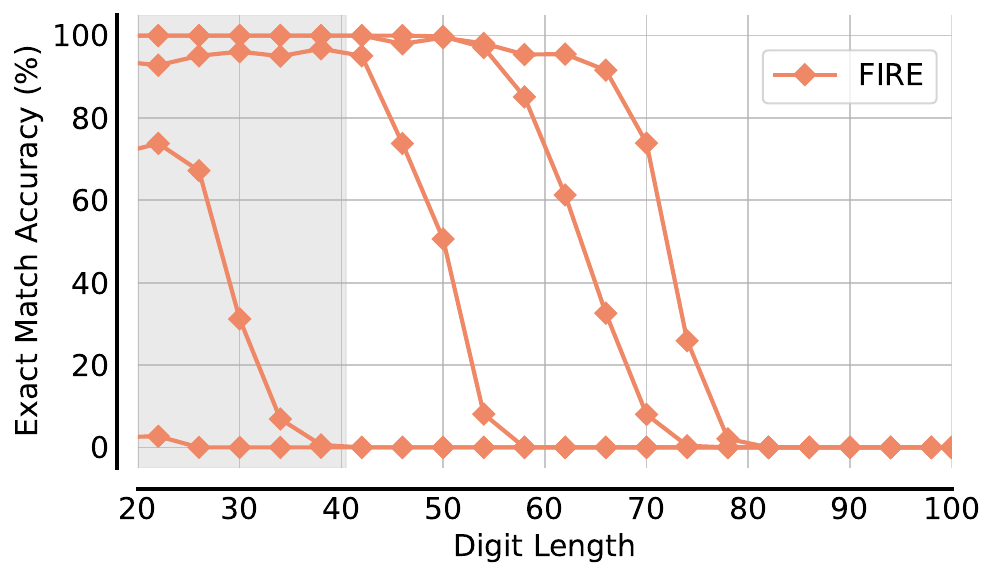}
\end{subfigure}
\vspace{-2mm}
\caption{Exact match accuracy on 20 to 100 digit addition of all 10 trials trained on up to 40-digit addition with index hint and reverse format using four different position encodings.
}\label{fig_app:seed_acc_tlen40_base}
\end{figure}

\begin{figure}[!h]
\centering
\begin{subfigure}[t]{0.45\textwidth}
\centering
\includegraphics[width=1.0\columnwidth]{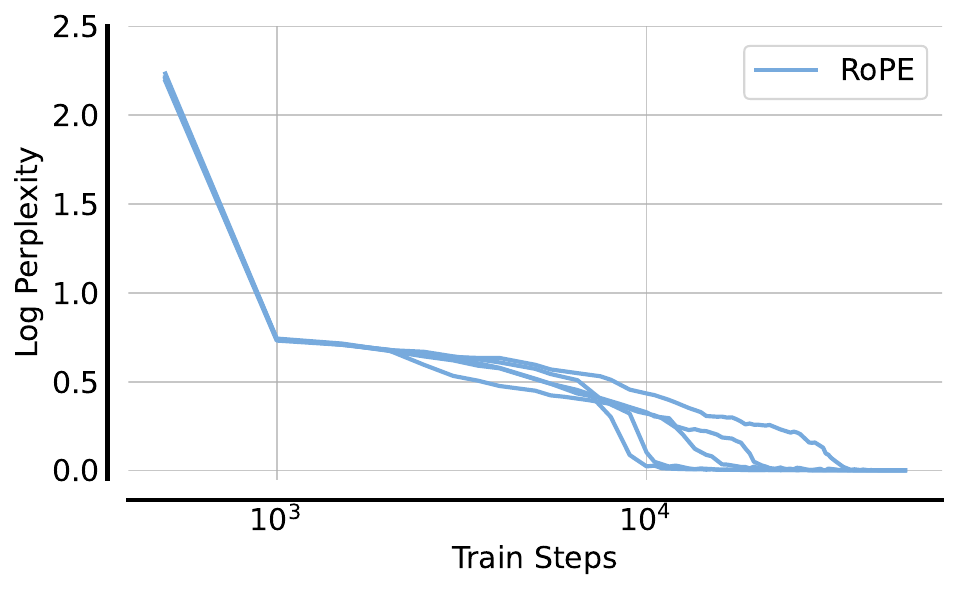}
\end{subfigure}
\begin{subfigure}[t]{0.45\textwidth}
\centering
\includegraphics[width=1.0\columnwidth]{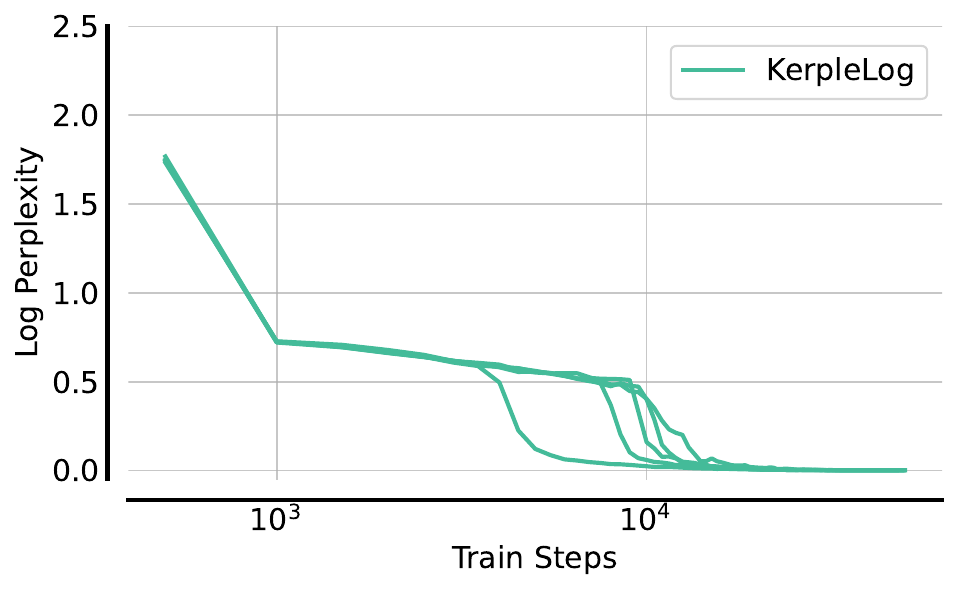}
\end{subfigure}
\begin{subfigure}[t]{0.45\textwidth}
\centering
\includegraphics[width=1.0\columnwidth]{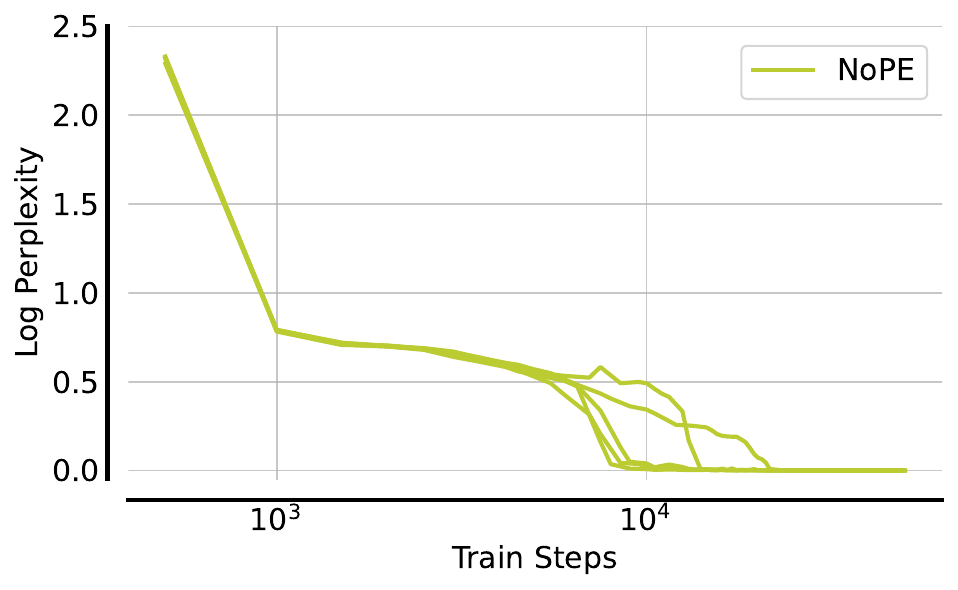}
\end{subfigure}
\begin{subfigure}[t]{0.45\textwidth}
\centering
\includegraphics[width=1.0\columnwidth]{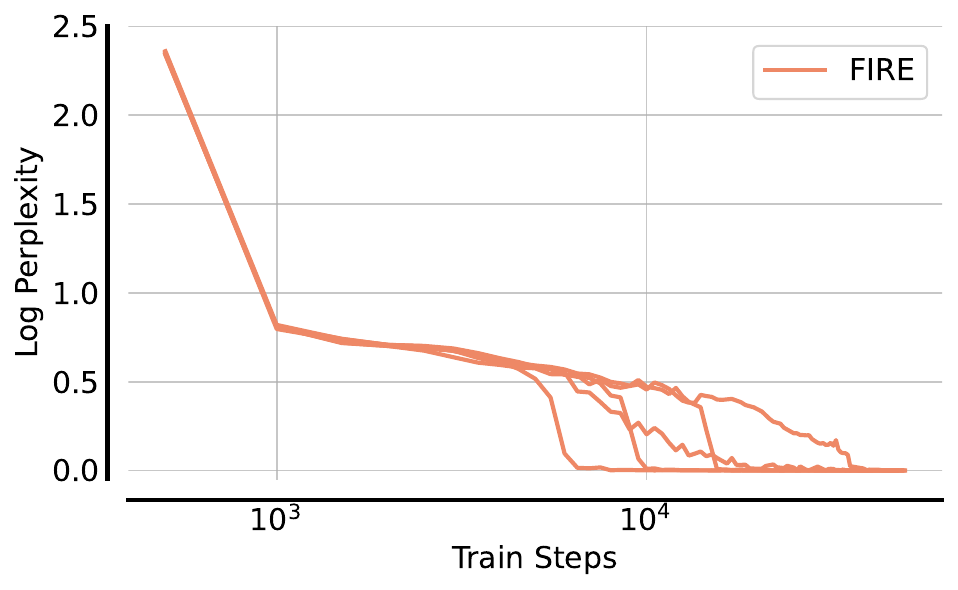}
\end{subfigure}
\vspace{-2mm}
\caption{Training loss over 10 trials in reverse formats. Despite similar nearly 0 log perplexity losses across runs after 10K training steps, different runs exhibit very different length generalization.
}\label{fig_app:logpplx_logx_all_base}
\end{figure}

\clearpage
\newpage
\subsection{Random Space Augmentation with Reverse Format with Index Hint trained up to 40-digit addition}\label{sec:loss_acc_reverse_index_hint_rs}
\begin{figure}[!h]
\centering
\begin{subfigure}[t]{0.45\textwidth}
\centering
\includegraphics[width=1.0\columnwidth]{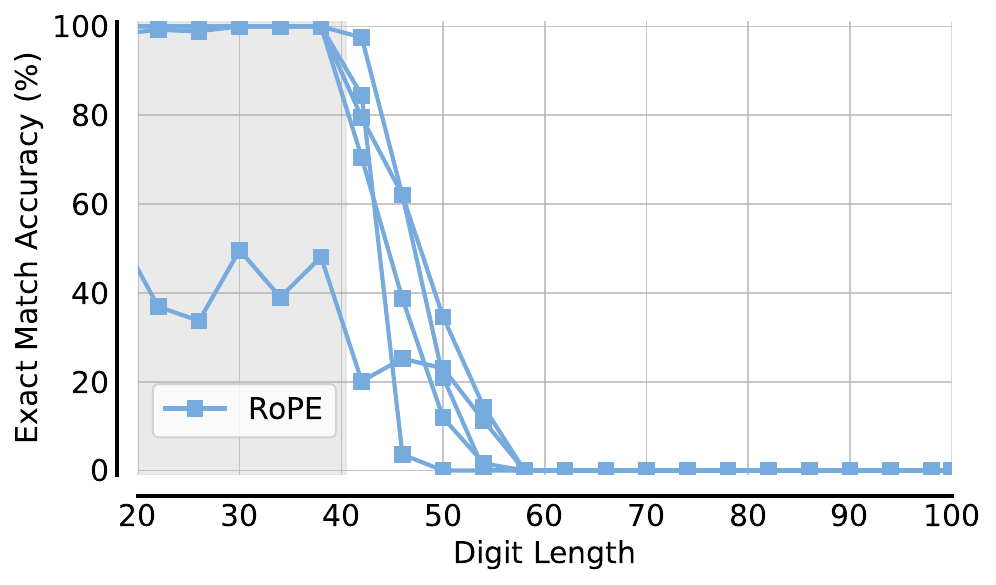}
\end{subfigure}
\begin{subfigure}[t]{0.45\textwidth}
\centering
\includegraphics[width=1.0\columnwidth]{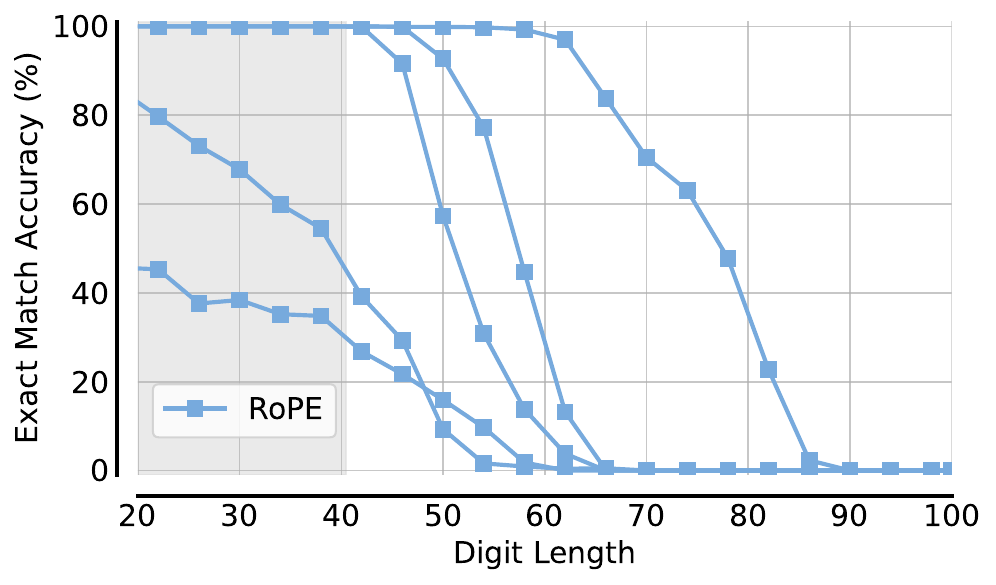}
\end{subfigure}
\begin{subfigure}[t]{0.45\textwidth}
\centering
\includegraphics[width=1.0\columnwidth]{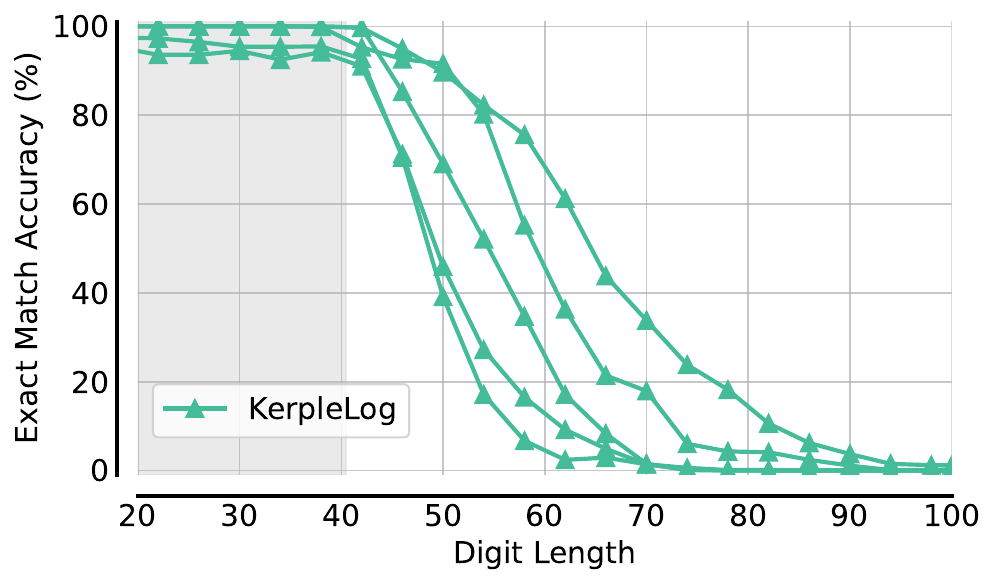}
\end{subfigure}
\begin{subfigure}[t]{0.45\textwidth}
\centering
\includegraphics[width=1.0\columnwidth]{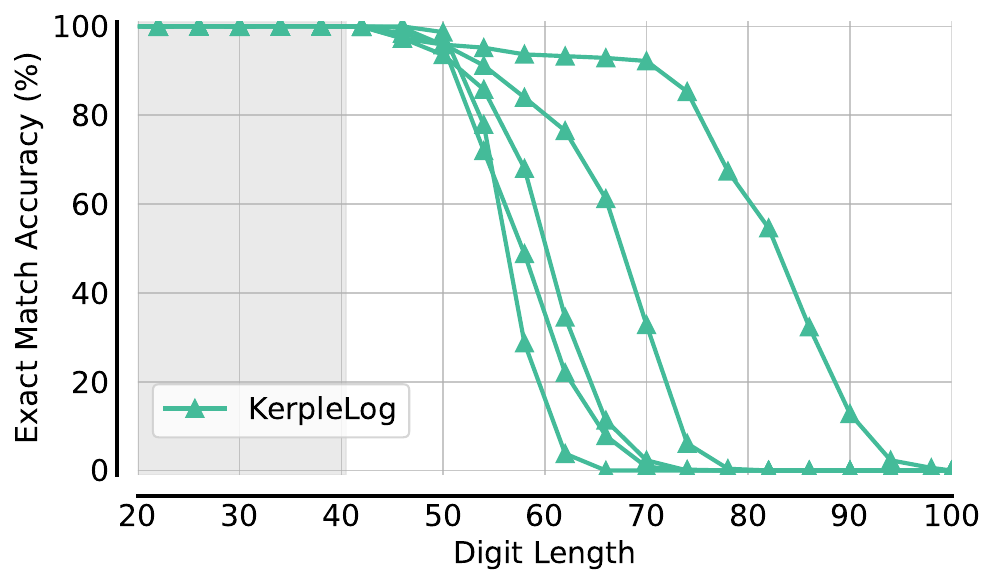}
\end{subfigure}
\begin{subfigure}[t]{0.45\textwidth}
\centering
\includegraphics[width=1.0\columnwidth]{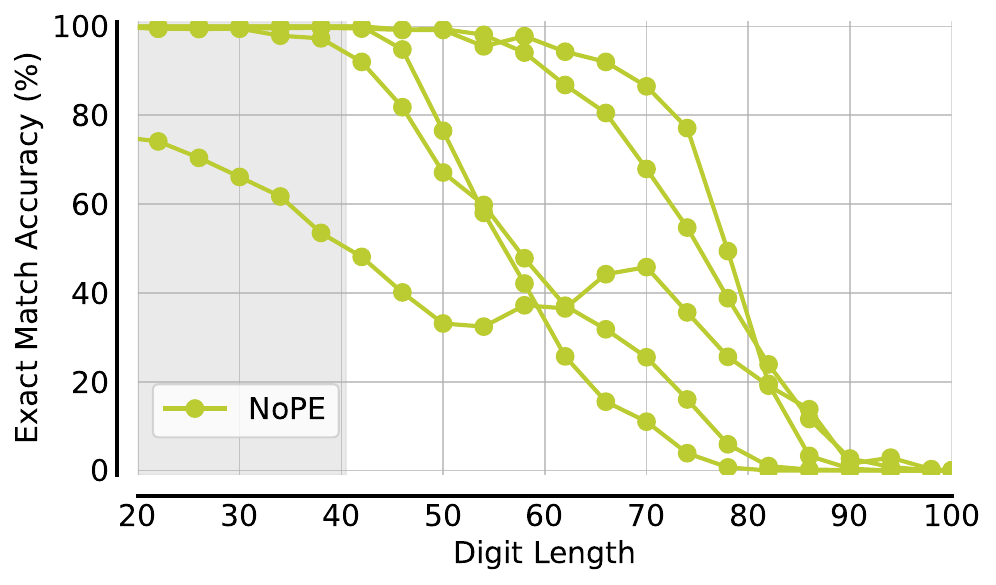}
\end{subfigure}
\begin{subfigure}[t]{0.45\textwidth}
\centering
\includegraphics[width=1.0\columnwidth]{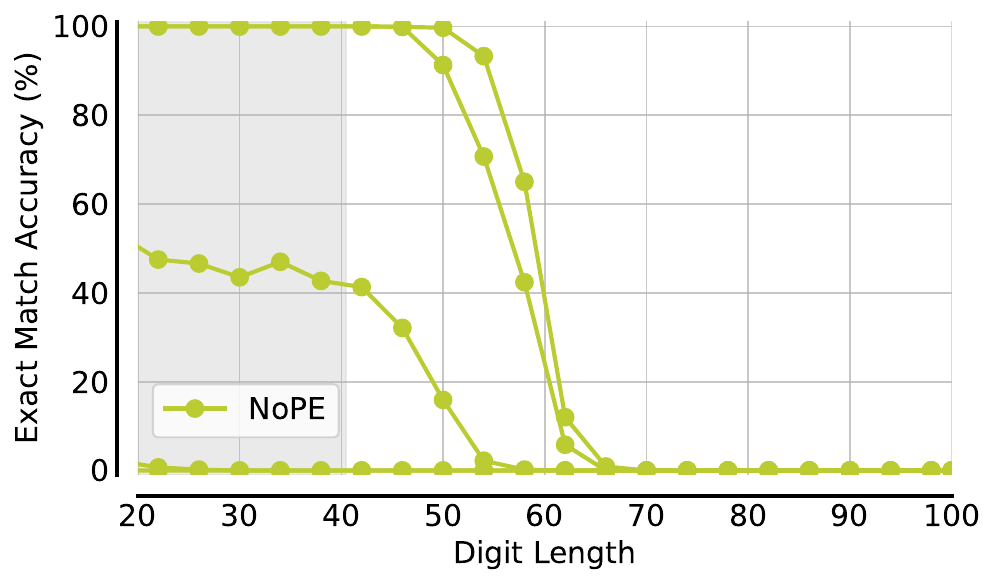}
\end{subfigure}
\begin{subfigure}[t]{0.45\textwidth}
\centering
\includegraphics[width=1.0\columnwidth]{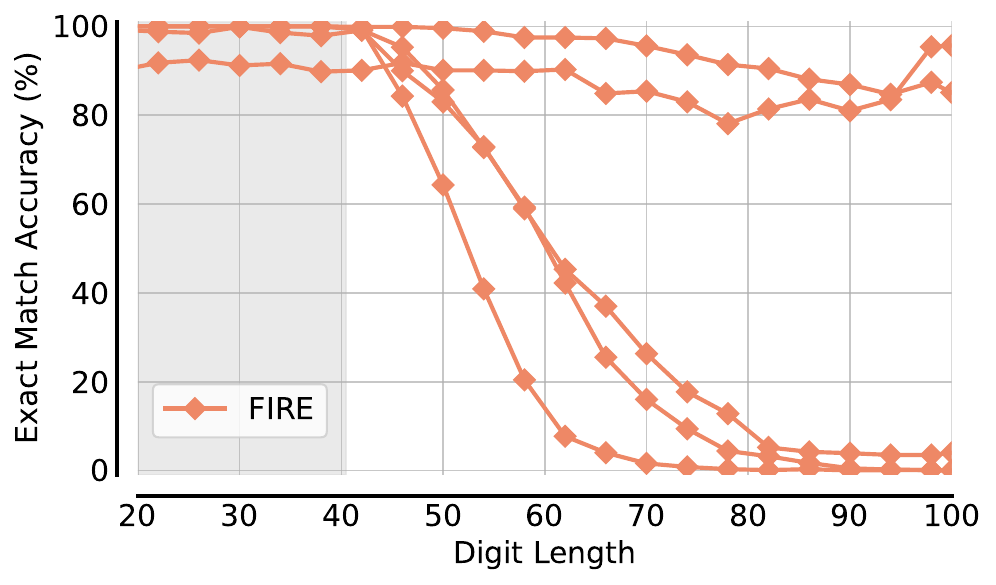}
\end{subfigure}
\begin{subfigure}[t]{0.45\textwidth}
\centering
\includegraphics[width=1.0\columnwidth]{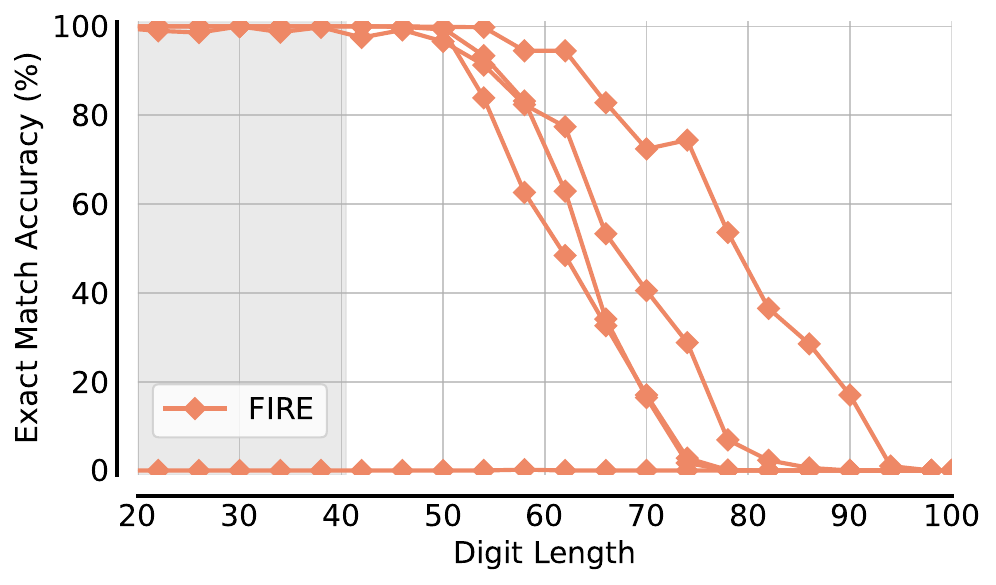}
\end{subfigure}
\vspace{-2mm}
\caption{(Left) With Random Space Augmentation. (Right) Without Random Space Augmentation. Exact match accuracy on 20 to 100 digit addition of all 10 trials trained on up to 40-digit addition with index hint and reverse format using four different position encodings.
}\label{fig_app:seed_acc_tlen40_rs}
\end{figure}

\begin{figure}[!h]
\centering
\begin{subfigure}[t]{0.45\textwidth}
\centering
\includegraphics[width=1.0\columnwidth]{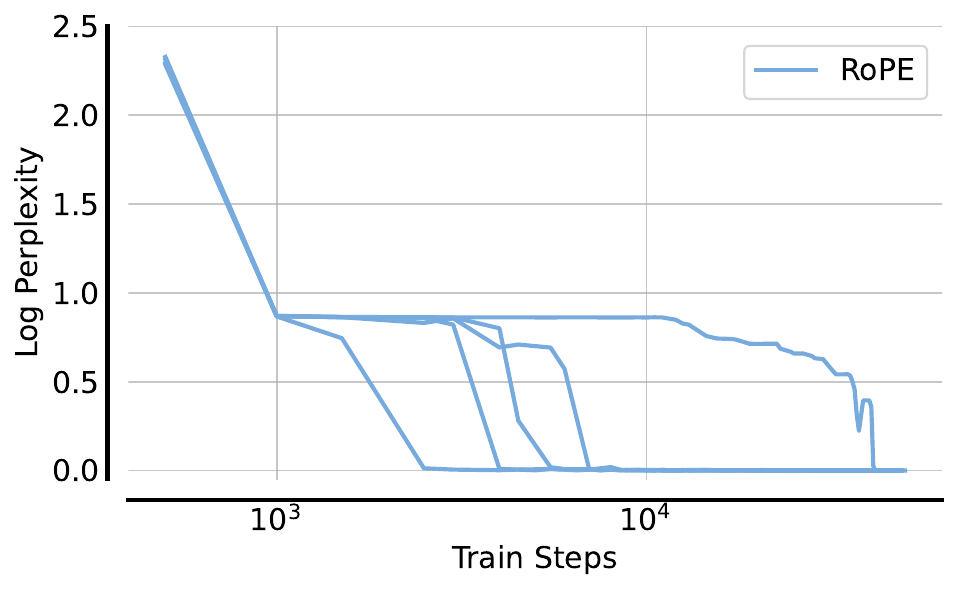}
\end{subfigure}
\begin{subfigure}[t]{0.45\textwidth}
\centering
\includegraphics[width=1.0\columnwidth]{figures/files/grokking_reverse_rs_rope_false_log}
\end{subfigure}
\begin{subfigure}[t]{0.45\textwidth}
\centering
\includegraphics[width=1.0\columnwidth]{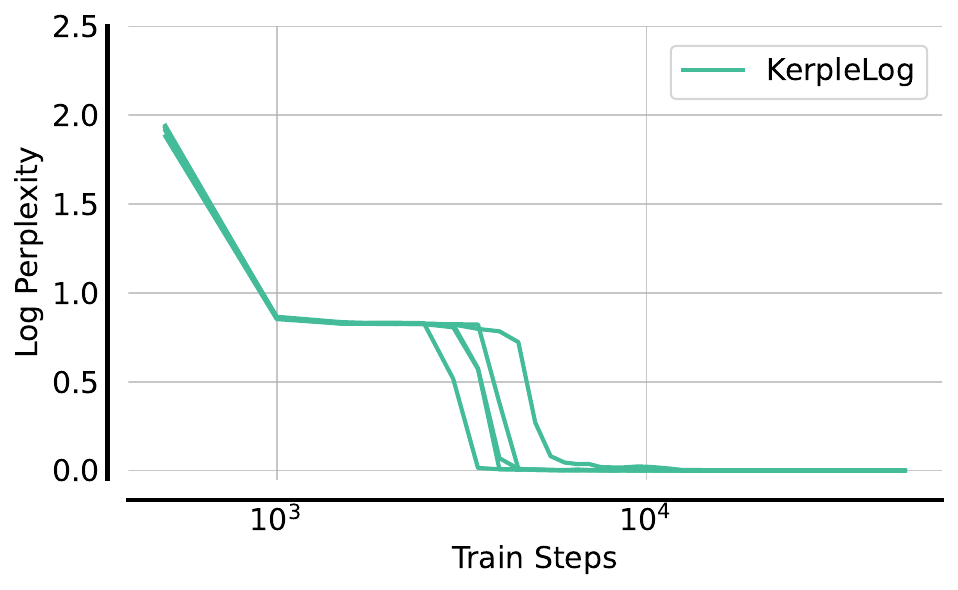}
\end{subfigure}
\begin{subfigure}[t]{0.45\textwidth}
\centering
\includegraphics[width=1.0\columnwidth]{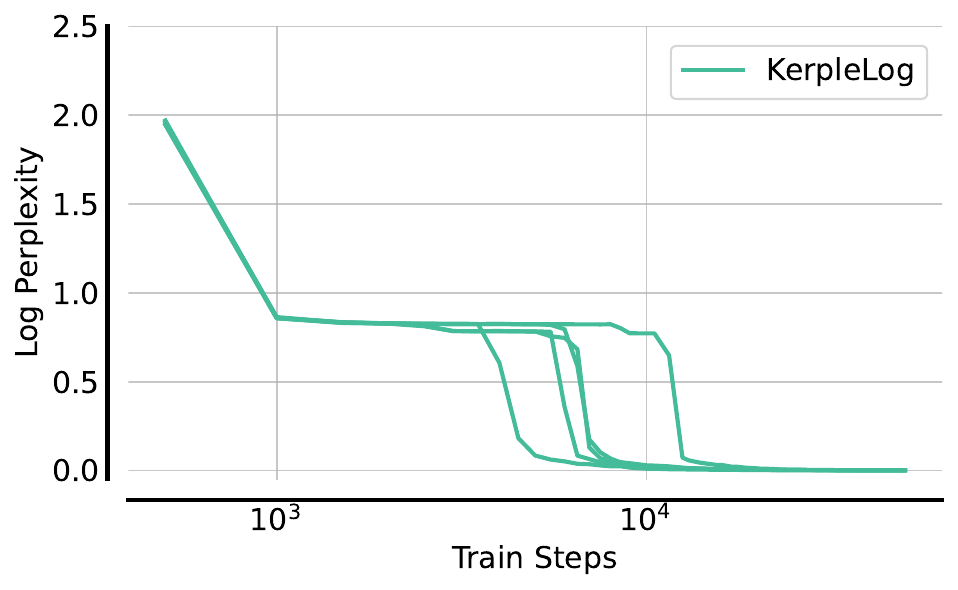}
\end{subfigure}
\begin{subfigure}[t]{0.45\textwidth}
\centering
\includegraphics[width=1.0\columnwidth]{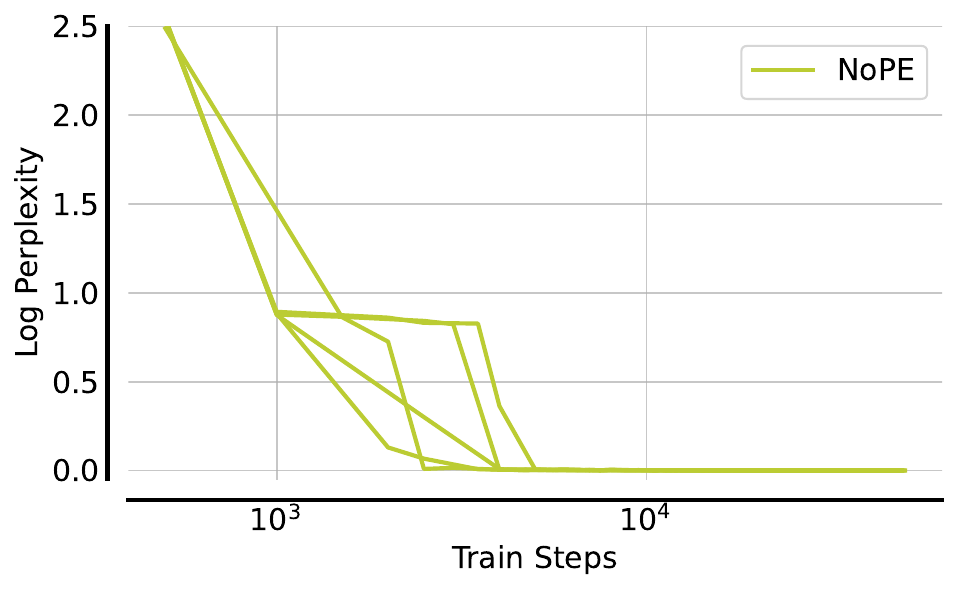}
\end{subfigure}
\begin{subfigure}[t]{0.45\textwidth}
\centering
\includegraphics[width=1.0\columnwidth]{figures/files/grokking_reverse_rs_nope_false_log}
\end{subfigure}
\begin{subfigure}[t]{0.45\textwidth}
\centering
\includegraphics[width=1.0\columnwidth]{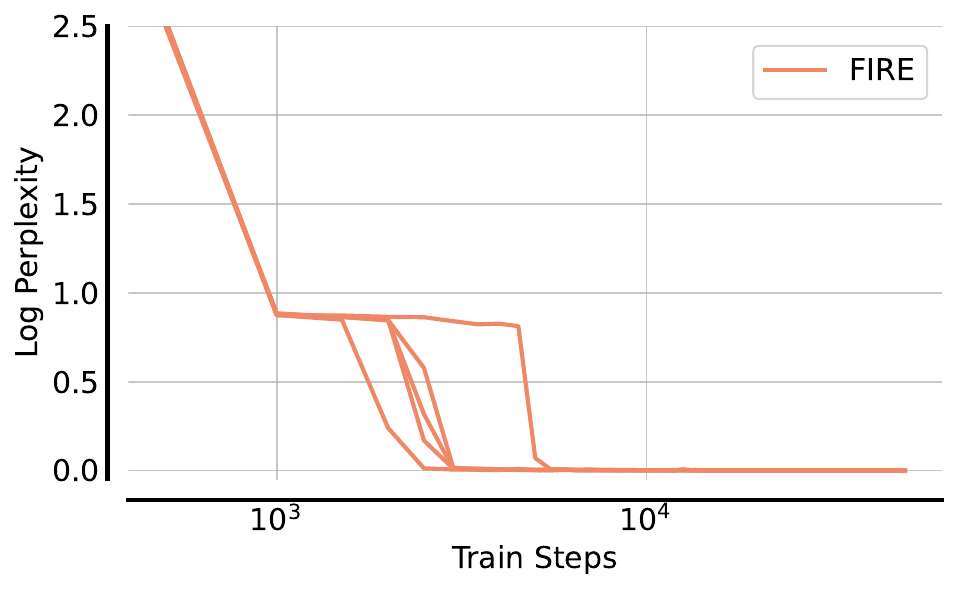}
\end{subfigure}
\begin{subfigure}[t]{0.45\textwidth}
\centering
\includegraphics[width=1.0\columnwidth]{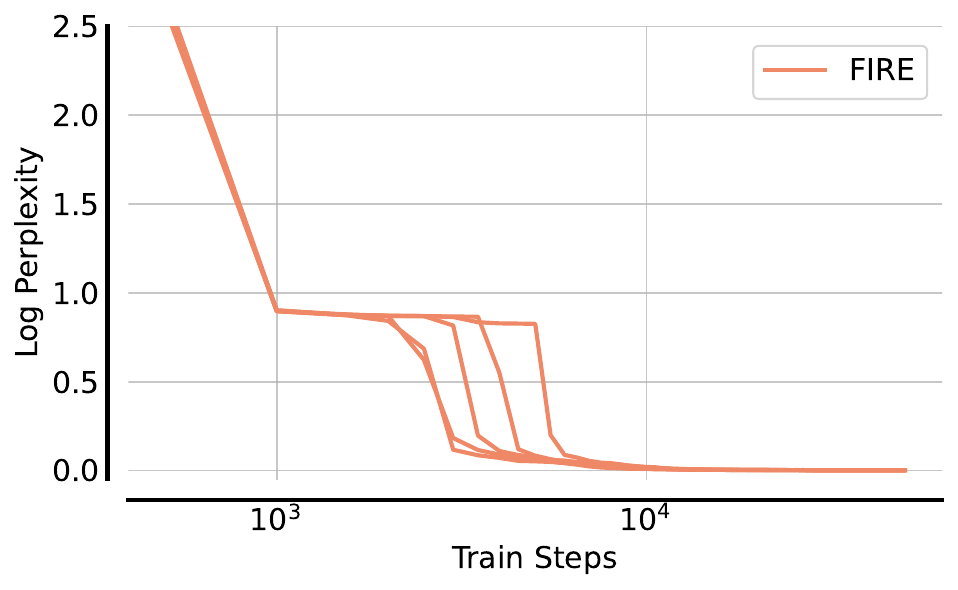}
\end{subfigure}
\vspace{-2mm}
\caption{(Left) Without Random Space Augmentation. (Right) With Random Space Augmentation. Training loss over 10 trials in reverse formats. Despite similar nearly 0 log perplexity losses across runs after 10K training steps, different runs exhibit very different length generalization.
}\label{fig_app:logpplx_logx_all_rs_tlen40}
\end{figure}

\clearpage
\newpage
\subsection{Randomized Position Encoding with Reverse Format with Index Hint trained up to 40-digit addition}\label{sec:loss_acc_reverse_index_hint_randomized_pe}
\begin{figure}[!h]
\centering
\begin{subfigure}[t]{0.45\textwidth}
\centering
\includegraphics[width=1.0\columnwidth]{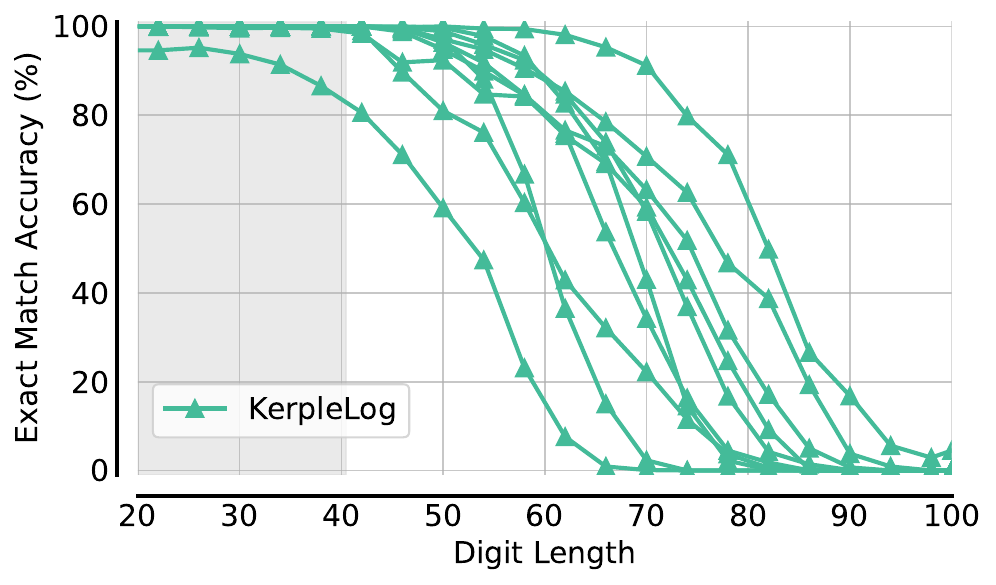}
\end{subfigure}
\begin{subfigure}[t]{0.45\textwidth}
\centering
\includegraphics[width=1.0\columnwidth]{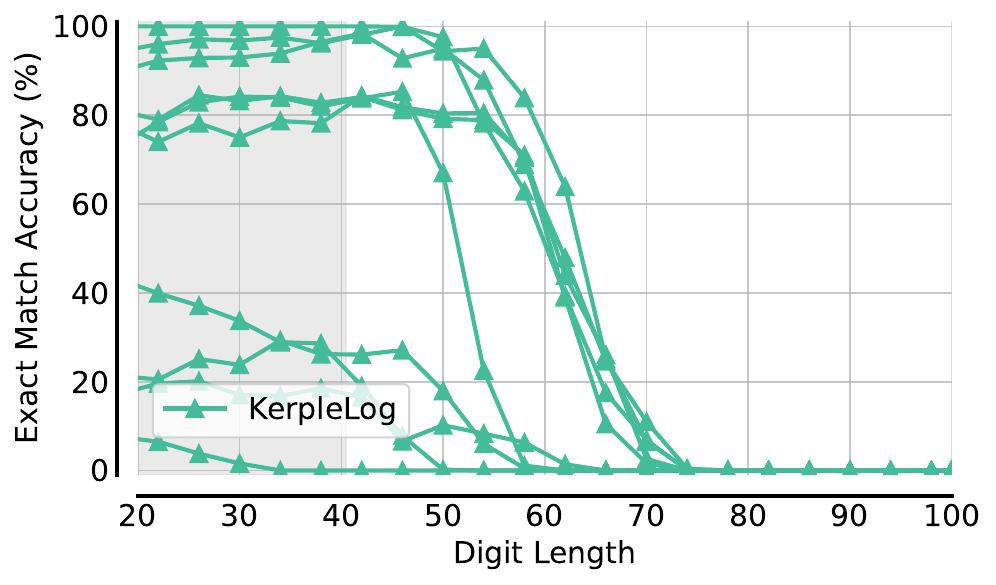}
\end{subfigure}
\begin{subfigure}[t]{0.45\textwidth}
\centering
\includegraphics[width=1.0\columnwidth]{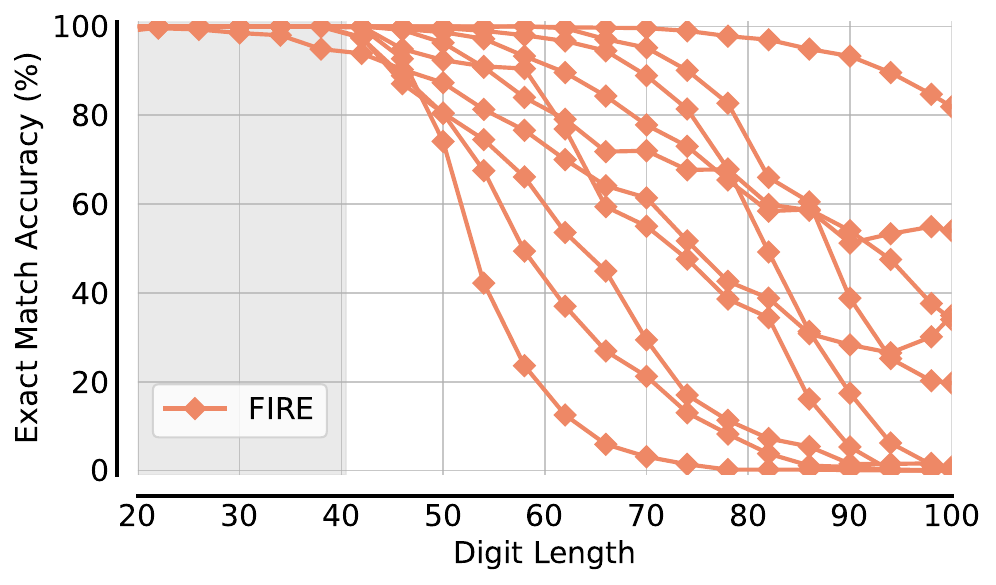}
\end{subfigure}
\begin{subfigure}[t]{0.45\textwidth}
\centering
\includegraphics[width=1.0\columnwidth]{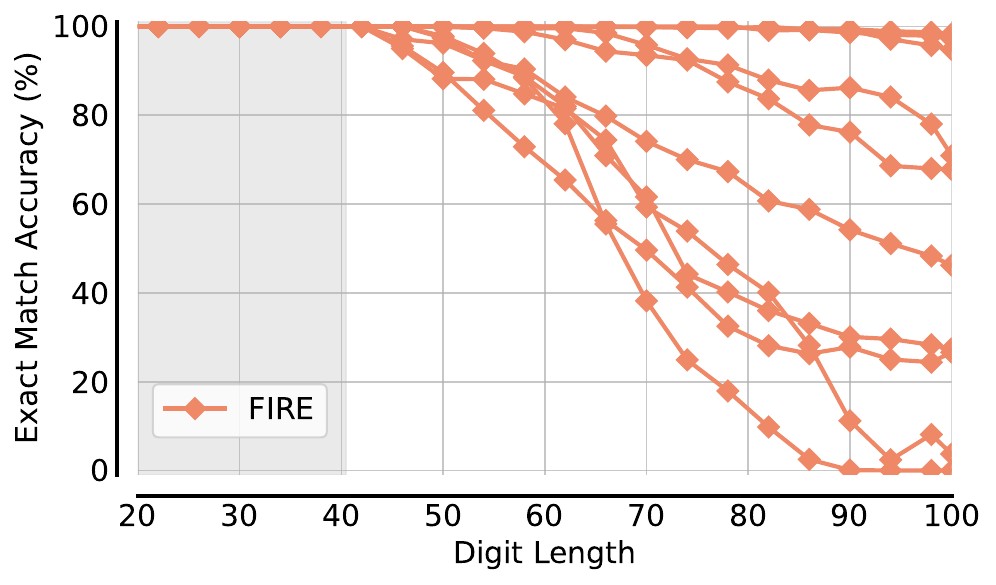}
\end{subfigure}
\vspace{-2mm}
\caption{(Left) Without Randomized Position Encoding (Right) With Randomized Position Encoding. Exact match accuracy on 20 to 100 digit addition of all 10 trials trained on up to 40-digit addition with index hint and reverse format using four different position encodings.
}\label{fig_app:seed_acc_tlen40_randomized_pe}
\end{figure}

\begin{figure}[!h]
\centering
\begin{subfigure}[t]{0.45\textwidth}
\centering
\includegraphics[width=1.0\columnwidth]{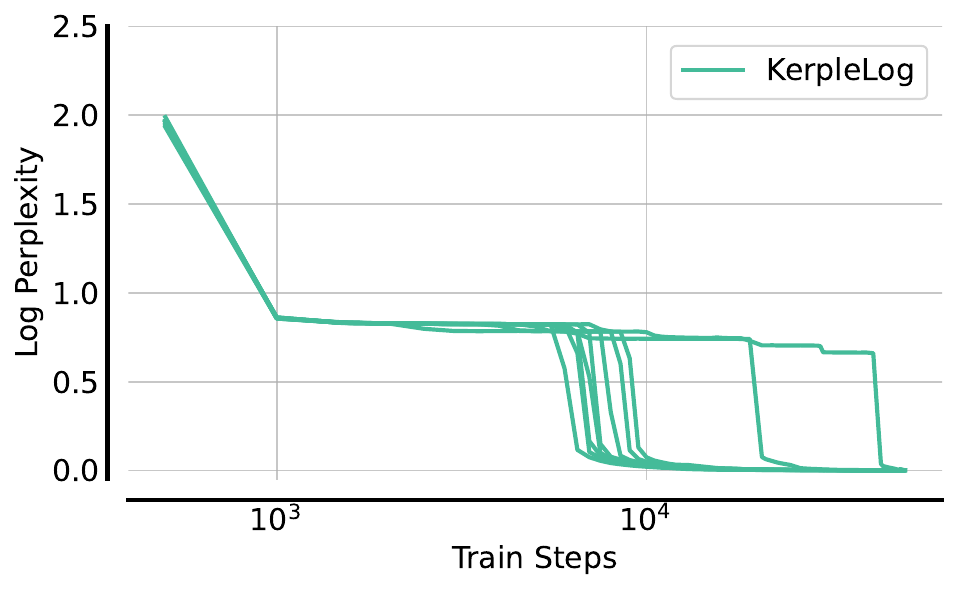}
\end{subfigure}
\begin{subfigure}[t]{0.45\textwidth}
\centering
\includegraphics[width=1.0\columnwidth]{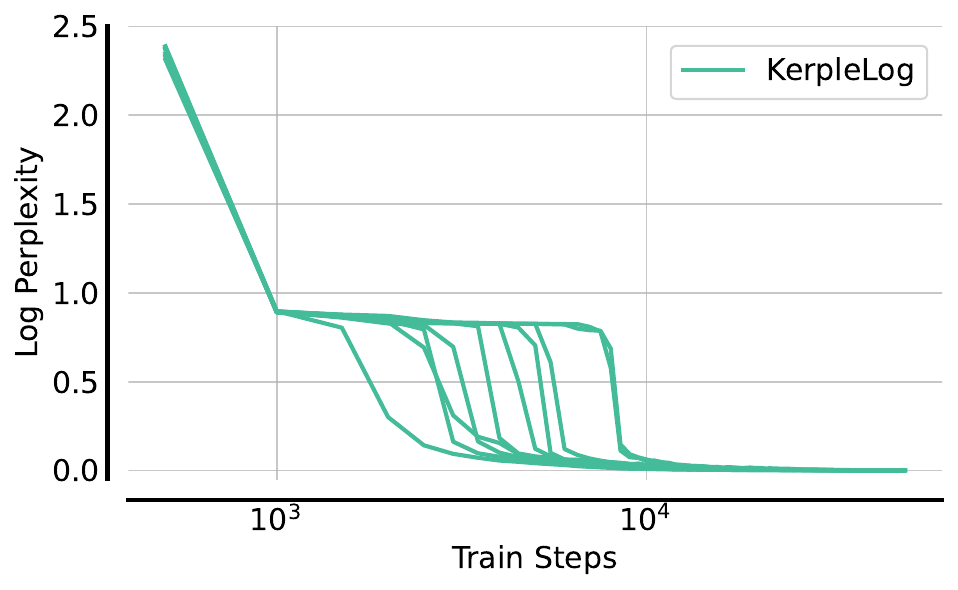}
\end{subfigure}
\begin{subfigure}[t]{0.45\textwidth}
\centering
\includegraphics[width=1.0\columnwidth]{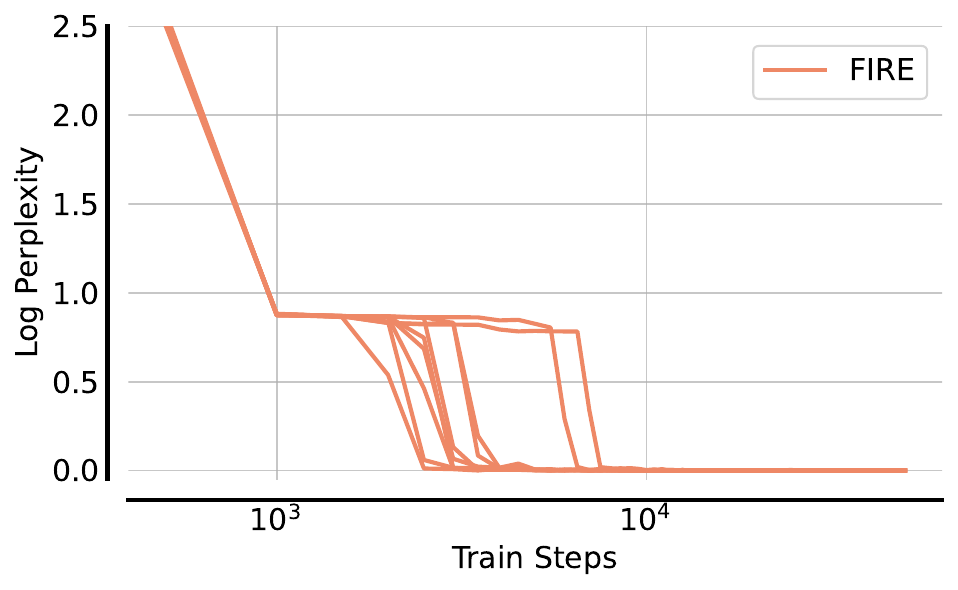}
\end{subfigure}
\begin{subfigure}[t]{0.45\textwidth}
\centering
\includegraphics[width=1.0\columnwidth]{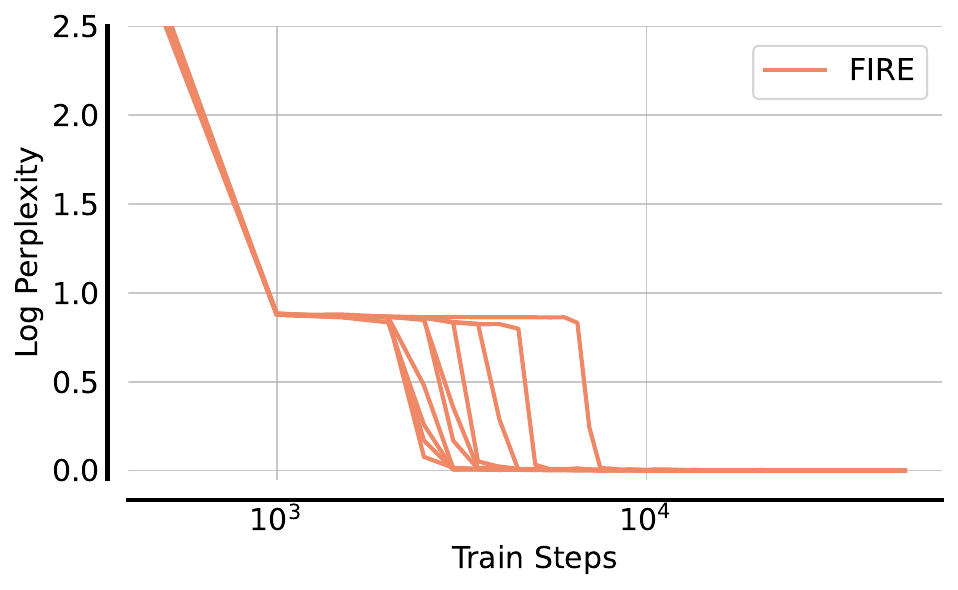}
\end{subfigure}
\vspace{-2mm}
\caption{(Left) Without Randomized Position Encoding (Right) With Randomized Position Encoding. Training loss over 10 trials in reverse formats.
}\label{fig_app:logpplx_logx_all_randomized_pe}
\end{figure}